\documentclass{article}

\usepackage[utf8]{inputenc} %

\usepackage[preprint]{jmlr_custom}
\makeatletter

\long\def\@makecaption#1#2{%
  \vskip 8pt%
  \begingroup
    \footnotesize
    \setbox\@tempboxa\hbox{\textbf{#1}: #2}%
    \ifdim \wd\@tempboxa >\hsize
      \noindent \textbf{#1}: #2\par
    \else
      \hbox to\hsize{\hfil\box\@tempboxa\hfil}%
    \fi
  \endgroup
}

\makeatother

\usepackage{graphicx} %
\usepackage{enumitem}
\setlist[enumerate]{label=(\roman*), leftmargin=*, itemsep=1pt, topsep=2pt}
\usepackage{microtype}
\usepackage{pdflscape}
\usepackage{amsmath}

\usepackage{etoolbox}
\BeforeBeginEnvironment{remark}{\vspace{6pt}}
\usepackage{amsfonts}
\usepackage{amssymb,amsthm}
\usepackage{bm}
\usepackage{bbm}
\usepackage{parskip}
\usepackage{rotating}
\usepackage{mathtools}
\usepackage{mathrsfs}
\usepackage{mathbbol}
\usepackage{xcolor}
\usepackage[dvipsnames]{xcolor}
\usepackage{adjustbox}
\usepackage[capitalize]{cleveref}
\usepackage{mathtools}
\usepackage[only,llbracket,rrbracket]{stmaryrd}

\usepackage{booktabs}  %
\usepackage{array}     %
\usepackage[most]{tcolorbox}  %

\usepackage{wrapfig}
\usepackage{tikz}
\usetikzlibrary{shapes.geometric, arrows.meta, positioning, calc}

\definecolor{intropath}{RGB}{93, 178, 96}
\definecolor{advancedpath}{RGB}{218, 153, 76}
\definecolor{expertpath}{RGB}{172, 109, 218}

\renewcommand{\arraystretch}{1.3}  %
\definecolor{rowgray}{gray}{0.95}  %

\definecolor{mybg}{HTML}{EAE9FF}   % light blue
\definecolor{outline}{HTML}{B5B2F8}  % slightly darker
\definecolor{mytitle}{HTML}{CAC8FF} % title background

\newtcolorbox{purplebox}[1][]{%
  enhanced,
  breakable,
  frame hidden,
  boxrule=0pt,
  parbox=false,
  borderline={1pt}{0pt}{outline},
  colback=mybg,
  colbacklower=mybg,
  arc=2mm,
  title={#1},
  fonttitle=\bfseries,
  coltitle=black, %
  attach boxed title to top left={xshift=2mm,yshift*=-2mm},
  varwidth boxed title,
  boxed title style={
    frame hidden,
    boxrule=0pt,
    colback=mytitle,
    arc=1mm,
    left=2mm, right=2mm, top=0.6mm, bottom=0.6mm
  }
}

\newcommand{\purple}[2][]{%
  \begin{purplebox}[#1]
    #2
  \end{purplebox}
}

\definecolor{customblue}{HTML}{4A90E2}

\definecolor{generalbg}{HTML}{FDF1DB}   
\definecolor{generaltitle}{HTML}{F3DEAA}

\newtcolorbox{generalbox}[1][]{%
  enhanced,
  breakable,
  frame hidden,
  boxrule=0pt,
  parbox=false,
  colback=generalbg,
  colbacklower=generalbg!4!gray,
  arc=2mm,
  title={#1},
  fonttitle=\bfseries,
  coltitle=black,
  attach boxed title to top left={xshift=2mm,yshift*=-2mm},
  varwidth boxed title,
  boxed title style={
    frame hidden,
    boxrule=0pt,
    colback=generaltitle,
    arc=1mm,
    left=2mm, right=2mm, top=0.6mm, bottom=0.6mm
  }
}

\usepackage{listings}
\usepackage{xcolor}

\definecolor{codebg}{rgb}{0.97,0.97,0.97}
\definecolor{codekw}{HTML}{50B2D7}
\definecolor{codestring}{HTML}{CD76FF}
\definecolor{codecomment}{HTML}{6B67EE}

\newcommand{\blfootnote}[1]{%
  \begingroup
  \renewcommand\thefootnote{}\footnote{#1}%
  \addtocounter{footnote}{-1}%
  \endgroup
}

\definecolor{figblue}{HTML}{3A62B4}
\definecolor{figred}{HTML}{C8143C}
\definecolor{figyellow}{HTML}{FDF2DC}
\definecolor{MyPurple}{HTML}{6B67EE}

\title{A Complete Guide to Spherical Equivariant Graph Transformers}

\author{%
\textbf{Sophia Tang}
\\[8pt]
{\normalfont Department of Computer and Information Science}\\
{\normalfont University of Pennsylvania}\\
{\normalfont Correspondence to: \href{sophtang@seas.upenn.edu}{\texttt{sophtang@seas.upenn.edu}}}
}

\begin{document}

\maketitle

\begin{abstract}
Spherical equivariant graph neural networks (EGNNs) provide a principled framework for learning on three-dimensional molecular and biomolecular systems, where predictions must respect the rotational symmetries inherent in physics. These models extend traditional message-passing GNNs and Transformers by representing node and edge features as spherical tensors that transform under irreducible representations of the rotation group SO(3), ensuring that predictions change in physically meaningful ways under rotations of the input. This guide develops a complete, intuitive foundation for spherical equivariant modeling — from group representations and spherical harmonics, to tensor products, Clebsch–Gordan decomposition, and the construction of SO(3)-equivariant kernels. Building on this foundation, we construct the Tensor Field Network and SE(3)-Transformer architectures and explain how they perform equivariant message-passing and attention on geometric graphs. Through clear mathematical derivations and annotated code excerpts, this guide serves as a self-contained introduction for researchers and learners seeking to understand or implement spherical EGNNs for applications in chemistry, molecular property prediction, protein structure modeling, and generative modeling.
\looseness=-1
\end{abstract}

\begin{center}
    \includegraphics[width=\linewidth]{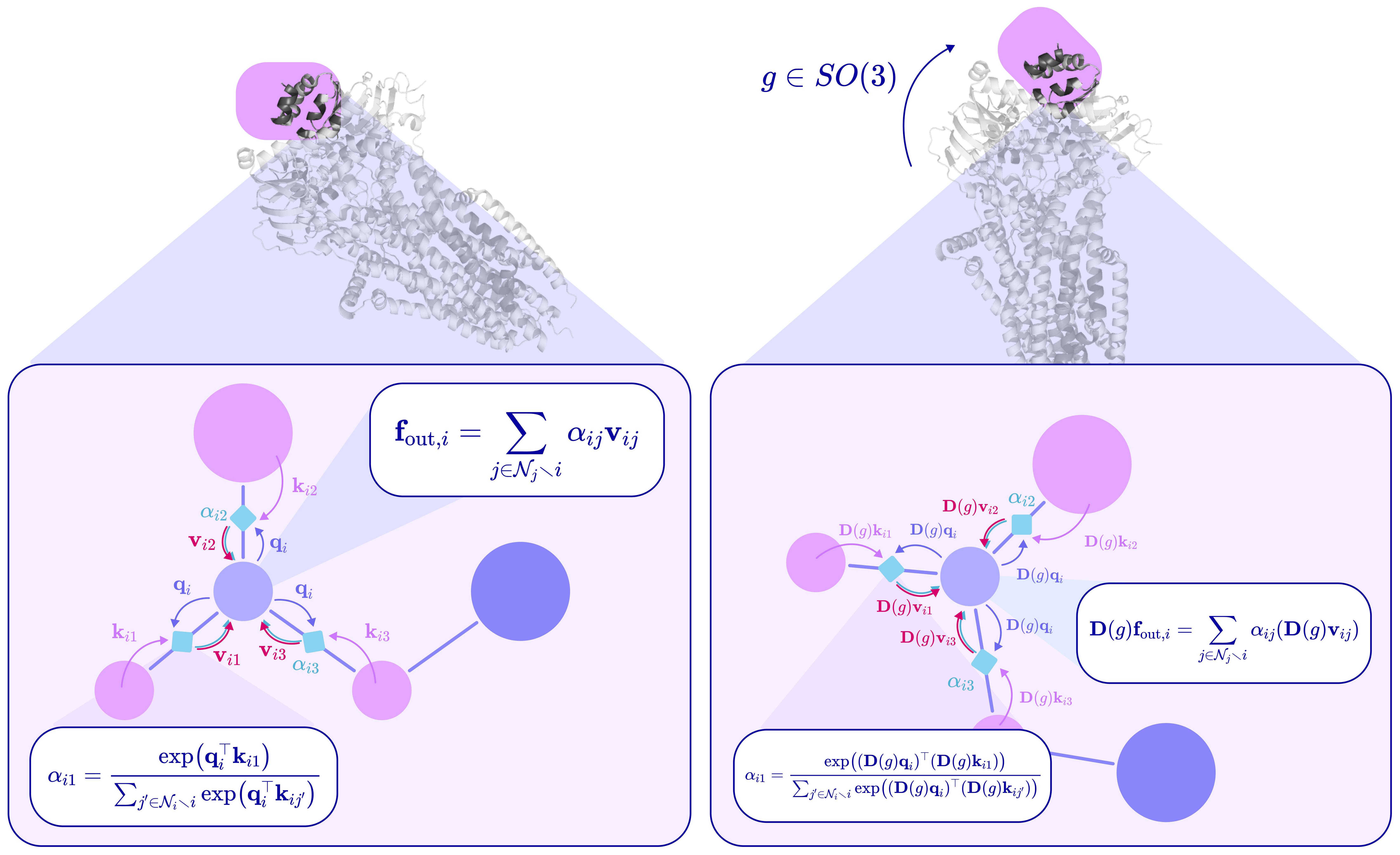}
    \label{fig:cover}
\end{center}

\newpage

\tableofcontents
\blfootnote{This paper is a technical version of the article originally published in Alchemy Bio\\ \href{https://alchemybio.substack.com/p/spherical-equivariant-graph-transformer}{\texttt{https://alchemybio.substack.com/p/spherical-equivariant-graph-transformer}}.}

\newpage

\section*{Introduction}
Modern machine learning models increasingly operate on three-dimensional data—from molecules and proteins to physical simulations, where the geometry of the system is fundamental to the task. Many properties of molecular systems do not depend on their absolute orientation in space, and the physical interactions between particles transform predictably under rotation. Standard neural architectures fail to capture these symmetries: a model trained on one orientation of a molecule may not recognize the same structure when rotated, leading to inefficiency, poor generalization, or physically inconsistent predictions. These limitations motivate the development of geometric deep learning models that respect the symmetry groups governing the underlying data.

Spherical equivariant graph neural networks \citep{geiger2022e3nn, thomas2018tensor,  fuchs2020se} address this need by enforcing SO(3) rotational equivariance at every stage of computation. Instead of treating features as arbitrary vectors, these models represent information as spherical tensors that transform under irreducible representations of SO(3). Using mathematical concepts from quantum mechanics, spherical EGNNs construct message-passing layers and attention mechanisms that guarantee physically meaningful behavior under rotation of the input graph. The result is an expressive yet symmetry-preserving framework that learns high-fidelity geometric relationships without data augmentation or hand-crafted invariants. 

A major application motivating this guide lies in representing biomolecular structures for applications across protein structure prediction \citep{jumper2021highly, baek2021accurate}, generative protein design \citep{watson2023novo, krishna2024generalized}, and molecular simulation \citep{batzner20223}, where interactions between atoms or residues depend on relative orientation and spatial geometry. However, the mathematical foundations underlying these architectures, such as spherical harmonics, irreducible representations, and equivariant kernel construction, are often presented in ways that are inaccessible to practitioners. The goal of this guide is to bridge this gap.

The sections are organized as follows:
\begin{itemize}
    \item Section \ref{sec:2} covers the \textbf{mathematical foundations} needed to understand how features transform under SO(3) rotations, including group theory, spherical tensors, spherical harmonics, tensor products, and Clebsch–Gordan coefficients.
    \item Section \ref{sec:3} derives the \textbf{SO(3)-equivariant kernel} by combining spherical harmonics, Clebsch–Gordan decomposition, and learnable radial functions, forming the basis for all rotation-equivariant message-passing layers.
    \item Section \ref{sec:4} explains how \textbf{Tensor Field Networks} implement equivariant message-passing and self-interaction using the kernels developed earlier, serving as a foundational architecture for spherical EGNNs.
    \item Section \ref{sec:5} extends TFNs into an \textbf{attention-based architecture} by constructing queries, keys, and values as spherical tensors to enable invariant attention scores and equivariant value updates.
    \item Section \ref{sec:6} demonstrates how SE(3)-Transformers are applied to real molecular data by detailing graph construction, model components, and training procedures on the QM9 dataset.
\end{itemize}

Throughout the guide, mathematical derivations are paired with annotated PyTorch and DGL implementations to demonstrate how theoretical concepts translate into practical code. By the end, readers will possess a deep understanding of spherical equivariant neural networks and the complete toolkit needed to implement them for molecular property prediction, biomolecular modeling, and other 3D learning tasks.

\newpage

\section{Preface}
\label{sec:1}

The majority of the notation used in this article aligns with the convention used in the SE(3)-Transformers paper \citep{fuchs2020se}. Here, we include specific indices distinguishing which kernels are unique, and the extension to multiple channels of each feature type for completeness. The meanings of the common mathematical symbols used in the sub- and superscripts of weights, kernels, vectors, and feature tensors are given in the table below.

\vspace{0.5em}
\begingroup
\renewcommand{\arraystretch}{1.25}
\begin{center}
\begin{tabular*}{\textwidth}{@{\extracolsep{\fill}}>{\centering\arraybackslash}p{1.4in} p{4.5in}@{}}
\hline
\textbf{Notation} & \textbf{Meaning} \label{table:notation1}
\\
\hline
in, out   & input features and output features (after message-passing) \\
$i$ & center node or destination node \\
$j$ & nodes in the neighborhood of node $i$ with an outgoing edge pointing towards node $i$ \\
$k$ & the type/degree of node features from the source or neighborhood nodes \\
$c_k$ & index of the type-$k$ feature channel \\
$l$ & the type/degree of node features from the center node \\
$c_l$ & index of the type-$l$ feature channel \\
$m_l$, $m_k$, $m$ & indices of the elements of the type-$l$, type-$k$, and type-$J$ spherical tensors, which also correspond to magnetic quantum numbers corresponding to the angular momentum numbers $l$, $k$, and $J$ \\
$J$ & the intermediate feature types for spherical harmonics projections ranging from $|k - l|$ to $|k + l|$ \\
$ij$ &  denotes an edge feature (displacement vector) or embedding stored in the edge (messages or key and value embeddings) from the neighborhood node $j$ to the center node $i$ \\
$l k$ & denotes equivariant kernels that transform tensors from type-$k$ to type-$l$ features \\
$\mathbf{Q}$, $\mathbf{K}$, $\mathbf{V}$ & denotes the kernels that transform features into query, key, and value embeddings, respectively \\
mi & total input channels (or multiplicity) of degree di \\
mo & total output channels of degree do \\
\hline
\end{tabular*}
\end{center}
\endgroup
\vspace{0.5em}

Throughout the article, annotated code excerpts from the official implementation of the SE(3)-Transformer \citep{fuchs2020se} are provided with slight modifications for clarity\footnote{Full code implementation found at \href{https://github.com/FabianFuchsML/se3-transformer-public}{\texttt{https://github.com/FabianFuchsML/se3-transformer-public}}}. We will walk through the majority of the code, but placing more emphasis on the mathematics and intuition surrounding the code.

The code uses a Python library called \textbf{Deep Graph Library (DGL)} \citep{wang2019deep} that allows the construction and handling of graph data. The library supports User-Defined Functions (UDFs) that enable users to construct novel functions that can be applied for message-passing across the entire graph (much of the code described here is encapsulated in a UDF)\footnote{Documentation for DGL can be found at \href{https://www.dgl.ai/dgl_docs/graphtransformer/index.html}{\texttt{https://www.dgl.ai/dgl\_docs/graphtransformer/index.html}}}. Here are some basic DGL notations that are useful in interpreting the code throughout the guide:

\begin{lstlisting}[language=Python]
import dgl
import dgl.function as fn

# initialize dgl graph
G = dgl.DGLGraph() 

# retrieve data from all nodes labeled ntype
G.ndata[ntype] 

# retrieve output features of type d from node data 
G.ndata[f'out{d}'] 

# retrieve data from all edges labeled etype
G.edata[etype] 

# retrieve edge kernels that transform from type di to type do
G.edata[f'({di},{do})'] 

# msg_func generates messages along edges and reduce_func aggregates the messages to send to the destination node
G.update_all(msg_func, reduce_func)

# calling built-in dgl function e_dot_v that computes a message on an edge by performing element-wise dot between features of e and v, and stores it as edge message labeled 'm
f = fn.e_dot_v('e', 'v', 'm') 

# applies the function f to update the features of the edges with the function
G.apply_edges(f)
\end{lstlisting}

This guide will discuss geometric tensors (spherical tensors and Cartesian tensors) and working with the tensor data structure in PyTorch, which refers to multidimensional arrays. To distinguish the two, we will use ‘\textbf{tensor}’ when referring to geometric tensors and ‘\textbf{arrays}’ when referring to the data structure\footnote{When describing the shape of an array, the batch size (which is typically the first dimension) is often omitted for simplicity, so the shape refers to the final dimensions of the tensor that are relevant to calculations.}.

\newpage
\section{Preserving Rotational Equivariance}
\label{sec:2}
Understanding how to preserve rotational equivariance is the primary challenge of understanding geometric GNNs. Conventional deep learning models generate predictions based on learned features on a \textbf{fixed reference frame}, but fail to detect those same features after transformations in space. 

Rotational symmetries are rooted in all physical systems, especially on the molecular scale. Thus, constructing models that understand how interactions between nodes change under rotation is critical for tasks involving biomolecular systems.

\subsection{Invariance and Equivariance}
\purple[]{Invariance and equivariance are the cornerstones of geometric GNNs because they describe how convolutions or filters must be constructed to recognize patterns and generate predictions on a global reference frame where the graph can appear in any location or orientation in space, but encode data that is the same or differing by a predictable transformation.}

When a function produces the \textbf{same} output for a given input regardless of its orientation or position in space, it \textbf{preserves invariance}. A feature of a physical system can also be called invariant if it does not change with permutations or rotations (e.g., atomic number, bond type, number of protons). Only functions that preserve invariance should be applied to invariant features.

For instance, the potential energy of an isolated molecule in a vacuum is constant no matter its orientation or position in space; therefore, a function that calculates the potential energy given a molecule input should produce the same value no matter its position or orientation; in other words, it should preserve invariance.

When a function produces a \textbf{predictable transformation of the original output as a direct consequence of a transformation on the input} (i.e., translation or rotation), \textbf{it preserves equivariance}.

A node feature is equivariant if it transforms predictably under transformations in the node’s position, and similarly, an edge feature is equivariant if it transforms predictably under transformations in either node it connects. Features on the node level of a geometric graph are often equivariant as they change with changes to the relative position and orientation of nodes, whereas system-level properties across the entire system are generally invariant. Only functions that preserve equivariance should be applied to equivariant features.

All chemical features represented by vectors (e.g., position, velocity, external forces on individual atoms) are equivariant, as they should transform with transformations in the input. In a molecule, changing the position of a negatively charged atom changes the direction of the attractive force between it and nearby charged atoms.

Preserving equivariance is crucial for modeling molecular systems, as their behavior is governed by conserved quantities of quantum mechanics, like angular momentum and energy, that follow strict sets of physical laws with inherent rotational symmetries.

Transformation equivariant models ensure that three-dimensional spatial transformations (i.e., translations in $x$, $y$, and $z$ directions and rotations around any axis) of the input graph structure and features result in predictable transformations in the model’s output using functions (called \textbf{kernels}) that learn patterns across positional and feature data and apply them equivariantly across the entire graph. These functions are applied across the graph and are constructed to handle location and orientation-invariant or equivariant features without needing to be trained on rotated or translated data.

Similar to how a convolutional neural network (CNN) \citep{o2015introduction} applies position-invariant filters to detect two-dimensional motifs regardless of their position in the input (i.e., image, heatmap), equivariant GNNs apply the same transformation-equivariant filters to detect motifs regardless of translations and rotations in 3D space.

\subsection{Group Representations and Transformation Operators}

A \textbf{group} in mathematics is defined as a \textbf{set}, denoted as $G$, of abstract actions (e.g., rotations and translations) and a \textbf{binary operation} $ab$ for all $a, b \in  G$ that operates on the elements in the set such that the following conditions hold:

\begin{enumerate}
    \item \textbf{Closure}: for all $a, b \in G$, the output of the binary operator is also in the set, $ab \in  G$.
    \item \textbf{Associativity}: for all $a, b, c \in  G$, the following equation holds: $(ab)c = a(bc)$
    \item \textbf{Identity Element}: every group has an identity element e that returns the element unchanged when applied to any element in the set with the binary operator. In other words, for all $a \in  G$, $ea = ae = a$.
    \item \textbf{Inverse Element}:  for all $a \in G$, there exists an inverse of $a$ (denoted as $a^{-1}$) in G such that $aa^{-1} = a^{-1}a = e$ (identity element). Note that a is also the inverse of $a^{-1}$.
\end{enumerate}

A \textbf{group representation} converts the abstract elements of a group into a set of $N \times N$ invertible square matrices, denoted as $GL(N)$. A group representation is generated via a \textbf{group homomorphism} $\rho: G \to GL(N)$ that takes a group as input and outputs a set of $N \times  N$ matrices corresponding to each group element while \textbf{preserving the function of the binary operator}:

\begin{align}
    \rho(ab)=\rho(a)\rho(b)\tag{$\forall a, b\in G$}
\end{align}

These representations can also be interpreted as \textbf{injective transformation operators} that act on $N$-dimensional vectors in a specific subspace, mapping the vector from \textbf{one point in the subspace to another point in the same subspace}. With the idea of group representations as transformation operators, we can define invariant and equivariant functions.

A function $\mathbf{W}$ that acts on a vector $\mathbf{f}$ is \textbf{invariant under a group} $G$ if the output is the same before and after the action $g \in  G$ is applied to the input, for all actions in the group.

\begin{align}
    \mathbf{W}(\rho_k(g)\mathbf{f})=\mathbf{W}\mathbf{f}\tag{$\forall\mathbf{f}\in\mathcal{X}_k,g\in G$}
\end{align}

A function is \textbf{equivariant under group} $G$ if the output of the function also \textbf{undergoes the same action} $g$ when the input is transformed by $g$.

\begin{align}
    \rho_l(g)(\mathbf{W}\mathbf{f})=\mathbf{W}(\rho_k(g)\mathbf{f})\tag{$\forall\mathbf{f}\in\mathcal{X}_k,g\in G$}
\end{align}

$\rho_k$ and $\rho_l$ are group representations of $G$, where $\rho_k$ acts in the same subspace $\mathcal{X}_k$ as the vector $\mathbf{f}$ and $\rho_l$ acts in the same subspace $\mathcal{X}_l$ as the output of the function $\mathbf{W}$.

\begin{align}
    \rho_l(g):\mathcal{X}_l\to\mathcal{X}_l,\;\rho_k(g):\mathcal{X}_k\to\mathcal{X}_k, \mathbf{W}: \mathcal{X}_k\to\mathcal{X}_l
\end{align}

\begin{figure}
    \centering
    \includegraphics[width=0.6\linewidth]{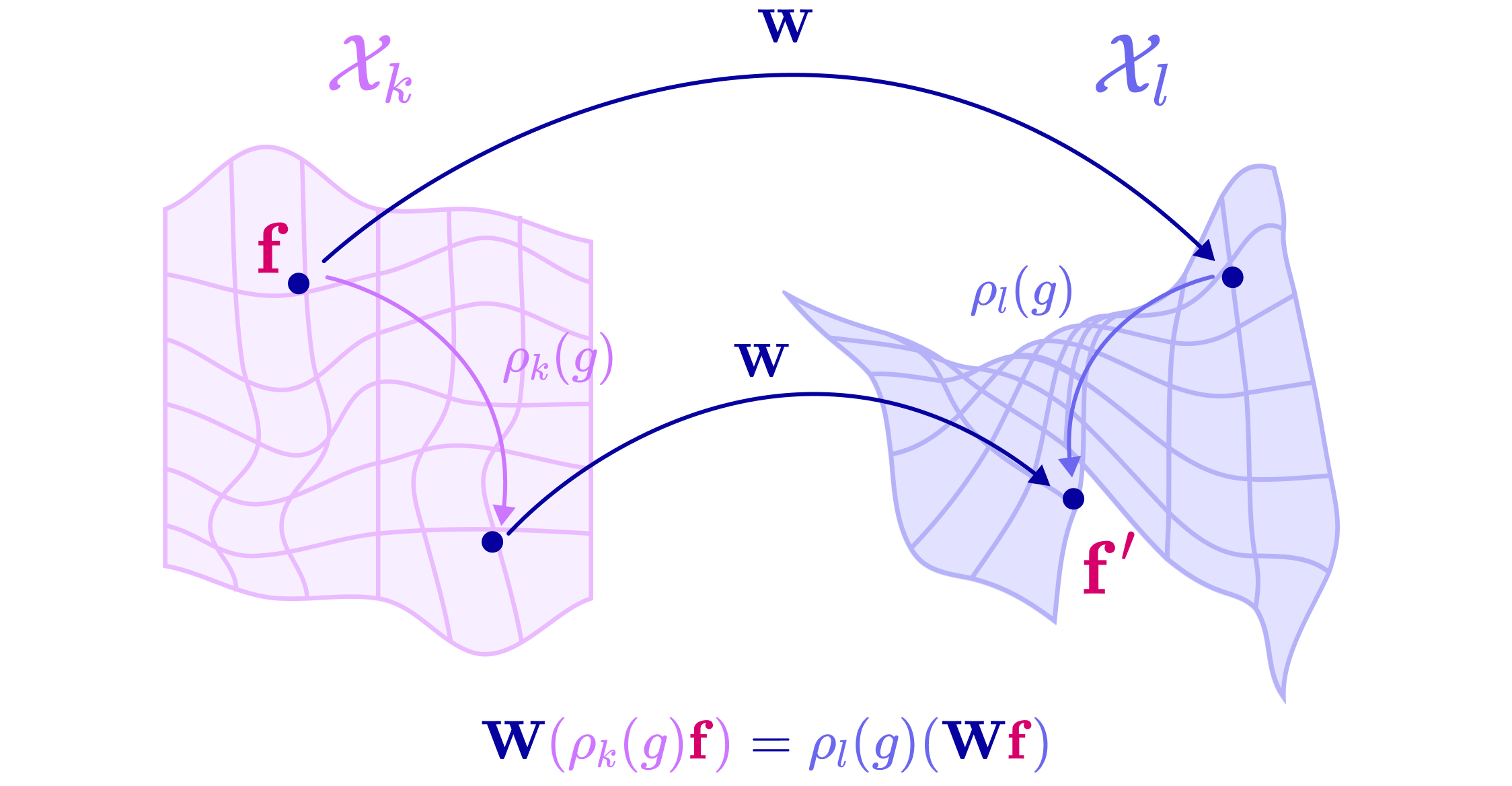}
    \caption{Diagram showing the commutative property of an equivariant function $\mathbf{W}$ that transforms a tensor $\mathbf{f}$ from one tensor space $\mathcal{X}_k$ to another tensor space $\mathcal{X}_l$.}
    \label{fig:equivariant}
\end{figure}

Applying two or more equivariant functions subsequently (or \textbf{composing the functions}) still satisfies the equivariance condition:

\begin{align}
    \mathbf{W}_2\big(\mathbf{W}_1(\rho_k(g)\mathbf{f})\big)&=\mathbf{W}_2\big(\rho_l(g)(\mathbf{W}_1\mathbf{f})\big)\\
    &=\rho_t(g)\big(\mathbf{W}_2(\mathbf{W}_1\mathbf{f})\big)\tag{$\forall\mathbf{f}\in\mathcal{X}_k,g\in G$}
\end{align}

$\mathbf{W}_2$  is a second equivariant function applied after $\mathbf{W}_1$ that transforms the input from the subspace $\mathcal{X}_l$ to the subspace $\mathcal{X}_t$. 

\begin{align}
    \rho_t(g):\mathcal{X}_t\to\mathcal{X}_t,\mathbf{W}_1: \mathcal{X}_k\to\mathcal{X}_l,\mathbf{W}_2: \mathcal{X}_l\to\mathcal{X}_t
\end{align}

This property allows us to compose as many equivariant functions as we want without worrying about breaking equivariance.

\begin{figure}[h!]
    \centering
    \includegraphics[width=0.8\linewidth]{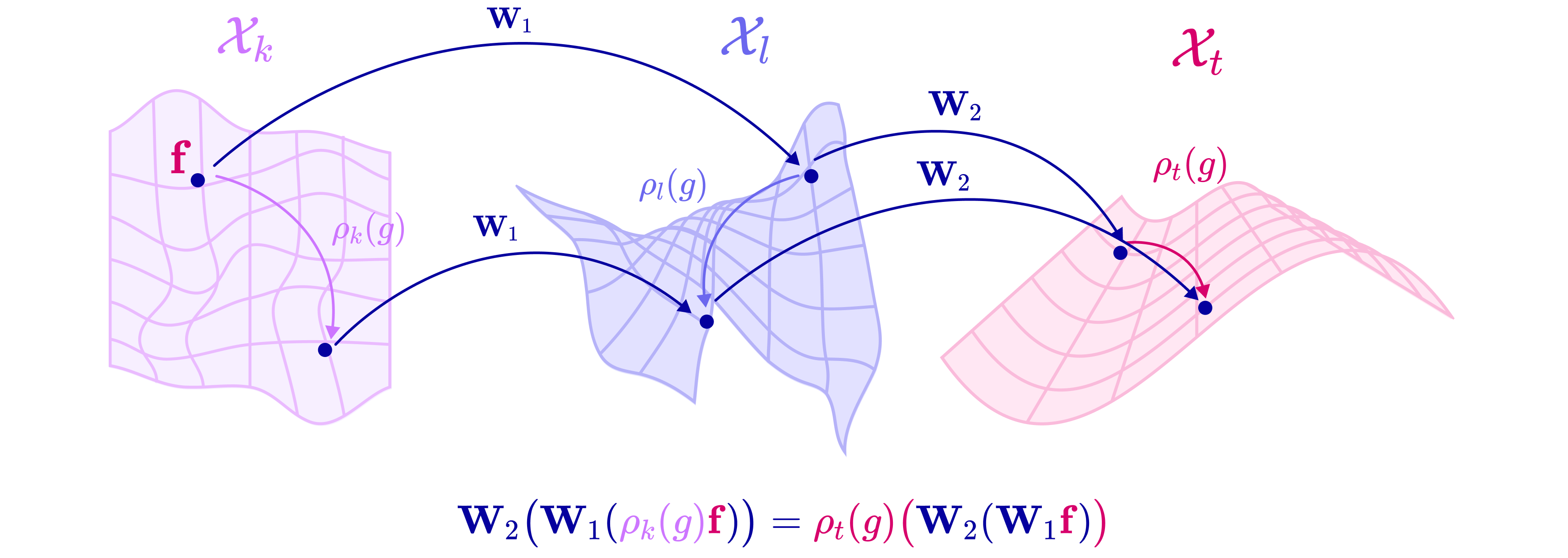}
    \caption{Diagram showing the commutative property of composing two equivariant functions $\mathbf{W}_1$ and $\mathbf{W}_2$ such that applying the group representation on the input before the two subsequent functions produces the same output as applying the group representation on the output of the composed functions.}
    \label{fig:equi2}
\end{figure}

Functions in geometric GNNs should satisfy three types of equivariance in 3D: \textbf{permutation} equivariance, \textbf{translation} equivariance, and \textbf{rotation} equivariance. \textbf{Permutation equivariance} states that permuting the indices of nodes should permute the output or produce the same output (permutation invariance). In most GNNs, nodes are treated as sets of objects rather than an ordered list, so these models are inherently permutation invariant.

Since geometric graphs represent isolated systems defined by relative displacement vectors and not absolute spatial information, geometric GNNs are by default \textbf{translation invariant}, meaning shifting the position of all nodes in the graph by a displacement vector does not change the output. Unfortunately, satisfying \textbf{rotational equivariance} in 3D is a lot more challenging, making it the focus of advancements in equivariant models.

\subsection{Spherical Tensors}

The \textbf{Special Euclidean Group in 3D}, known as the \textbf{SE(3) group}, is the set of all rigid 3D transformations, including rotations and translations. We will focus on the subset of SE(3) called the \textbf{Special Orthogonal Group in 3D}, known as the \textbf{SO(3) group}, which is the set of all 3D rotations.

\begin{figure}[h!]
\centering
\includegraphics[width=0.8\linewidth]{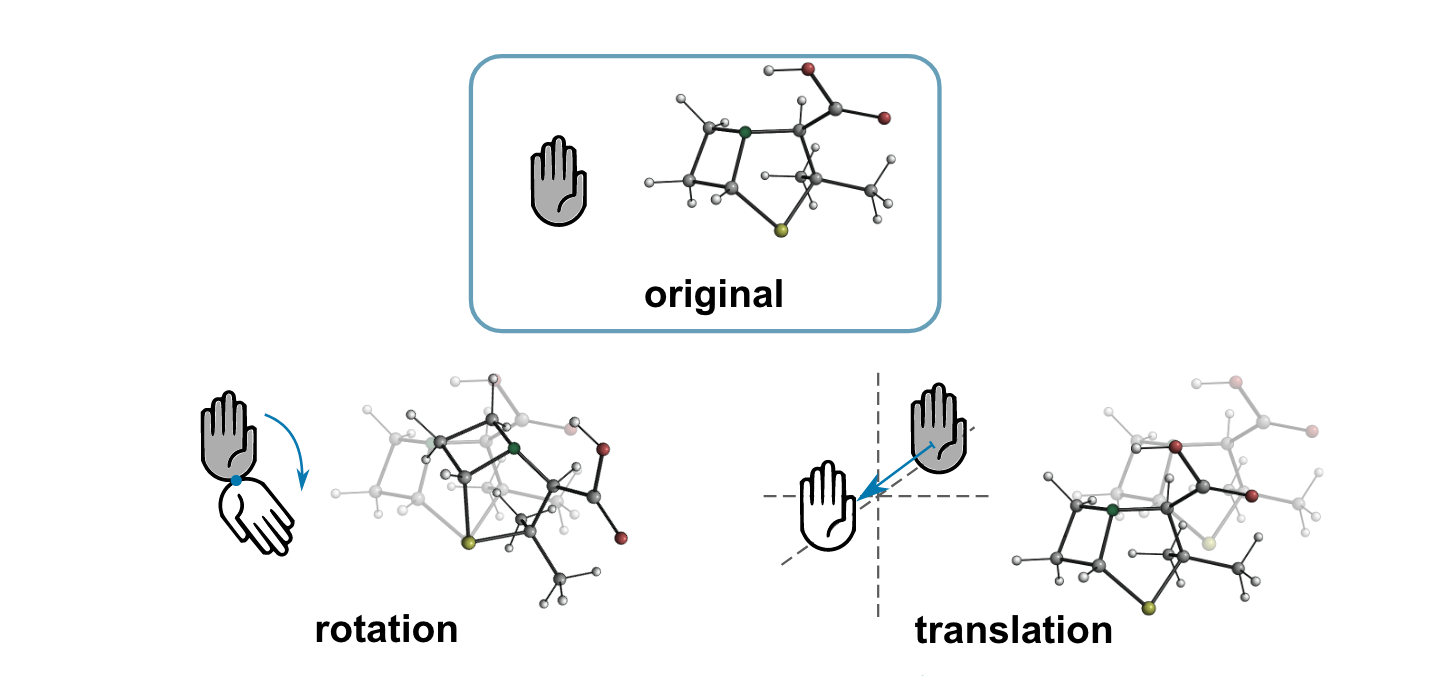}
\caption{Molecules transformed under the SE(3) group including rotations (SO(3) group) and translations. (Source: \citet{atz2021geometric})}
\label{fig:equi3}
\end{figure}

A representation of SO(3) is a set of invertible $N \times N$ square matrices that assign a specific matrix to every possible 3D rotation defined by the three Euler angles alpha $\alpha$, beta $\beta$, and gamma $gamma$, which define the rotation angles about the $x$, $y$, and $z$-axes, respectively. These matrices are orthogonal and have a determinant of 1, meaning they \textbf{preserve length and relative angles between vectors}.

Since higher-dimensional tensors require more complex representations, we need a way of decomposing complex representations into smaller building blocks that can be used to rotate across tensors of increasing dimensions. These building blocks are called irreducible representations (\textit{irreps}) of SO(3), which is a subset of rotation matrices that can be used to construct larger rotation matrices that operate on higher-dimensional tensors.

All group representations can be decomposed into the \textbf{direct sum} $\otimes$ (concatenation of matrices along the diagonal) of \textit{irreps}. This block diagonal matrix can then be used to transform a higher-dimensional tensor after first applying an $N \times N$ change of basis matrix $\mathbf{Q}$. Thus, we can write all representations of SO(3) in the following form:

\begin{align}
    \mathbf{D}(g)=\mathbf{Q}^{\top}\left[\bigoplus_J\mathbf{D}_J(g)\right]\mathbf{Q}
\end{align}

In the equation above, $\mathbf{Q}$ decomposes the input tensor into a direct sum of type-$J$ spherical tensors aligned with each block in the block-diagonal Wigner-D matrix, and the transpose of $\mathbf{Q}$ converts the rotated spherical tensors back into their original basis.

There is a special subset of tensors called \textbf{spherical tensors} that transform directly under the \textit{irreps} of SO(3) \textbf{without the need for a change in basis}. Spherical tensors are considered \textbf{irreducible types} because all Cartesian tensors can be decomposed into their spherical tensor components, but spherical tensors cannot be decomposed further. Spherical tensors have \textbf{degrees} numbered by non-negative integers $l$ that we call the tensor \textbf{type}. Type-$l$ tensors are $(2l + 1)$-dimensional vectors that transform under a corresponding set of type-$l$ irreducible representations of SO(3). We will describe both of these ideas more explicitly in the upcoming sections.

\purple[Spherical Tensors in Quantum Physics]{In quantum physics, spherical tensors are used to represent the \textbf{orbital angular momentum of quantum particles} like electrons. The degree of spherical tensors corresponds to the \textbf{angular momentum quantum number} (conventionally denoted with the letter $l$) that indicates the magnitude of angular momentum. The dimensions correspond to the $(2l + 1)$ possible \textbf{magnetic quantum numbers}, which have integer values ranging from $-l$ to $l$ (denoted with the letter $m$) and are equal to the projected angular momentum on the $z$-axis relative to an external magnetic field. 

Since angular momentum can be represented as a vector in 3D space, the value of $m$ must be between $-l$ (directly opposing the magnetic field) and $l$ (perfectly aligned with the magnetic field), and since angular momentum is quantized, $m$ \textbf{must be an integer}. In physics, $|l,m⟩$ is used to denote a specific dimension $m$ of a type-$l$ spherical tensor which represents an \textbf{eigenstate} of a quantum particle, where the angular momentum is considered to be \textbf{well-defined} (more on this later).}

\subsection{From Point Cloud to Geometric Graph}

A \textbf{point cloud} is a \textbf{finite set of 3D coordinates} (or 3-dimensional vectors) where every point has a corresponding \textbf{feature vector}. Nodes can represent atoms in molecules, residues (C-alpha atoms) in proteins, or any unit in a system that carries information about itself in the form of feature tensors.

The feature vector $\mathbf{f}$ corresponding to each node in the point cloud contains data on its properties (e.g., atomic number, charge, hydrophobicity, etc.). The feature vector can be arranged into a \textbf{tensor list}, a multi-dimensional list of spherical tensors with three axes in the following order: a \textbf{tensor} axis, a \textbf{channel} axis, and a \textbf{tensor-component} axis.

\begin{figure}[h!]
    \centering
    \includegraphics[width=0.6\linewidth]{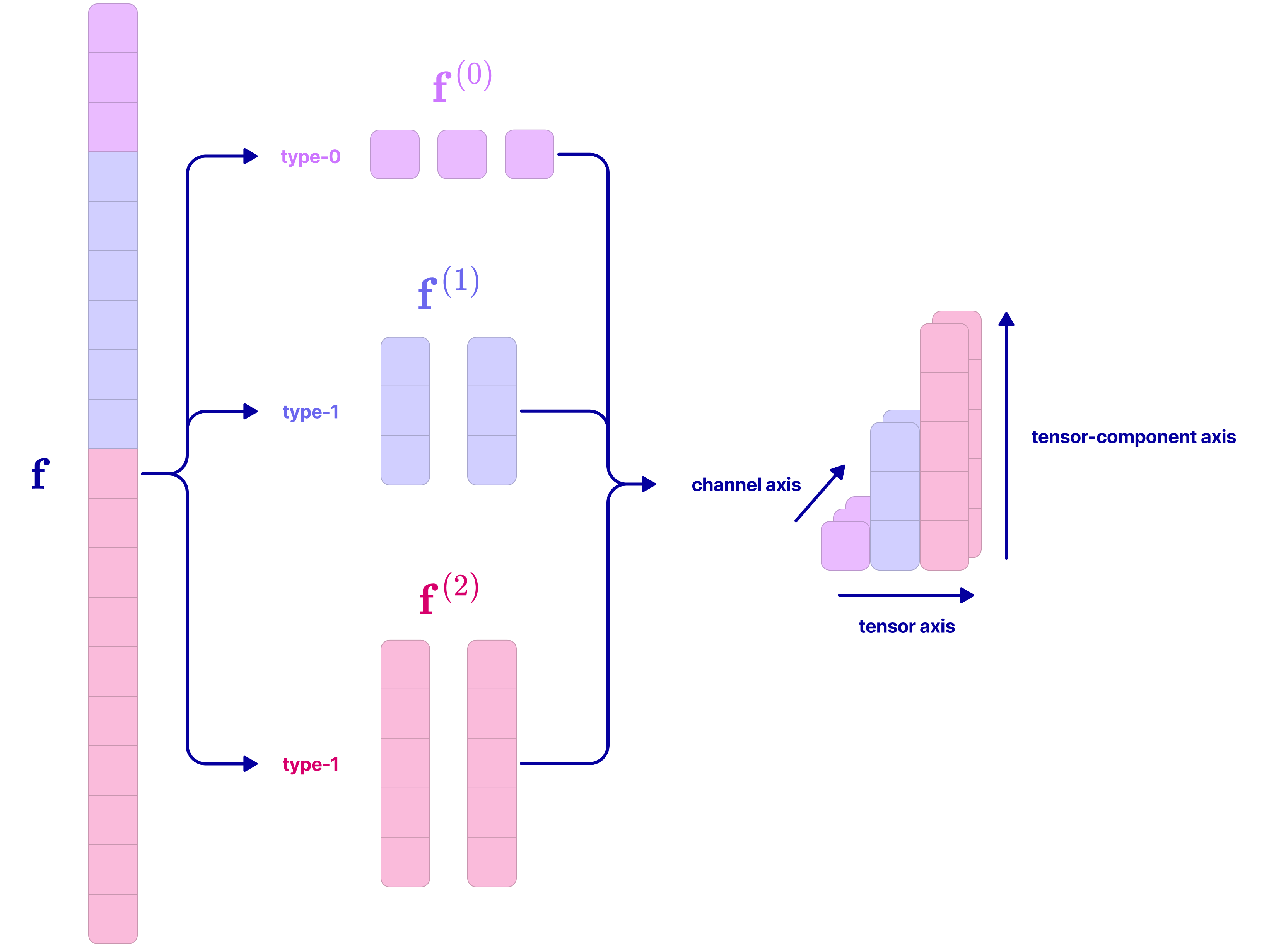}
    \caption{A feature vector $\mathbf{f}$ is split into its type-0, type-1, and type-2 components and arranged into a feature tensor with a tensor axis, a channel axis, and a tensor-component axis.}
    \label{fig:feature}
\end{figure}

\begin{enumerate}
    \item The \textbf{channel axis} represents the number of features of each type at the node. If a node contains three type-2 features, that feature has 3 channels.
    \item The \textbf{tensor axis} represents the number of different types of spherical tensor features at the node. If a node has type-0, type-1, and type-2 features, it has a tensor axis dimension of 3.
    \item The \textbf{tensor-component axis} represents the dimensions of each type of spherical tensor. For the type-$k$ spherical tensors at the node, the tensor-component axis has dimension $2k + 1$. Type-0 tensors are 1-dimensional scalars, type-1 tensors are 3-dimensional vectors, type-2 tensors are 5-dimensional vectors, and type-$k$ tensors are $(2k+1)$-dimensional vectors.
\end{enumerate}

From the point cloud, we want to generate a \textbf{directional graph} that contains directional edges between nodes that point in the direction of message-passing. An edge in the point cloud is defined by a \textbf{displacement vector that points from node} $j$ (source node or neighborhood node) \textbf{to node} $i$ (destination node or center node).

\begin{align}
    \mathbf{x}_{ij}=\mathbf{x}_i-\mathbf{x}_j\tag{directional edge}
\end{align}

This can be decomposed into the \textbf{radial distance} (scalar distance between nodes) and the \textbf{angular unit vector} (vector with length 1 in the direction of the displacement vector). In the upcoming sections, we will see how both components are incorporated in constructing the equivariant kernel for message-passing.

\begin{align}
    \|\mathbf{x}_{ij}\|&=\sqrt{x^2+y^2+z^2}\tag{radial distance}\\\hat{\mathbf{x}}_{ij}&=\frac{\mathbf{x}_{ij}}{\|\mathbf{x}_{ij}\|}\tag{angular unit vector}
\end{align}

Note that in most geometric graphs, edges are \textbf{bidirectional}, meaning there is an edge from node $i$ to node $j$ and an edge from node $j$ to node $i$. The displacement vector of bidirectional edges has the same radial distance and angular unit vectors pointing in opposite directions.

Another way of representing a point cloud is as a continuous function $\mathbf{f}$ that takes in a 3-dimensional vector representing a point in 3D space ($\mathbf{x}$) and outputs a feature vector ($\mathbf{f}$) if $\mathbf{x}$ is in the point cloud $(\mathbf{x} = \mathbf{x}_j)$ or the zero vector if $\mathbf{x}$ is not in the point cloud.

\begin{align}
    \mathbf{f}(\mathbf{x})=\sum_{j=1}^N\mathbf{f}_j \delta(\mathbf{x}-\mathbf{x}_j)\tag{point cloud}
\end{align}

Since point clouds are defined by the relative spatial displacements between nodes in a global (non-fixed) reference frame, they often represent objects with rotational symmetries that preserve equivariance. This means that rotating the entire point cloud should generate rotated system-level outputs, and rotating individual nodes should generate rotated node-level updates.

Representing graphs as continuous functions facilitates \textbf{continuous convolutions} that are applied to every point in space, also known as \textbf{point convolutions}. However, for the sake of clarity, we will consider graphs as a \textbf{finite set of points} on which the convolutions (which we will refer to as \textbf{kernels}) are applied. We do this by using summation notation, which will become clear in later sections.

\subsection{Wigner-D Matrices}
An \textbf{irreducible representation (or \textit{irrep})} of SO(3) defines how a type-$l$ spherical tensor transforms under 3D rotation. The type-$l$ irrep is a set of $(2l + 1) \times  (2l + 1)$ matrices called \textbf{Wigner-D matrices} that rotate a type-$l$ spherical tensor by an element $g \in  SO(3)$. For a specific 3D rotation $g \in SO(3)$, the Wigner-D matrices for type-$l$ tensors can be denoted as:

\begin{align}
    \mathbf{D}_l(g)\in \mathbb{R}^{(2l+1)\times(2l+1)}
\end{align}

\begin{figure}
\centering
\includegraphics[width=0.6\linewidth]{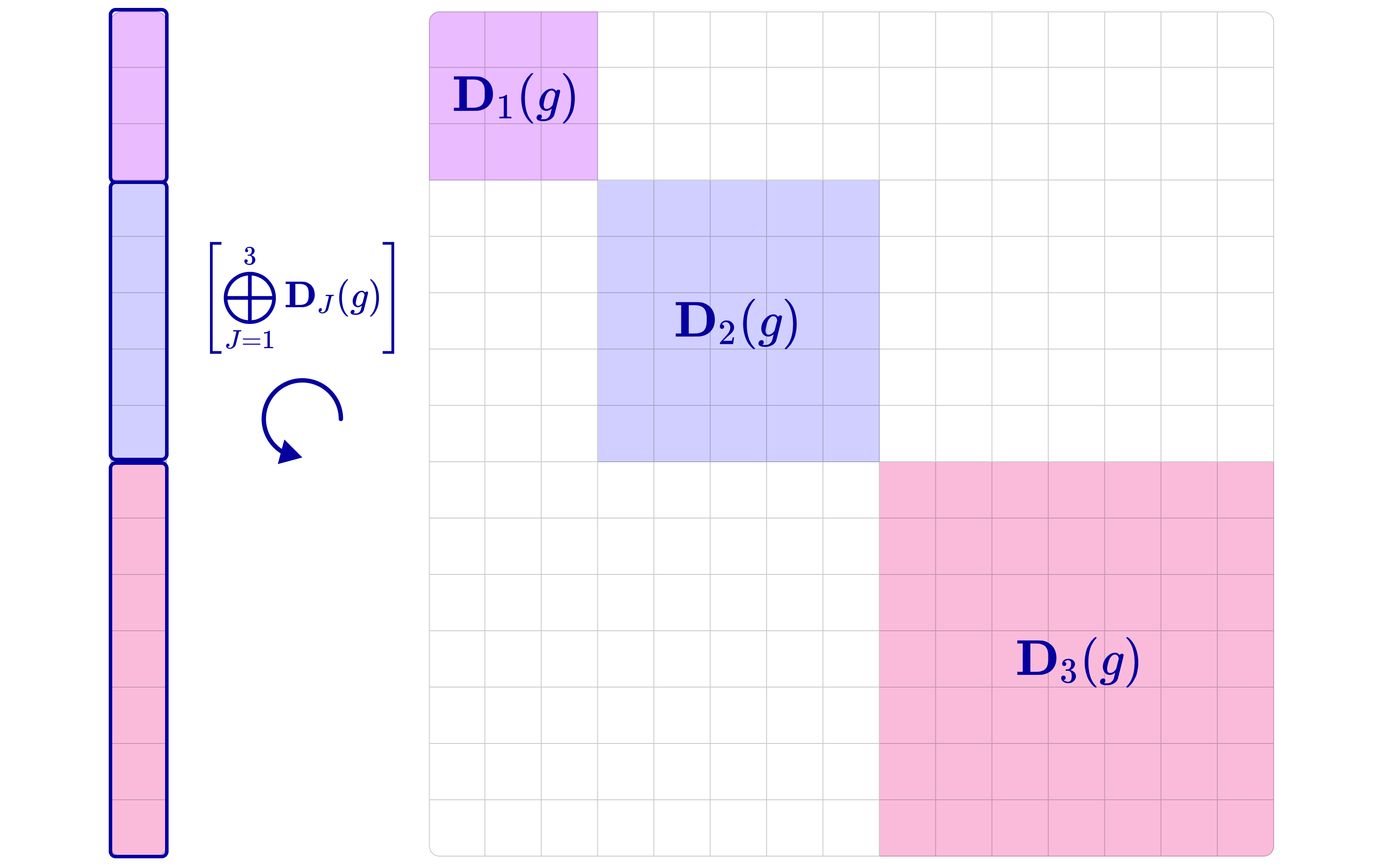}
\caption{The feature vector composed of stacked type-1, type-2, and type-3 spherical tensors rotates under the block diagonal matrix formed by the type-1, type-2, and type-3 Wigner-D matrices along the diagonal.}
\label{fig:rot}
\end{figure}

\begin{enumerate}
    \item \textbf{Type-0 tensors} are \textbf{scalars} and remain unchanged under 3D rotation. The Wigner-D matrix for a type-0 vector is a $1 \times 1$ matrix with a single entry of 1.
    \begin{align}
        \mathbf{D}_0(g)=[1]
    \end{align}

    \item \textbf{Type-1 tensors} are \textbf{3-dimensional vectors} and simply transform by the standard $3 \times 3$ rotation matrices under 3D rotation. All rotation matrices $\mathbf{R}$, and thus Wigner-D matrices for type-1 tensors, can be constructed by multiplying the basic rotation matrices that rotate by an angle $\alpha$ about the $x$-axis ($\mathbf{R}_x$), $\beta$ about the $y$-axis ($\mathbf{R}_y$), and $\gamma$ about the $z$-axis ($\mathbf{R}_z$).
    \begin{align}
        \mathbf{D}_1(g)=\mathbf{R}_x(\alpha)\mathbf{R}_y(\beta)\mathbf{R}_z(\gamma),
    \end{align}
    where:
    \begin{small}
    \begin{align}
        \underbrace{\mathbf{R}_x(\alpha)=\begin{bmatrix}1 & 0 & 0 \\
0 & \cos\alpha & -\sin\alpha \\
0 & \sin\alpha & \cos\alpha
\end{bmatrix}}_{\text{Rotation by $\alpha$ about $x$-axis}}, \underbrace{\mathbf{R}_y(\beta) = \begin{bmatrix}
\cos\beta & 0 & \sin\beta \\
0 & 1 & 0 \\-\sin\beta & 0 & \cos\beta \end{bmatrix}}_{\text{Rotation by $\beta$ about $y$-axis}}, \underbrace{\mathbf{R}_z(\gamma) = \begin{bmatrix}
\cos\gamma & -\sin\gamma & 0 \\
\sin\gamma & \cos\gamma & 0 \\
0 & 0 & 1
\end{bmatrix}}_{\text{Rotation by $\gamma$ about $z$-axis}}
    \end{align}
    \end{small}

    \item \textbf{Higher-order tensors} ($l > 2$) transform by the corresponding $(2l + 1) \times (2l + 1)$ type-$l$ Wigner-D matrix that we denote with a subscript $l$.
\end{enumerate}

Now, we can define the \textbf{equivariance condition using Wigner-D matrices}. A function (or kernel), which we will denote as $\mathbf{W}$, that transforms a type-$k$ spherical tensor to a type-$l$ spherical tensor, is equivariant if it satisfies the following equivariance condition:

\begin{align}
    \mathbf{D}_l(g)(\mathbf{W}\mathbf{f})=\mathbf{W}(\mathbf{D}_k(g)\mathbf{f})\tag{$\forall\mathbf{f}\in\mathcal{X}_k,g\in G$}
\end{align}

Since the feature vector is a stack of spherical tensors, it can be transformed via a matrix composed of concatenated Wigner-D matrices along the diagonal (Figure \ref{fig:rot}). In Section \ref{subsec:tensor-product} on tensor products, we walk through how to decompose any Cartesian tensor into its spherical tensor components, enabling them to transform directly and equivariantly under the Wigner-D matrices.

\subsection{Spherical Harmonics}
\purple[]{Spherical harmonics represent a \textbf{complete} and \textbf{orthonormal} basis for rotations in SO(3). They are functions that \textbf{project 3-dimensional vectors into spherical tensors} that transform equivariantly and directly under Wigner-D matrices, without requiring a change in basis. A spherical harmonic evaluated on a rotated 3-dimensional unit vector is \textbf{equal} to evaluating the spherical harmonic on the unrotated vector and transforming the output by an \textit{irrep} of SO(3). Vectors of spherical harmonic functions are used to \textbf{project the angular unit vector to spherical tensors}, which are a fundamental building block of equivariant kernels.}

\begin{figure}[h!]
    \centering
    \includegraphics[width=0.8\linewidth]{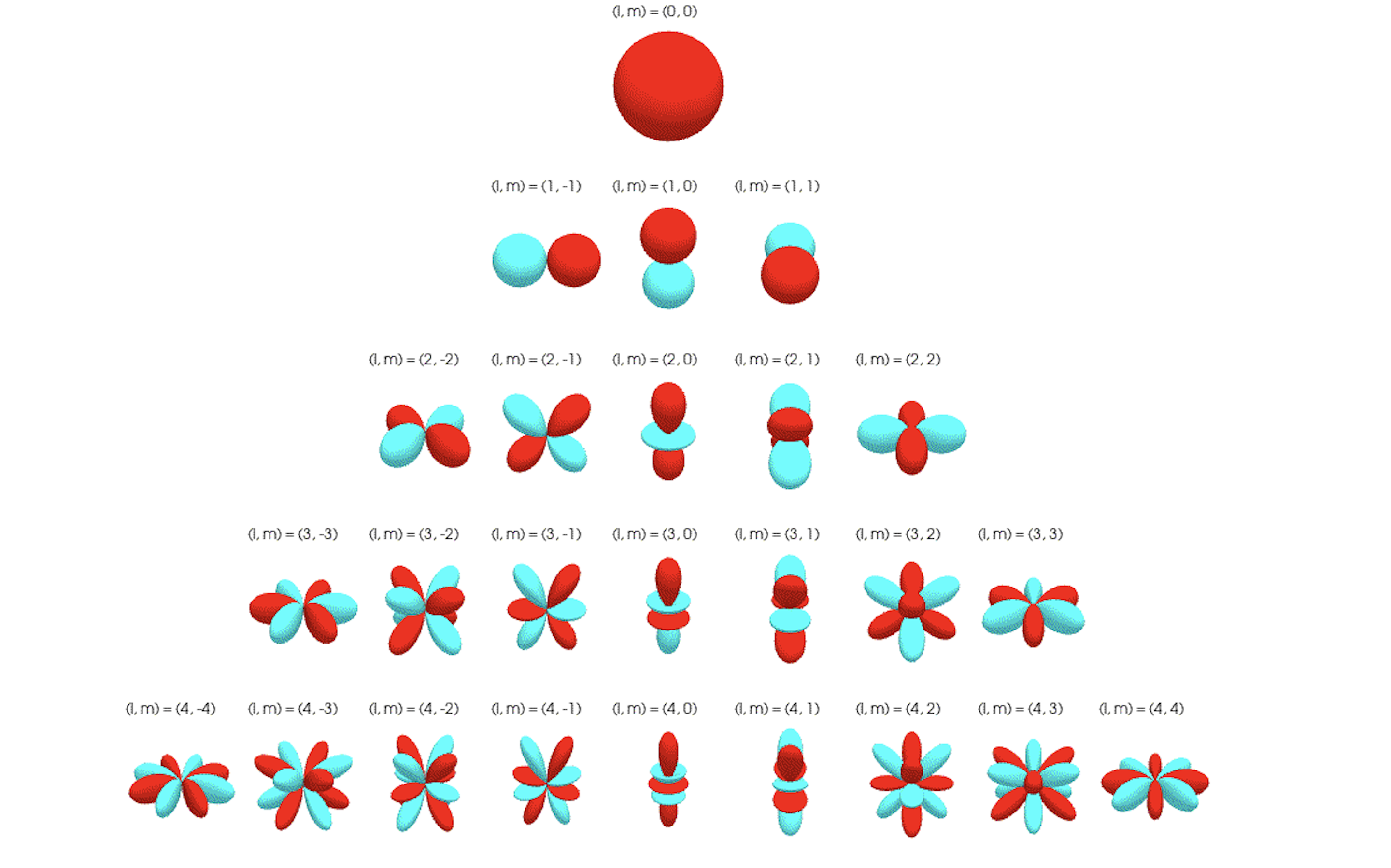}
    \caption{Real spherical harmonics for angular quantum numbers l = 0, 1, 2, 3, 4. For every point on the unit sphere, the spherical harmonic function outputs a real number (represented by the color and distance from the origin). The farther the distance from the origin, the larger the magnitude of the spherical harmonic, and the color indicates whether the harmonic is positive (red) or negative (blue). (Taken from \href{https://en.wikipedia.org/wiki/Table_of_spherical_harmonics}{Wikipedia})}
    \label{fig:spherical}
\end{figure}

The orthonormal basis functions with which all SO(3)-equivariant spherical tensors can be constructed are called \textbf{spherical harmonics}. We can also consider spherical harmonics as sets of functions that project 3-dimensional vectors onto \textbf{orthogonal tensor subspaces where Wigner-D matrices operate}. Each spherical harmonic function (indexed by its degree $l$ and order $m$) takes a unit vector (length of 1) on the unit sphere ($S^2$) and returns a real number.

\begin{align}
    Y^{(l)}_{m}(\hat{\mathbf{x}}):S^2\to \mathbb{R}
\end{align}

For every type of spherical tensor $l$, there is a corresponding vector of $2l + 1$ spherical harmonic functions indexed by $m$ that transform points on the unit sphere into type-$l$ spherical tensors that rotate directly under type-$l$ Wigner-D matrices.

\begin{align}
    \mathbf{Y}^{(l)}(\hat{\mathbf{x}})=\begin{bmatrix} 
Y^{(l)}_{-l}(\hat{\mathbf{x}})
\\
\\
\vdots\\ 
\\Y^{(l)}_{l-1}(\hat{\mathbf{x}})\\\\
Y^{(l)}_{l}(\hat{\mathbf{x}})\end{bmatrix}
\end{align}

This spherical harmonic vector transforms a point on the unit sphere to a type-$l$ spherical tensor:

\begin{align}
    \mathbf{Y}^{(l)}(\hat{\mathbf{x}}):S^2\to \mathbb{R}^{2l+1}
\end{align}

The explicit expressions defining the real spherical harmonics \footnote{Real spherical harmonics are often used in machine learning applications to reduce computational complexity; however, complex spherical harmonics are used in quantum mechanics.} (or tesseral spherical harmonics) can be written in terms of the \textbf{angle from the $z$-axis} (polar angle $\theta$) and the \textbf{angle from the $x$-axis} of the orthogonal projection onto the $xy$-plane (azimuthal angle $\phi$):

\begin{align}
    Y^{(l)}_m(\theta,\phi)=\underbrace{\sqrt{\frac{(2l+1)}{4\pi}\frac{(l-m)!}{(l+m)!}}}_{\text{normalization constant}}\underbrace{P_l^{|m|}(\cos\theta)}_{\text{polar dependence}}\cdot \underbrace{\begin{cases}\sin(|m|\phi)&m<0\\1&m=0\\\cos(m\phi)&m>0\end{cases}}_{\text{azimuthal dependence}}
\end{align}

\begin{figure}[h!]
    \centering
    \includegraphics[width=0.3\linewidth]{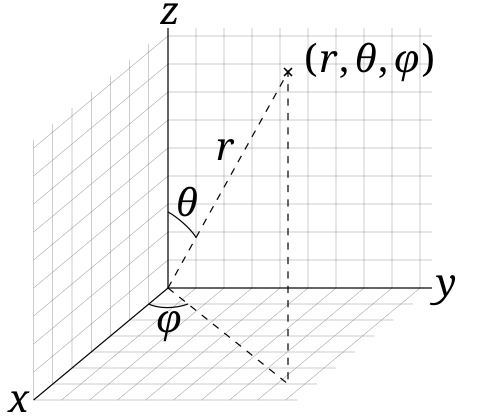}
    \caption{Diagram showing the polar ($\theta$) and azimuthal ($\phi$) angles that are used to calculate the spherical harmonics. (Taken from \href{https://en.wikipedia.org/wiki/Spherical_coordinate_system}{Wikipedia})}
    \label{fig:spherical}
\end{figure}

The function dependent on $\cos(\theta)$ is the \textbf{associated Legendre polynomial (ALP)} with degree $l$ and order m given by the following equation:

\begin{align}
    P^{|m|}_l(x)=(-1)^m(1-x^2)^{\frac{m}{2}}\underbrace{\frac{d^m}{dx^m}(\underbrace{P_l(x)}_{\text{degree-}l \text{ Legendre polynomial}})}_{m\text{th derivative}}
\end{align}

In the equation above, $P_l$ denotes the \textbf{Legendre polynomial} with maximum degree $l$. Visually, you can think of the Legendre polynomial where $x=\cos(\theta)$ as a wave on the unit sphere. The zeros of the polynomial between $[-1, 1]$ are \textbf{nodes} on the spherical harmonic where the harmonic changes sign. The set of Legendre polynomials is \textbf{mutually orthogonal over the interval $[-1, 1]$}, meaning that the inner product of any two polynomials in the set is 0. 

\begin{align}
    \langle P_l(x), P_k(x)\rangle=\int_{x=-1}^1P_l(x)P_k(x)dx=0\tag{$l\neq k$}
\end{align}

The Legendre polynomials are also complete, such that all \textbf{square-integrable} functions on the interval $[-1, 1]$ can be approximated as a linear combination of the Legendre polynomials, where the coefficients are independent of one another\footnote{A function is square-integrable on $[-1, 1]$ if the integral (area under the curve) of the function squared on the domain $[-1, 1]$ is less than infinity (bounded). This is an important property of wavefunctions in quantum mechanics because the square of a wavefunction represents a probability distribution with an integral of 1.}.

When $x=\cos(\theta)$, the $(1-x^2)$ term becomes $\sin^2\theta$, and we can \textbf{rewrite the ALP as}:

\begin{align}
    P^{|m|}_l(\cos\theta)=(-1)^m(\sin\theta)^{m}\frac{d^m}{d(\cos\theta)^m}(P_l(\cos\theta))
\end{align}

The \textbf{associated Legendre polynomials (ALPs)} are obtained by taking the mth derivative of the Legendre polynomials and scaling by a factor of $(\sin\theta)^m$. Since the $(\sin\theta)^m$ factor is zero for $\theta = 0$ (north pole) and $\theta = \pi$ (south pole) for all non-zero m, there is a node at the north and south poles. As the exponent m increases, the \textbf{ALPs start to approach zero farther from the poles, and the peak between 0 and $\pi$ becomes narrower}, effectively decreasing the number of possible nodes between $\theta = 0$ and $\pi$ generated from the Legendre polynomial term. These properties are reflected in the graph below of all the ALPs of degree $l=5$ with non-negative orders of $m$.

\begin{figure}[h!]
    \centering
    \includegraphics[width=0.4\linewidth]{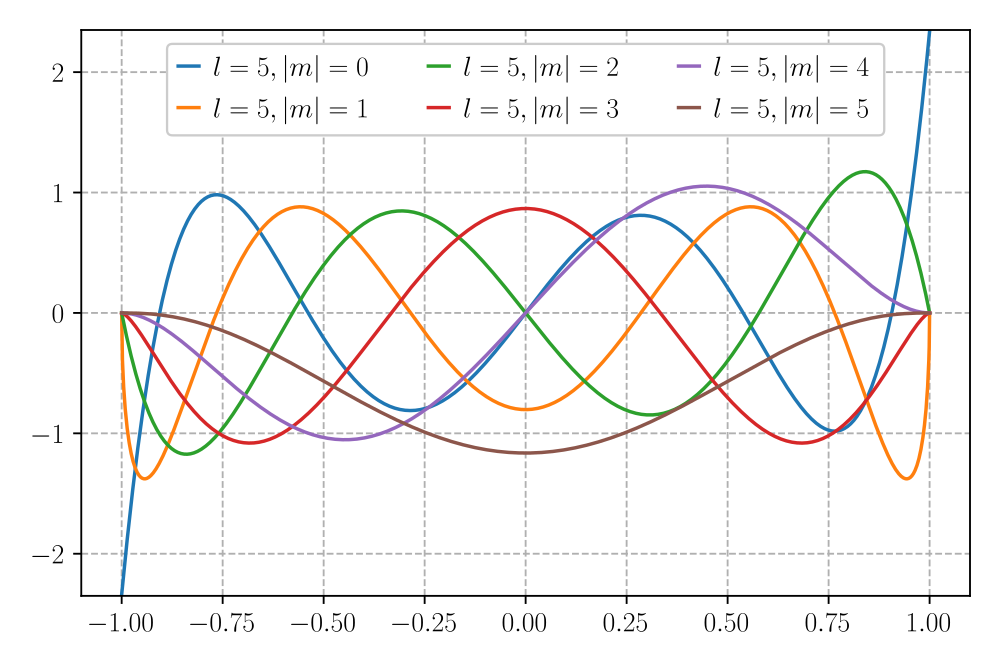}
    \caption{The associated Legendre polynomials for $l =5$ and all non-negative values of $m$ on the interval $[-1,1]$. (Taken from \href{https://en.wikipedia.org/wiki/Legendre_function}{Wikipedia})}
    \label{fig:spherical}
\end{figure}

The purpose of including the ALP in spherical harmonics as opposed to the unmodified Legendre polynomials is to \textbf{allow dependence on the azimuthal angle} ($\phi$). Since the azimuthal angle is undefined at the poles, the $(\sin\theta)^m$ factor ensures that there is no contribution from the azimuthal term at the poles for all non-zero values of $m$. 

\begin{align}
    Y^{(l)}_m(\theta,\phi)=\underbrace{\sqrt{\frac{(2l+1)}{4\pi}\frac{(l-m)!}{(l+m)!}}}_{\text{normalization constant}}\underbrace{P_l^{|m|}(\cos\theta)}_{\text{polar dependence}}\cdot \underbrace{\begin{cases}\sin(|m|\phi)&m<0\\1&m=0\\\cos(m\phi)&m>0\end{cases}}_{\text{azimuthal dependence}}
\end{align}

When $m = 0$, the $(\sin\theta)^m$ factor is 1, and the polynomial is unbounded at the poles, and there is no azimuthal dependence in the spherical harmonic.

\begin{figure}[h!]
\centering
\includegraphics[width=\linewidth]{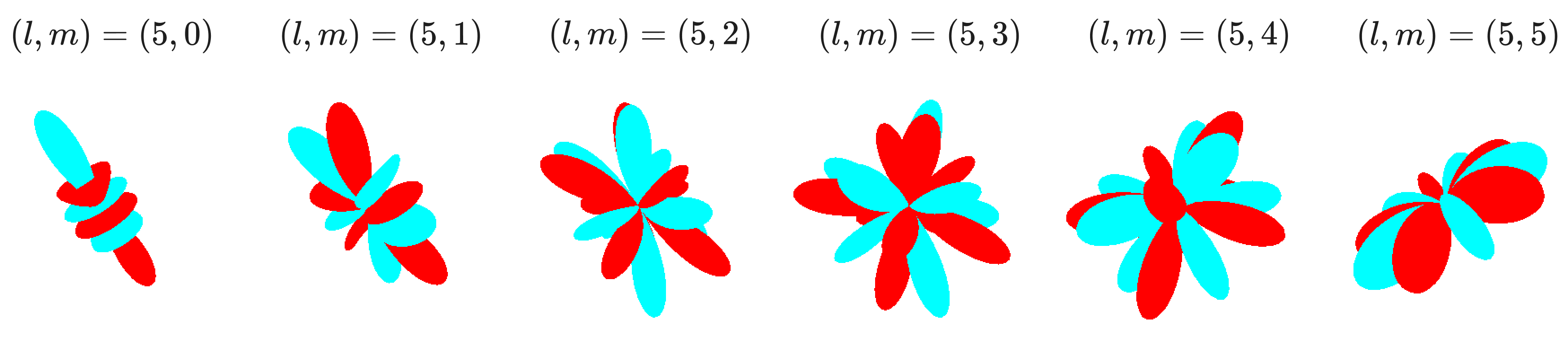}
\caption{Spherical harmonics corresponding to the associated Legendre polynomials for l=5 and all non-negative values of m. The nodes at the poles introduce azimuthal dependence as shown by the changes in sign around the z-axis. As m increases, the number of peaks between the poles decreases from taking higher derivatives of the sine factor. (Adapted from \href{https://en.wikipedia.org/wiki/Table_of_spherical_harmonics}{Wikipedia})}
\label{fig:key}
\end{figure}

Including the sine or cosine function, dependent on the azimuthal angle, allows spherical harmonics to describe a larger range of functions on the sphere and orientations of the angular momentum vector in a magnetic field.

It is also worth noting that taking the \textbf{$l$th derivative} of a degree-$l$ Legendre polynomial is a \textbf{constant}, and taking even higher order derivatives returns zero. This means that including the ALP term aligns with the condition that the magnitude of the angular momentum projected on the $z$-axis $|m|$ must be less than the magnitude of the angular momentum vector $|m| \leq l$.

Spherical harmonics inherit the properties of the Legendre polynomials and ALPs applied to the unit sphere, forming a \textbf{complete} and \textbf{orthogonal} basis for spherical tensors with the following properties:

\begin{enumerate}
    \item The \textbf{completeness} of spherical harmonics means that all Cartesian tensors can be represented as a set of orthogonal projections on the spherical tensor subspaces that rotate directly under Wigner-D matrices via a change of basis.
    \item The \textbf{orthogonality} of spherical harmonics means that each type of spherical tensor transforms independently under rotation by the type-$l$ Wigner-D matrix. This means that the \textbf{direct sum} (or vector concatenation) of spherical tensors rotates under a block-diagonal matrix with Wigner-D matrices along the diagonal and zeros everywhere else. The zero entries indicate that there are no cross dependencies between the tensors under rotation, a crucial property for our upcoming discussion on constructing equivariant layers.
    \begin{align}
        \int_{S^2}Y^{(l)}_{m_l}(\theta,\phi)Y^{(k)}_{m_k}(\theta,\phi)d\Omega=\delta_{l k}\delta_{m_l m_k}
    \end{align}
    
    The \textbf{integral of the product of spherical harmonics over the unit sphere} (inner product) is equal to the product of two Kronecker deltas $\delta$, which is equal to 1 if $l = k$ or $m_l = m_k$, and zero otherwise. This means that the inner product of two real spherical harmonics is \textbf{non-zero only when they share the same degree $l$ and order $m$, and is zero for all distinct pairs of $l$ and $m$}.  
\end{enumerate}

To gain some physical intuition, we will describe the role of spherical harmonics in quantum mechanics, where they are used to describe the angular component of electron wavefunctions.

\purple[Spherical Harmonics in Quantum Mechanics]{
Spherical harmonics are \textbf{eigenfunctions} of the \textbf{orbital angular momentum operators} $L^2$ and $L_z$ that act on electron wavefunctions. When the squared total angular momentum operator ($L^2$) is applied to a spherical harmonic, it returns the function scaled by the eigenvalue (scalar) $\hbar^2l(l+1)$ consisting of the total angular momentum number $l$, where $\hbar$ is the Planck constant.

\begin{align}
    \hat{L}^2Y_m^{(l)}(\theta,\phi)=l(l+1)\hbar^2Y_m^{(l)}(\theta,\phi)\tag{$l=0,1,\dots$}
\end{align}

In addition, when the $z$-component angular momentum operator ($L_z$) is applied to a spherical harmonic, it returns the function scaled by the eigenvalue $m\hbar$ containing the magnetic quantum number $m$.

\begin{align}
    \hat{L}_zY^{(l)}_m(\theta,\phi)=m\hbar Y^{(l)}_m(\theta,\phi)\tag{$m=-l,\dots,l$}
\end{align}

These eigenvalues give the quantized values of orbital angular momentum, that is, the angular momentum of an electron orbital cannot take any value but is limited to certain quantities defined by the integer values of $l$ and $m$.

The orbitals with defined angular momentum $|l,m\rangle$ are called \textbf{eigenstates}, and they have wavefunctions that map every point in space to a probability amplitude. The square of the wavefunction describes the probability of an electron existing in a specific quantum state. Wavefunctions can describe the probability distribution of position, angular momentum, spin, or energy of an electron. The spherical harmonic corresponding to $|l,m\rangle$ is the \textbf{angular component of the wavefunction of an isolated electron, which describes how the probability densities vary with orientation around the origin}.

This illustrates how spherical harmonics capture the foundational rotational symmetries of physical systems, which make them the perfect basis for constructing functions on the sphere that detect rotationally symmetric patterns in graph features. }

\begin{wrapfigure}{r}{0.4\linewidth}
    \centering
    \includegraphics[width=\linewidth]{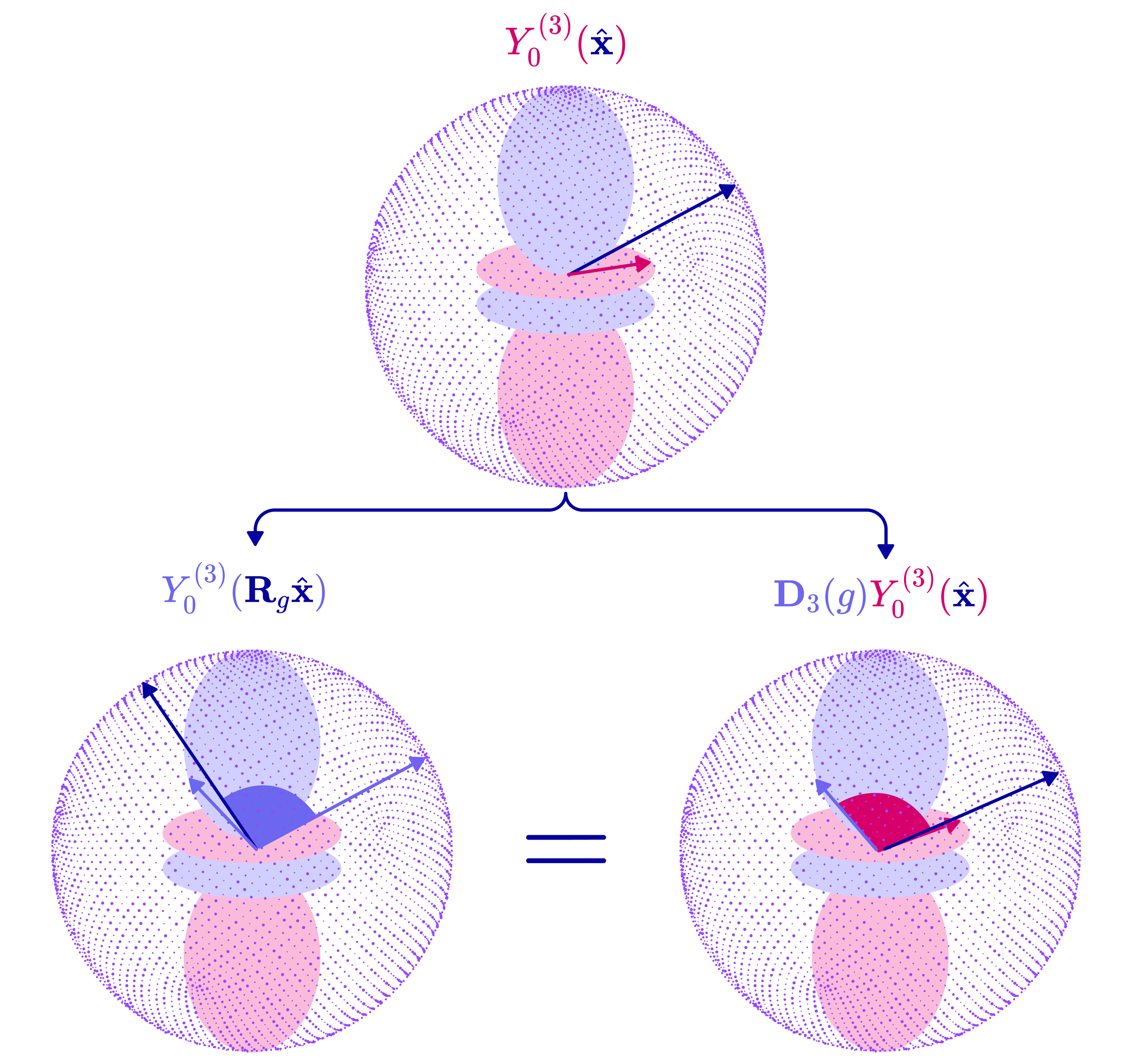}
    \caption{Spherical harmonic functions are equivariant under rotation in SO(3). Rotating the 3D vector input to the spherical harmonic corresponding to $l=3$ and $m=0$ gives the same output as rotating the output of the spherical harmonic evaluated on the unrotated vector by the corresponding Wigner-D matrix.}
    \label{fig:spherical2}
\end{wrapfigure}

Critically, the spherical harmonics are \textbf{equivariant functions}, meaning that a rotation of the input 3-dimensional vector by the $3 \times 3$ rotation matrix $\mathbf{R}$ for $g \in SO(3)$ is equivalent to rotating the spherical harmonic projection by the type-$l$ Wigner-D matrix for $g$.

\begin{align}
    \mathbf{Y}^{(l)}(\mathbf{R}_g\hat{\mathbf{x}})=\mathbf{D}_l(g)\mathbf{Y}^{(l)}(\hat{\mathbf{x}})\tag{$\forall g\in SO(3)$}
\end{align}

The type-$J$ vectors of spherical harmonic functions with elements $m = -J$ to $J$ are used to \textbf{project angular unit vectors of edges into higher-degree spherical tensors that can model different frequencies of rotationally symmetric features}. These higher-degree spherical tensors form the \textbf{basis set of SO(3)-equivariant kernels} that can combine to capture complex rotationally symmetric chemical properties with high precision, which we will discuss in depth throughout the guide.

\begin{figure}[h!]
    \centering
    \includegraphics[width=0.7\linewidth]{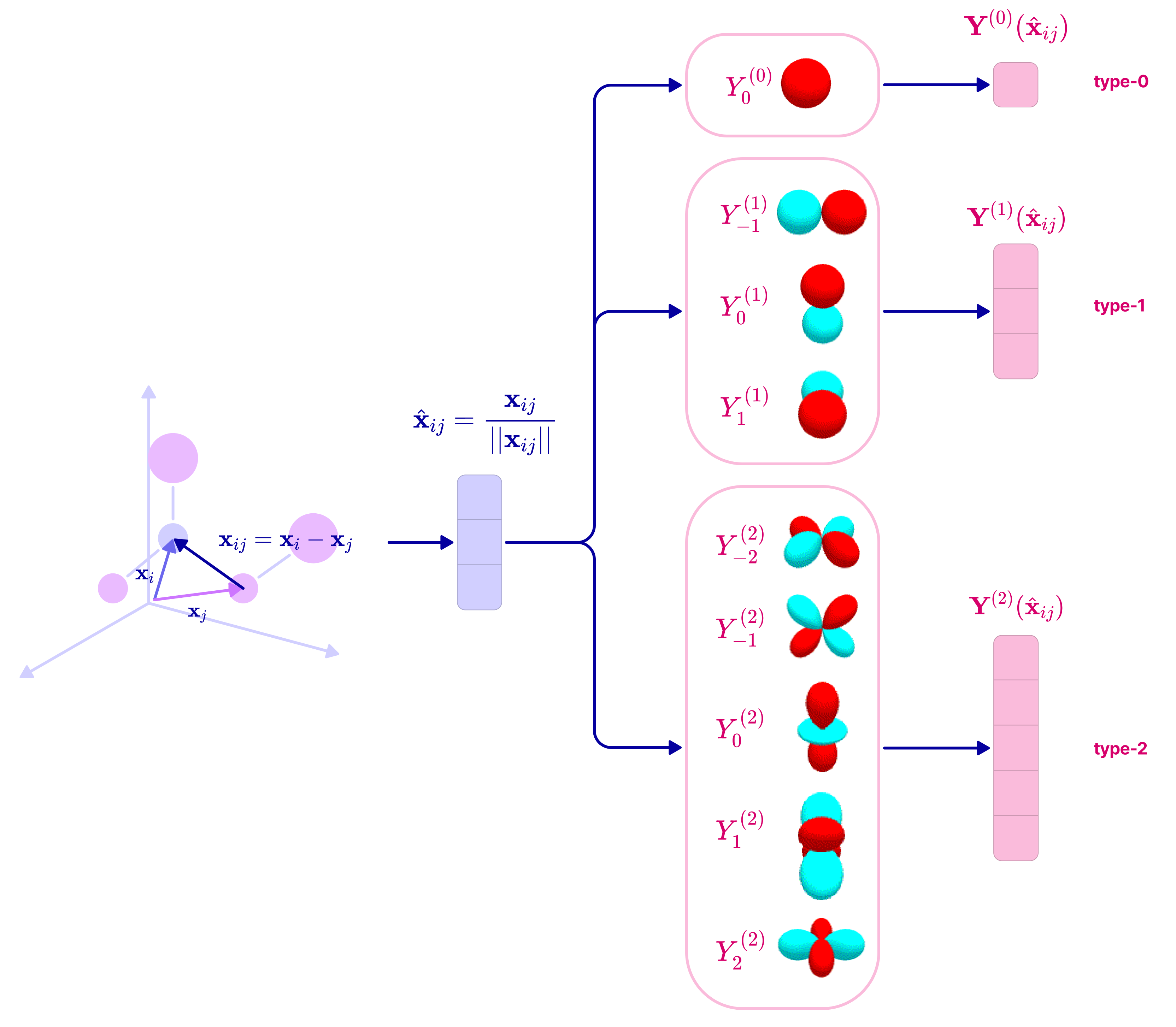}
    \caption{Diagram showing how the displacement vector between two nodes is projected to type-0, type-1, and type-2 spherical tensors via spherical harmonic functions.}
    \label{fig:proj}
\end{figure}

Like how the Fourier series decomposes periodic signals into sine and cosine components with specific periodic frequencies, spherical harmonics \textbf{decompose rotationally symmetric, or SO(3)-equivariant, features on the unit sphere into components that change with specific angular frequencies}. This decomposition is crucial to model how SO(3)-equivariant features vary across positions and orientations on the unit sphere.

The \textbf{degree} or \textbf{type} of a spherical harmonic determines its frequency or how rapidly it oscillates on the sphere. Lower-degree harmonics can model features with broader, smoother variations under rotations, while higher-degree harmonics capture features with sharper, finer variations under rotation \citep{cen2024high}.

Just as low-frequency sinusoids fail to accurately approximate high-frequency functions in Fourier analysis, low-degree spherical harmonics are not sensitive enough to handle chemical properties that vary dramatically with subtle changes in atomic orientation and position. In Section \ref{subsec:basis-kernels} on computing the basis kernels, we will deconstruct how to precompute the spherical harmonic functions using recursive relations. 

\textbf{Now that we have defined the basis for spherical tensors, we will discuss how to combine and convert between tensors of different types using the tensor product.}

\subsection{Tensor Product}
\label{subsec:tensor-product}
\purple[]{
The tensor product is a \textbf{bilinear} and \textbf{equivariant} operation that combines two spherical tensors to produce a higher-dimensional tensor. Since the output higher-dimensional tensors are generally not spherical tensors themselves, we must decompose them into their spherical tensor components using change-of-basis matrices formed with Clebsch-Gordan coefficients.
}

Suppose we want to exchange information between type-$k$ and type-$l$ features. Since they are different types, they are transformed differently under 3D rotation. \textit{How do we exchange information between these features without breaking equivariance?}

This is where the \textbf{tensor product} comes in, which is denoted by $\otimes$.

The tensor product converts the type-$k$ and type-$l$ spherical tensors into a $(2l+1) \times (2k+1)$ matrix by calculating the product of every pair of dimensions indexed by $m_k$ and $m_l$.

\begin{align}
    \mathbf{s}^{(k)}\otimes \mathbf{t}^{(l)}=\mathbf{s}^{(k)}\mathbf{t}^{(l)\top}=\begin{bmatrix}s^{(k)}_{-m_k}t^{(l)}_{-m_l}&s^{(k)}_{-m_k}t^{(l)}_{-m_l+1}&\dots&s^{(k)}_{-m_k}t^{(l)}_{m_l}\\s^{(k)}_{-m_k+1}t^{(l)}_{-m_l}&s^{(k)}_{-m_k+1}t^{(l)}_{-m_l+1}&\dots&\vdots\\\vdots & \vdots & \ddots &\vdots \\s^{(k)}_{m_k}t^{(l)}_{-m_l}&s^{(k)}_{m_k}t^{(l)}_{-m_l+1}&\dots&s^{(k)}_{m_k}t^{(l)}_{m_l}\end{bmatrix}
\end{align}

We can flatten this matrix into a $(2l+1)(2k+1)$-dimensional tensor. However, this higher-dimensional tensor is not spherical, and \textbf{we must define the representation $\mathbf{D}(g)$ under which it rotates equivariantly}. Since the tensor product is equivariant, it satisfies the equivariance condition, which states that rotating the tensor product by $g$ is equivalent to rotating the individual spherical tensors by their respective Wigner-D matrices and then taking the tensor product. This translates into the following equation:

\begin{align}
    \mathbf{D}(g)(\mathbf{s}^{(k)}\otimes \mathbf{t}^{(l)})=(\mathbf{D}_k(g)\mathbf{s}^{(k)})\otimes (\mathbf{D}_l(g)\mathbf{t}^{(l)})
\end{align}

Using the tensor product identity below:

\begin{align}
    \text{vec}(\mathbf{AXB})=(\mathbf{B}^{\top}\otimes \mathbf{A})\text{vec}(\mathbf{X})
\end{align}

We can manipulate the above equation to isolate the Wigner-D matrices:

\begin{align}
    \text{vec}\left((\mathbf{D}_k(g)\mathbf{s}^{(k)})\otimes (\mathbf{D}_l(g)\mathbf{t}^{(l)})\right)&=\text{vec}\left((\mathbf{D}_k(g)\mathbf{s}^{(k)})(\mathbf{D}_l(g)\mathbf{t}^{(l)})^{\top}\right)\\
    &=\text{vec}\left(\mathbf{D}_k(g)\mathbf{s}^{(k)}\mathbf{t}^{(l)\top}\mathbf{D}_l(g)^{\top}\right)\\
    &=\left(\mathbf{D}_k(g)\otimes\mathbf{D}_l(g)\right)\text{vec}(\mathbf{s}^{(k)}\otimes\mathbf{t}^{(l)})
\end{align}

This means that the tensor product of a type-$k$ and a type-$l$ tensor rotates under the representation of SO(3) derived from the \textbf{Kronecker product} ($\otimes$) of the type-$k$ and type-$l$ Wigner-D matrices:

\begin{align}
    \mathbf{D}_{k\otimes l}(g)=\mathbf{D}_k(g)\otimes\mathbf{D}_l(g)
\end{align}

The \textbf{Kronecker product} operation is analogous to the tensor product for matrices that produces a ‘\textit{matrix of matrices}’ where \textbf{each block of the outer matrix is an inner matrix derived from scaling the second matrix in the product by the corresponding element of the first matrix}.

The Kronecker product of two SO(3) representations is another representation; but since every group representation can be written as the \textbf{direct sum} of \textit{irreps} (where each block along the diagonal is an \textit{irrep} and the remaining entries are zero) coupled with an orthogonal change-of-basis matrix and its transpose, we can decompose the Kronecker product of the type-$k$ and type-$l$ Wigner-D matrices into the \textbf{direct sum of Wigner-D matrices} coupled with the change of basis matrix composed of a special set of coefficients called the \textbf{Clebsch-Gordan coefficients} (which we will dive into in Section \ref{subsec:cb-decomp}). 

\begin{align}
    \mathbf{D}_k(g)\otimes\mathbf{D}_l(g)&=\mathbf{Q}^{l k\top}\left[\bigoplus_{J=|k-l|}^{k+l}\mathbf{D}_J(g)\right]\mathbf{Q}^{l k}\\
    &=\mathbf{Q}^{l k\top}\begin{bmatrix}\mathbf{D}_{|k-l|}(g)\\\\&&\ddots\\\\&&&&\mathbf{D}_{k+l}(g)\end{bmatrix}\mathbf{Q}^{l k}
\end{align}

where $\mathbf{Q}$ is a $(2l+1)(2k+1) \times (2l+1)(2k+1)$ orthogonal change-of-basis matrix where each element is a Clebsch-Gordan coefficient.

From this equation, we see that the Clebsch-Gordan change of basis matrix can be used to \textbf{transform the $(2l+1)(2k+1)$-dimensional tensor product into the direct sum of exactly one spherical tensor of each type ranging from $|k-l|$ to $k+l$} stacked into a single vector that rotates under the direct sum of Wigner-D matrices ranging from $|k-l|$ to $k+l$. We can think of the change-of-basis operation as projecting the tensor from the combined space into several orthogonal subspaces that rotate under defined representations of SO(3).

\begin{align}
    \mathbf{s}^{(k)}\otimes \mathbf{t}^{(l)}\in k\otimes l=|k-l|\oplus|k-l+1|\oplus...\oplus(k+l-1)\oplus(k+l)
\end{align}

\begin{figure}[h!]
    \centering
    \includegraphics[width=\linewidth]{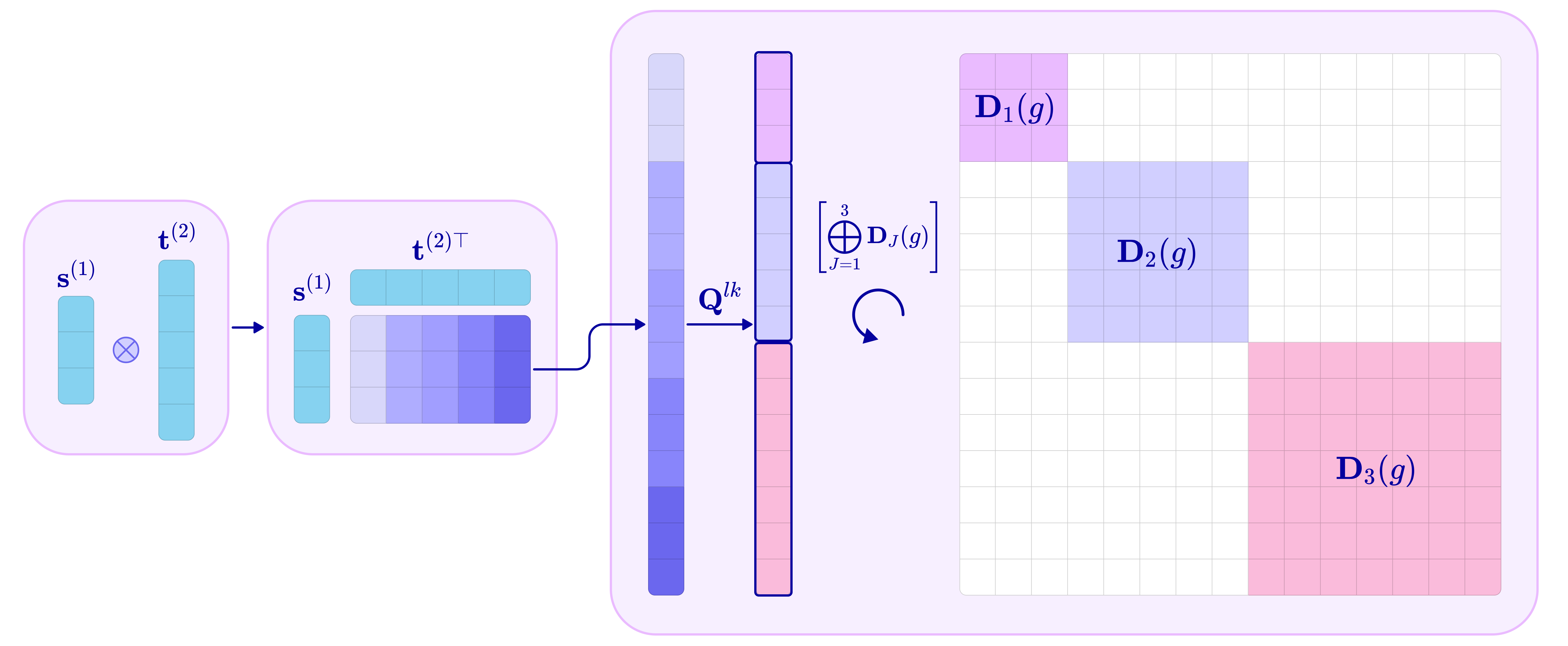}
    \caption{Diagram showing the tensor product of the type-1 spherical tensor $\mathbf{s}$ and the type-2 spherical tensor $\mathbf{t}$, which produces a 15-dimensional tensor. This tensor is decomposed into the direct sum of exactly one type-1 ($|2-1|$), type-2, and type-3 $(2+1)$ spherical tensor, which rotate independently under the direct sum of the type-1, type-2, and type-3 Wigner-D matrices.}
    \label{fig:tensor-product}
\end{figure}

Let's solidify this abstract idea with a familiar example. Consider the tensor product of two type-1 tensors (3-dimensional vectors) $\mathbf{a}$ and $\mathbf{b}$, which gives a $3 \times 3$ matrix (or 9-dimensional tensor):

\begin{align}
    \mathbf{a}\mathbf{b}^{\top}=\begin{bmatrix}a_x\\a_y\\a_z\end{bmatrix}\begin{bmatrix}b_x&b_y&b_z\end{bmatrix}=\begin{bmatrix}a_xb_x&a_xb_y&a_xb_z\\a_yb_x&a_yb_y&a_yb_z\\a_zb_x&a_zb_y&a_zb_z\end{bmatrix}
\end{align}

\begin{figure}
    \centering
    \includegraphics[width=\linewidth]{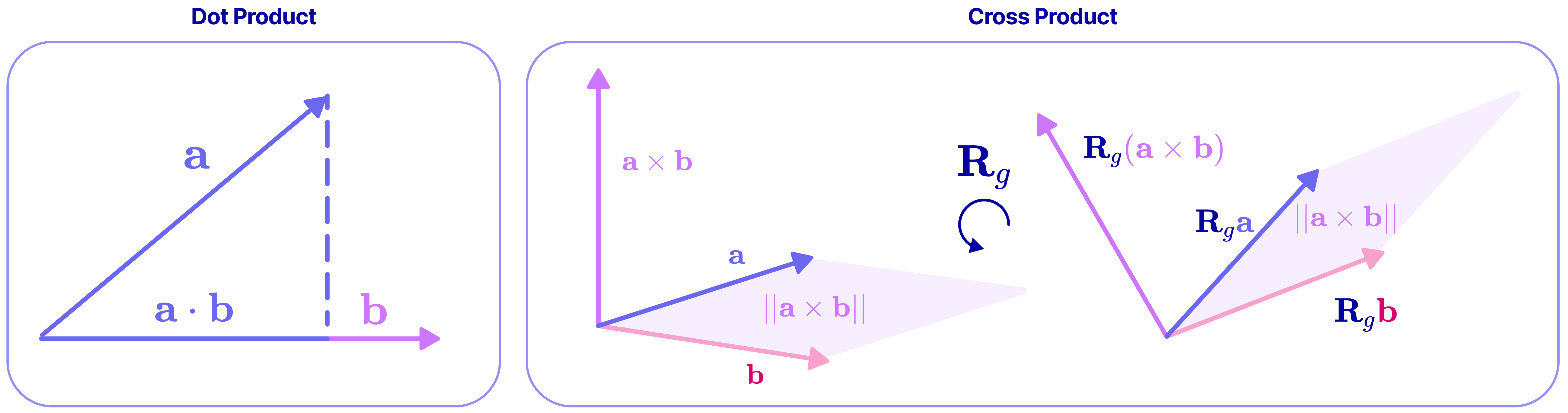}
    \caption{\textbf{(Left)} The dot product of two 3-dimensional vectors $\mathbf{a}$ and $\mathbf{b}$ is the length of the projection of a onto the line spanned by $\mathbf{b}$. The dot product is invariant to rotation. \textbf{(Right)} The cross product between two 3-dimensional vectors a and b is the vector perpendicular to both $\mathbf{a}$ and $\mathbf{b}$ (direction obtained from the right-hand rule) with a magnitude equal to the area of the parallelogram formed by the two vectors. The cross product is equivariant under rotation by the type-1 Wigner-D matrices.}
    \label{fig:dot-cross}
\end{figure}
We can extract some familiar values from this matrix:

\begin{enumerate}
    \item We can see that the trace of the matrix (sum of values along the diagonal) is equal to the \textbf{dot product} of $\mathbf{a}$ and $\mathbf{b}$.
    \begin{align}
        \mathbf{a}\cdot \mathbf{b}=a_xb_x+a_yb_y+a_zb_z
    \end{align}    
    The dot product can be interpreted as the length of the projection of the vector $\mathbf{a}$ on the line created by vector $\mathbf{b}$. If we rotate both a and b, the length of the projection shouldn’t change because the lengths of the individual vectors and the angle between them remain constant under rotation. So we can think of the trace as the type-0 spherical tensor component of the 9-dimensional Cartesian tensor that transforms invariantly under rotation.
    \item We can also extract the cross-product from this $3 \times 3$ matrix from the antisymmetric elements:
    \begin{align}
        \mathbf{a}\times \mathbf{b}=\begin{bmatrix}a_yb_z-a_zb_y\\a_zb_x-a_xb_z\\a_xb_y-a_yb_x\end{bmatrix}
    \end{align}
    The cross-product can be interpreted as the vector perpendicular to both vectors $\mathbf{a}$ and $\mathbf{b}$ with length equivalent to the area of the parallelogram formed by the two vectors. If we rotate a and b by the rotation matrix $\mathbf{R}$,, the cross-product should rotate by the same matrix $\mathbf{R}$, but the length would remain constant since rotations preserve lengths and angles. Thus, we can think of the cross product as the type-1 spherical tensor component of the 9-dimensional Cartesian tensor that transforms under the type-1 Wigner-D matrices (standard $3 \times 3$ rotation matrices).

    \item Unfortunately, the type-2 spherical tensor component of the $3 \times 3$ matrix does not have a concrete physical interpretation. However, we can think of it as the traceless, symmetric part of the matrix that rotates under the type-2 Wigner-D matrices.

    \begin{align}
        \begin{bmatrix}c(a_xb_z+a_zb_x)\\c(a_xb_y+a_yb_x)\\2a_yb_y-a_xb_x-a_zb_z\\c(a_yb_z+a_zb_y)\\c(a_zb_z+a_xb_x)\end{bmatrix}
    \end{align}
\end{enumerate}

The decomposed 9-dimensional Cartesian tensor is the direct sum of the \textbf{type-0} (trace), \textbf{type-1} (asymmetric), and \textbf{type-2} (traceless, symmetric) spherical tensor components. 

\begin{figure}[h!]
\centering
\includegraphics[width=0.9\linewidth]{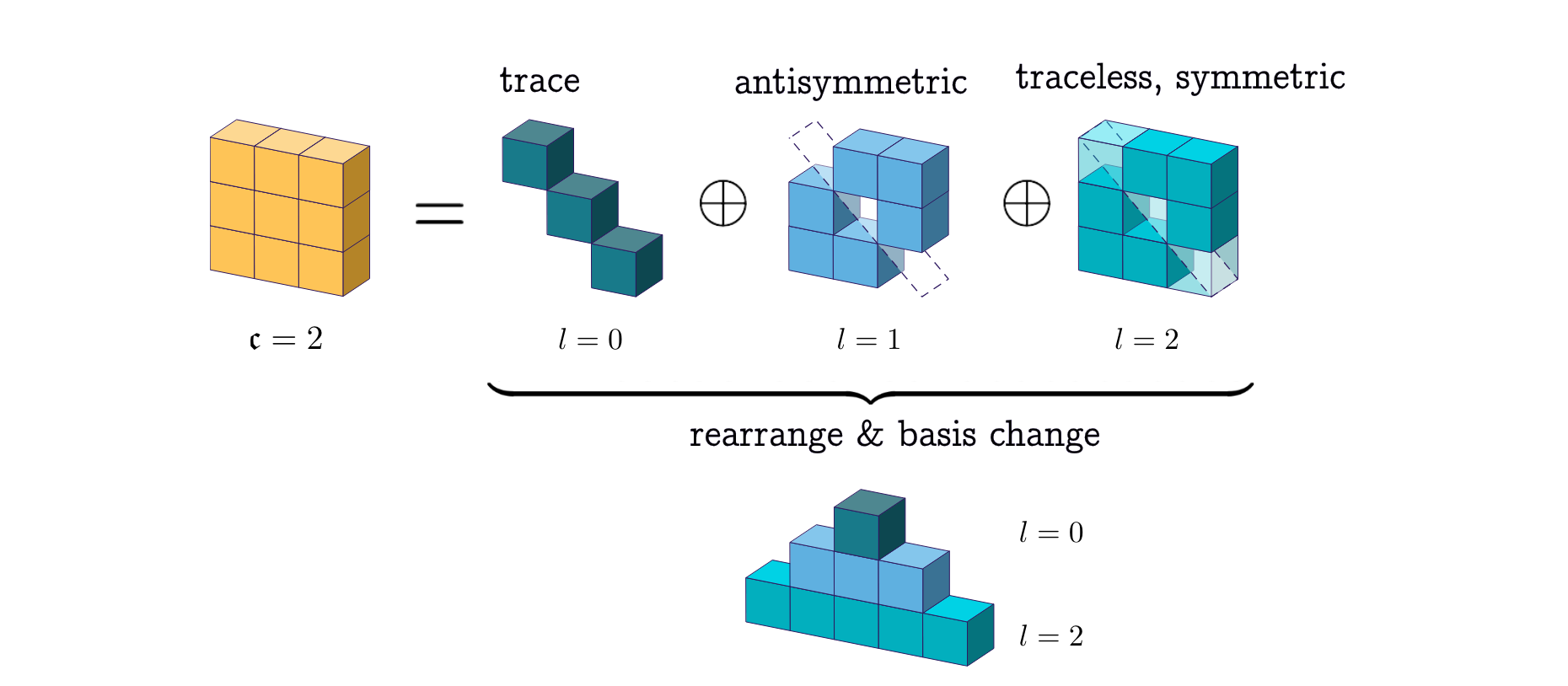}
\caption{Decomposition of a 9-dimensional Cartesian tensor or 3 x 3 matrix into its type-0, type-1, and type-2 spherical tensor components. (Source: \citep{duval2023hitchhiker})}
\label{fig:decomp}
\end{figure}

If we stack all three representations into a $(1+3+5)$-dimensional tensor, the resulting vector concatenation \textbf{rotates under the direct sum of the type-0, type-1, and type-2 Wigner-D matrices}: 

\begin{align}
\begin{bmatrix}1&&\\&\mathbf{D}_0(g)&\\&&\mathbf{D}_2(g)\end{bmatrix}\begin{bmatrix}a_xb_x+a_yb_y+a_zb_z\\a_yb_z-a_zb_y\\a_zb_x-a_xb_z\\a_xb_y-a_yb_x\\c(a_xb_z+a_zb_x)\\c(a_xb_y+a_yb_x)\\2a_yb_y-a_xb_x-a_zb_z\\c(a_yb_z+a_zb_y)\\c(a_zb_z+a_xb_x)\end{bmatrix}
\end{align}

\purple[Tensor Product in Quantum Mechanics]{
The \textbf{tensor product} in quantum mechanics is used to describe the \textbf{overlap (or coupling) of two electron orbitals} with well-defined angular momentum states $|l, m_l\rangle $ and $|k, m_k\rangle$. These are considered angular momentum \textbf{eigenstates} because the values of the total angular momentum ($l$ and $k$) and magnetic quantum number ($m_l$ and $m_k$) are \textbf{eigenvalues} of the angular momentum operators, with their eigenfunction being the corresponding spherical harmonic function. Since angular momenta are vector quantities, we can consider the uncoupled angular momentum vectors of each eigenstate as a 3-dimensional vector, and the coupled angular momentum as the vector addition of the eigenstate vectors.

When the angular momentum vectors of both uncoupled eigenstates are perfectly aligned, their coupled state has a maximum momentum of $k + l$. When they are perfectly disaligned, their coupled state has a minimum magnitude of $|k - l|$. The two eigenstates can overlap in any relative orientation, so the magnitude of the angular momentum vector of the coupled state can theoretically be anywhere between these two boundaries.

However, we discussed earlier that angular momentum is \textbf{quantized}, and can only have discrete values corresponding to non-negative integer values of $l$ and integer values of $m$ from $-l$ to $l$. This means that we can think of the coupled state as a \textbf{probability distribution of coupled eigenstates} with well-defined angular momenta corresponding to integer values of $l$ between $|k - l|$ to $k + l$. The $m$ value of the coupled eigenstates must be equal to the sum of the uncoupled eigenstates ($m = m_{l} + m_k$) since the projection of angular momentum on the z-axis is a scalar value without directionality. The probabilities of finding each coupled eigenstate in the total coupled state are represented by the \textbf{Clebsch-Gordan coefficients}, which can be used to decompose tensor products (total coupled state) into their spherical tensor components (coupled eigenstates).}

\subsection{Clebsch-Gordan Decomposition}
\label{subsec:cb-decomp}
Now, we will define the \textbf{Clebsch-Gordan coefficients (CG coefficients)} that form the change-of-basis matrices transforming tensors from the combined tensor product space into their spherical tensor components.

First, we will develop some intuition about the purpose of the CG coefficients in the context of angular momentum coupling.

\begin{figure}[h!]
    \centering
    \includegraphics[width=\linewidth]{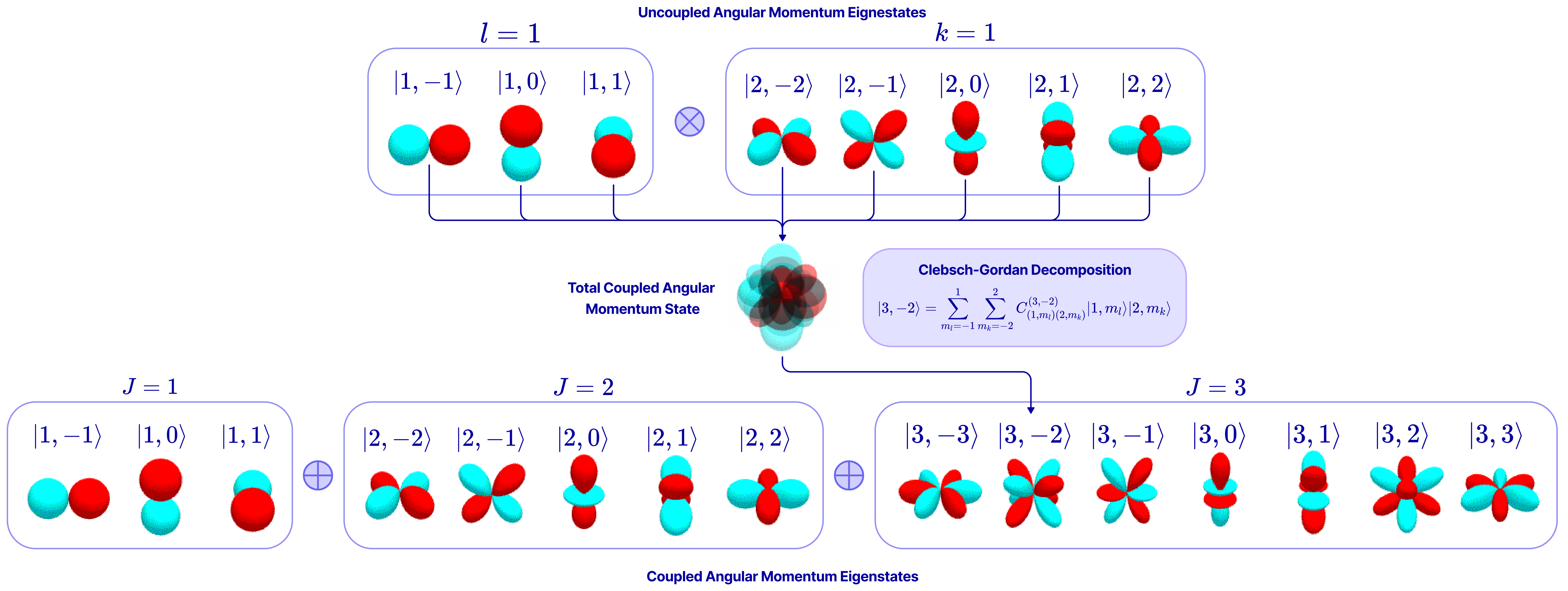}
    \caption{The coupling of two angular momentum eigenstates (which is equivalent to the tensor product) can be written as the direct sum of coupled angular momentum eigenstates ranging from $|k-l|$ to $k+l$.}
    \label{fig:clebsch}
\end{figure}

\purple[Clebsch-Gordan Coefficients and Angular Momentum]{The coupled angular momentum state is a \textbf{probability distribution of coupled eigenstates} with well-defined angular momenta. Each \textbf{Clebsch-Gordan coefficient} indicates the \textbf{amplitude of the wavefunction} corresponding to an eigenstate $|J, m\rangle$ in the wavefunction of the coupled state $|l, m_l\rangle |k, m_k\rangle$. The square of the absolute value of the CG coefficients is the probability of finding the eigenstate $|J, m\rangle$ in the coupled state $|l, m_l\rangle |k, m_k\rangle$, which means the probabilities across all coupled eigenstates must sum to 1:
\begin{align}
    \sum_{J=|k-l|}^{|k+l|}|C^{(J, m)}_{(l, m_l)(k, m_k)}|^2=1
\end{align}

Furthermore, we can obtain the wavefunction of the coupled eigenstate $|J, m\rangle$ for a defined value of $J$ as a linear combination of coupled states $|l, m_l\rangle|k, m_k\rangle$ for different values of $m_l$ and $m_k$ scaled by the CG coefficients.

\begin{align}
    |J, m\rangle=\sum_{m_l=-l}^l\sum_{m_k=-k}^kC^{(J, m)}_{(l, m_l)(k, m_k)}|l, m_l\rangle|k, m_k\rangle
\end{align}
}

There are $(2l + 1)(2k + 1)(2J + 1)$ CG coefficients needed for the decomposition from the tensor product to one type-$J$ spherical tensor component, which can be represented as a $(2l + 1)(2k + 1) \times (2J + 1)$ matrix. This is only a slice of the $(2l + 1)(2k + 1) \times (2l + 1)(2k + 1)$ matrix needed for the decomposition of the tensor product for all values of $J$.

When decomposing the tensor product of a type-$k$ and type-$l$ tensor, each CG coefficient \textbf{scales the product of the $m_l$ dimension of the type-$l$ tensor and the $m_k$ dimension of the type-$k$ tensor} to give the \textbf{$m$th dimension of the type-$J$ spherical tensor component}.

\begin{align}
    (\mathbf{s}^k\otimes\mathbf{t}^l)^{(J)}_m=\sum_{m_l=-l}^l\sum_{m_k=-k}^kC^{(J, m)}_{(l, m_l)(k, m_k)}s^{(k)}_{m_k}t^{(l)}_{m_l}
\end{align}

\begin{wrapfigure}{r}{0.5\linewidth} % or {l} instead of {r}
    \centering
    \includegraphics[width=\linewidth]{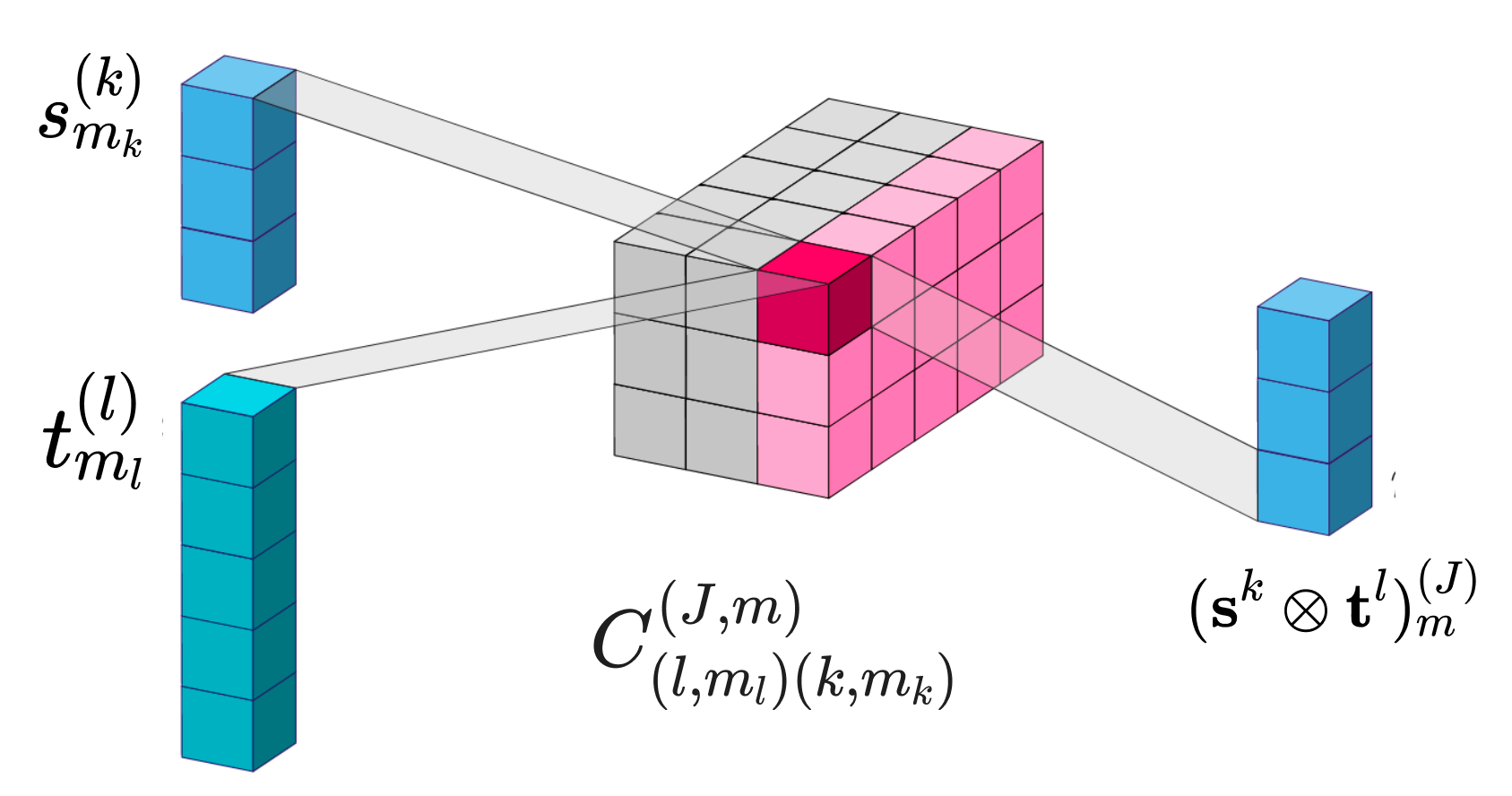}
    \caption{A visual representation of all the Clebsch-Gordan coefficients corresponding to the type-$J$ component of the tensor product between type-$k$ and type-$l$ spherical tensors arranged in a 3-dimensional list. Each element of the list is a single CG coefficient that scales the $(m_k, m_l)$ element of the tensor product to obtain the $m$th element of the type-$J$ spherical tensor component. (Source: \citet{duval2023hitchhiker})}
    \label{fig:cg-coe}
\end{wrapfigure}

We can consider each CG coefficient $C$ as a scaling factor that projects the $(m_k, m_l)$ element of the tensor in the combined $k \otimes l $ tensor space to the orthogonal $m$th element of the orthogonal type-$J$ spherical tensor subspace.

In Section \ref{subsec:basis-kernels} on computing the basis kernels, we will break down how to calculate the Clebsch-Gordan change-of-basis matrices using the Sylvester equation.

\subsection{Parameterizing the Tensor Product}

A core mechanism of spherical equivariant geometric GNNs is combining tensors of various types with learnable weights and \textbf{training them to learn rotationally symmetric relationships between nodes}. To do this, we must introduce learnable parameters into the tensor product without breaking equivariance.

As we defined in Section \ref{subsec:tensor-product}, the tensor product of a type-$k$ and type-$l$ spherical tensor can be decomposed into $k + l- |k - l| + 1 = 2\min(l, k)+1$ spherical tensors of types ranging from $J = |k - l|$ to $k + l$. Since the output of the tensor product is no longer a spherical tensor, directly applying learnable weights to the tensor product will break equivariance since the elements do not transform predictably under rotation. 

Instead, we can apply learnable scalar weights separately to each component of the decomposed spherical tensor components, since they are \textbf{orthogonal} and \textbf{rotate independently under the \textit{irreps} of SO(3)}.

\begin{align}
    \mathbf{D}_J(g)w_{(l, k,J)}(\mathbf{s}^k\otimes\mathbf{t}^l)_J=w_{(l, k,J)}\mathbf{D}_J(g)(\mathbf{s}^k\otimes\mathbf{t}^l)_J\tag{$\forall g\in G$}
\end{align}

The equivariance condition holds since the weight $w$ is a type-0 scalar that is invariant to rotations.

We can define the parameterized tensor product of a type-$k$ spherical tensor $\mathbf{s}$ and a type-$l$ spherical tensor $\mathbf{t}$ as the direct sum of the type-$J$ spherical tensor components each scaled by a weight indexed by the type of first input ($k$), the type of the second input ($l$), and the type of the component it is applied to ($J$) from the tensor product decomposition.

\begin{figure}[h!]
    \centering
    \includegraphics[width=0.4\linewidth]{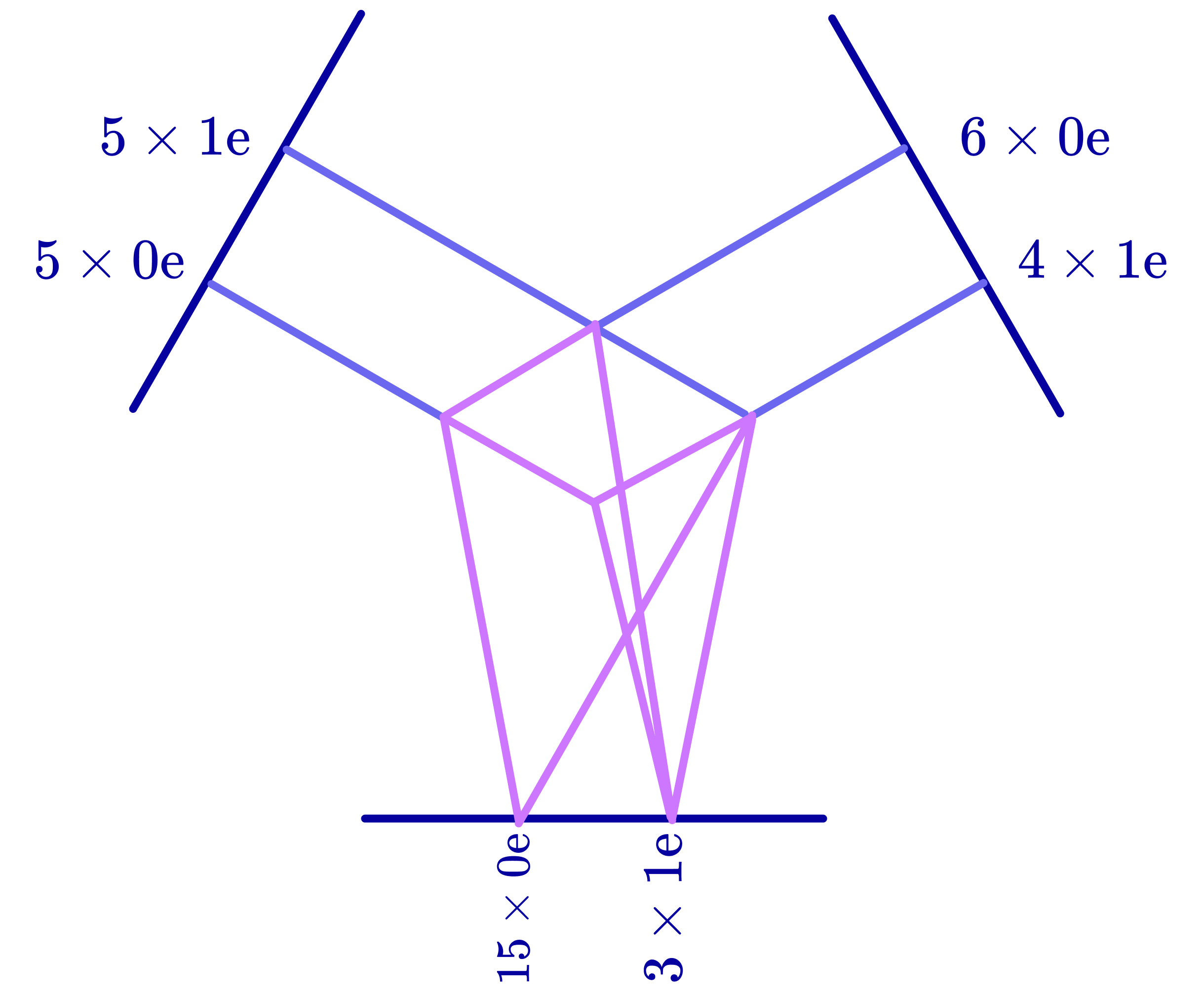}
    \caption{Representation of paths in a tensor product of two spherical tensor lists: one with 5 type-0 tensors and 5 type-1 tensors, and the other with 6 type-0 tensors and 4 type-1 tensors. If the output is restricted to only type-0 and type-1 tensors (for computational efficiency), we would get 15 type-0 tensors, 3 type-1 tensors, and a total of 18 paths. (Adapted from \href{https://docs.e3nn.org/en/latest/api/o3/o3_tp.html}{e3nn docs})}
    \label{fig:tp}
\end{figure}

\begin{align}
    \bigoplus_{J=|k-l |}^{k+l}w_{(l, k, J)}(\mathbf{s}^k\otimes\mathbf{t}^l)_J\label{eq:tensor-prod}
\end{align}
where the subscript $J$ denotes the type-$J$ component of the tensor product decomposition.

When taking the tensor product between two lists of spherical tensors of multiple channels of multiple feature types, we can think of the \textbf{combination} of a single tensor type from the first tensor list, a single type from the second tensor list, and the type of the decomposed tensor product component $(k, l, J)$ as a \textbf{path in the tensor product and assign a learnable parameter to each path} (visualized in Fig \ref{fig:tp}). In the case of (\ref{eq:tensor-prod}), where we took the tensor product between a single type-$k$ and type-$l$ spherical tensor, the number of learnable weights equals the $2\min(l, k) +1$ possible values of $J$.

\textbf{First, we will generalize to multi-type input tensors with a single channel of each type before extending to multiple channels in the section on Tensor Field Networks.} The tensor product of the tensor list $\mathbf{s}$ of types ranging from $k = 0$ to $K$ with the tensor list $\mathbf{t}$ of types ranging from $l = 0$ to $L$ can be written as follows:

\begin{align}
    \mathbf{s}^{k=0..K}\otimes\text{ }\mathbf{t}^{l=0..L}= \bigoplus_{J=0}^{K+L}\left(\sum_{\text{Path}(l, k, J)}w_{(l, k, J)}(\mathbf{s}^k\otimes\mathbf{t}^l)_J\right)
\end{align}

\begin{enumerate}
    \item For every combination of input types $(k, l )$ where $k = 0$ to $K$ and $l  = 0$ to $L$ where $J$ is in the range $|k-l|$ to $|k+l|$, we \textbf{extract the type-$J$ component of the decomposed tensor product between the type-$l$ channel and the type-$k$ channel of the tensor list}. Since each extracted type-$J$ tensor corresponds to a single path $(k, l, J)$, we scale the output with a unique learnable weight $w$. 
    
    \item Then, we take the \textbf{element-wise sum of the weighted type-$J$ tensors from every path $(k, l , J)$ that shares the same value for $J$}, since they are all $(2J+1)$-dimensional and transform with the same \textit{irreps} under rotation. 
    
    \item By repeating steps 1 and 2, we get a \textbf{single tensor for every possible degree $J$ from 0 to $K+L$}, each of which is the weighted sum of all the decomposed type-$J$ components generated from the tensor products between different combinations of degrees $l$ and $k$ from the input tensor lists.
    
    \item These vectors can be concatenated into a single vector containing subvectors of types $J = 0$ to $K+L$.
\end{enumerate}

Now, we can explicitly define each \textbf{entry} of the type-$J$ spherical tensor component ($m = -J to J$) of the tensor product decomposition using the Clebsch-Gordan coefficients:

\begin{align}
    (\mathbf{s}^k\otimes\mathbf{t}^l)^{(J)}_m=\sum_{m_l=-l}^l\sum_{m_k=-k}^kC^{(J, m)}_{(l , m_l)(k, m_k)}s^{(k)}_{m_k}t^{(l )}_{m_l}
\end{align}

Let’s break down how to calculate the parameterized tensor product with an example that applies to equivariant kernels: the tensor product between the list of feature tensors and the list of spherical harmonics projections of the angular unit vector.

Suppose we have a feature list f that contains a type-0 and type-1 tensor. We also have the angular unit vector $\hat{\mathbf{x}}$ between nodes $i$ and $j$ projected into type-0 and type-1 tensors using the type-0 and type-1 spherical harmonics. \textbf{We can denote each tensor list as a vector of stacked spherical tensors:}

\begin{align}
    \vec{\mathbf{f}}^{(0:1)}=\begin{bmatrix}\mathbf{f}^{(0)}\\\mathbf{f}^{(1)}\end{bmatrix}\tag{feature list}\\\vec{\mathbf{Y}}^{(0:1)}(\hat{\mathbf{x}}_{ij})=\begin{bmatrix}\mathbf{Y}^{(0)}\\\mathbf{Y}^{(1)}\end{bmatrix}\tag{projection list}
\end{align}

First, we can determine every path $(l, k, J)$ that results in a type-$J$ spherical tensor component for all possible values of $J$ ranging from $|0-0|=0$ to $1+1=2$.

\begin{align}
J&=0: (0, 0, 0), (1, 1, 0)\\J&=1: (0, 1, 1), (1, 0, 1), (1, 1, 1)\\J&=2: (1, 1, 2)
\end{align}

Then, we compute the tensor products between every pair of tensors in the input lists, decompose them into their spherical tensor components, scale each path by a weight, and take the sum over all the outputs with the same degree.

The \textbf{type-0} ($J = 0$) sum of the paths $(0, 0, 0)$ and $(1, 1, 0)$ is a \textbf{scalar}:

\begin{align}
    w_{(0,0,0)}(\mathbf{f}^{(0)}\otimes \mathbf{Y}^{(0)})_{0}+w_{(1,1,0)}(\mathbf{f}^{(1)}\otimes \mathbf{Y}^{(1)})_{0}
\end{align}

The \textbf{type-1} ($J = 1$) sum of the paths $(0, 1, 1)$, $(1, 0, 1)$, and $(1, 1, 1)$ is a \textbf{3-dimensional vector}:

\begin{align}
    w_{(0,1,1)}(\mathbf{f}^{(0)}\otimes \mathbf{Y}^{(1)})_1+w_{(1,0,1)}(\mathbf{f}^{(1)}\otimes \mathbf{Y}^{(0)})_1+w_{(1,1,1)}(\mathbf{f}^{(1)}\otimes \mathbf{Y}^{(1)})_1
\end{align}

The \textbf{type-2} ($J = 2$) output of the path (1, 1, 2) is a \textbf{5-dimensional vector}:

\begin{align}
    w_{(1,1,2)}(\mathbf{f}^{(1)}\otimes \mathbf{Y}^{(1)})_2
\end{align}

Finally, we can concatenate each weighted sum into a single 9-dimensional vector to get the output of the tensor product between the feature tensor list and the list of spherical harmonics projections of the angular unit vector.

\begin{align}
\begin{bmatrix}\mathbf{f}^{(0)}\\\mathbf{f}^{(1)}\end{bmatrix}\otimes\begin{bmatrix}\mathbf{Y}^{(0)}\\\mathbf{Y}^{(1)}\end{bmatrix}=\begin{bmatrix}w_{(0,0,0)}(\mathbf{f}^{(0)}\otimes \mathbf{Y}^{(0)})_0+w_{(1,1,0)}(\mathbf{f}^{(1)}\otimes \mathbf{Y}^{(1)})_0\\\\w_{(0,1,1)}(\mathbf{f}^{(0)}\otimes \mathbf{Y}^{(1)})_1+w_{(1,0,1)}(\mathbf{f}^{(1)}\otimes \mathbf{Y}^{(0)})_1+w_{(1,1,1)}(\mathbf{f}^{(1)}\otimes \mathbf{Y}^{(1)})_1\\\\\underline{w}_{(1,1,2)}(\mathbf{f}^{(1)}\otimes \mathbf{Y}^1)_2\end{bmatrix}
\end{align}

\subsection{Review}

Since we have covered quite a lot of concepts, let’s synthesize these concepts in the context of SO(3)-equivariance:

\begin{enumerate}
    \item The group of rotations in three dimensions is called the \textbf{SO(3) group}, and the group representations are $N \times N$ orthogonal matrices that can be decomposed into irreducible representations (\textit{irreps}) called \textbf{Wigner-D matrices}.
    
    \item \textbf{Wigner-D matrices} can act on any tensors after applying a change-of-basis matrix; however, they act directly on special types of tensors called \textbf{spherical tensors} that are generated by \textbf{spherical harmonics}. Spherical harmonics form a complete, orthonormal basis of functions on the unit sphere that can project vectors on the unit sphere to spherical tensors that can be transformed directly and equivariantly with Wigner-D matrices.
    
    \item These \textbf{special spherical tensors} are divided into \textbf{types (or degrees)} that are denoted by a non-negative integer ($l = 0, 1, \dots $) and are $(2l+1)$-dimensional. The Wigner-D matrices that act directly on type-$l$ tensors have dimensions $(2l+1) \times (2l+1)$, and there is a set of $2l +1$ spherical harmonic functions that project vectors into type-$l$ tensors. Higher-degree spherical tensors change more rapidly under rotation and are represented by higher-frequency spherical harmonic functions on the unit sphere.
    
    \item The \textbf{tensor product} is a tensor operator that \textbf{transforms two lower-degree tensors into a higher-degree tensor}. The tensor product of two spherical tensors is no longer a spherical tensor but can be separated into exactly one spherical tensor of each type, ranging from $|k - l|$ to $k + l$ by multiplication with a change of basis matrix containing Clebsch-Gordan coefficients.
    
    \item The tensor product allows us to \textbf{pass messages from a type-$k$ feature from a neighborhood node to a type-$l$ feature at the center node without breaking equivariance}. This is done by extracting the type-$l$ spherical tensor component of the tensor product between the type-$k$ feature with a spherical tensor generated from spherical harmonics.
    
    \item \textbf{Weights or learnable parameters} can only be applied \textit{after} decomposing the tensor product into its spherical tensor components, which means a maximum of one parameter can be applied for every set of values $(l, k, J)$ corresponding to the two input types and the tensor product component type, respectively.
\end{enumerate}

Now, we will put these ideas into practice to generate an \textbf{equivariant kernel} that transforms between degrees of spherical tensors and generates messages between nodes.

Before getting started, the code implementation of SE(3)-Transformers uses a novel data structure called \textbf{fibers} to keep track of node and edge features. The structure of a fiber is a \textbf{list of tuples} that are used to define the degrees and number of channels that are inputted and outputted from equivariant layers in the form [(multiplicity or number of channels, type or degree)]. The code implementation will often extract a (multiplicity, degree) pair from a fiber structure in the following way\footnote{Full implementation of the data structure can be found at\\ \href{https://github.com/FabianFuchsML/se3-transformer-public/blob/master/equivariant_attention/fibers.py}{\texttt{https://github.com/FabianFuchsML/se3-transformer-public/blob/master/equivariant\_attention/fibers.py}}}:

\begin{lstlisting}[language=Python]
# extract the multiplicity and degree of the input fiber
for (mi, di) in f_in.structure
\end{lstlisting}

\newpage
\section{Constructing an Equivariant Kernel}
\label{sec:3}
To facilitate message-passing between spherical tensors of different types, we want to construct a \textbf{kernel} that can take a single type-$k$ input feature and directly transform it into a type-$l$ feature.

In a non-equivariant setting, this is simple. All we need is to multiply by a $(2l + 1) \times (2k + 1)$ kernel of learnable weights to transform the $(2k + 1)$-dimensional type-$k$ tensor to a $(2l + 1)$-dimensional type-$l$ tensor. However, multiplying a randomly initialized kernel would break equivariance, so we must carefully define how to construct a kernel $\mathbf{W}$ of the same dimensions that linearly transforms type-$k$ to type-$l$ tensors while preserving SO(3)-equivariance.

This kernel should also be dependent on the displacement vector from node $j$ to $i$, which encodes the distance and relative angular relationship between the two nodes.

With these ideas in mind, we define the kernel $\mathbf{W}$ as a \textbf{function that takes the displacement vector as input and outputs a $(2l + 1) \times (2k + 1)$ linear transformation matrix}.

\begin{align}
    \mathbf{W}^{l k}(\mathbf{x}_{ij}):\mathbb{R}^3\to \mathbb{R}^{(2l+1)\times (2k+1)}
\end{align}

\subsection{Deriving the Equivariant Kernel Constraint}
\purple[]{
To construct an equivariant kernel, we must first define how it must operate under rotations. In this section, we will derive the kernel constraint from scratch, as it is the fundamental building block of equivariant GNNs. 
}

We know that $\mathbf{W}$ transforms a type-$k$ input feature from node $j$ to type-$l$ features for message passing to node $i$:

\begin{align}
    \mathbf{f}^l_{\text{out},j}=\mathbf{W}^{l k}(\mathbf{x}_{ij})\mathbf{f}^k_{\text{in,}j}
\end{align}

This must satisfy the equivariance constraint on SO(3):

\begin{align}
    \mathbf{D}_l(g)\mathbf{f}^l_{\text{out},j}=\mathbf{W}^{l k}(\mathbf{x}_{ij})\mathbf{D}_k(g)\mathbf{f}^k_{\text{in,}j}
\end{align}

We can think of the kernel applied to a specified type-$k$ feature as a \textbf{tensor field}. A field is a mathematical object that assigns a mathematical object to every point in space. In this case, for every point in 3-dimensional space (defined by the vector $\mathbf{x}$), there is an assigned type-$l$ tensor:

\begin{align}
    \mathbf{W}^{l k}(\mathbf{x}_{ij})\mathbf{f}^k_{\text{in,}j}:\mathbb{R}^3\to\mathbb{R}^{(2l+1)}
\end{align}

Rotating a tensor field is not as simple as rotating the tensors themselves since \textbf{both the orientation and position of the tensors must rotate}.

\begin{align}
    \mathbf{D}_l(g)\mathbf{f}^l_{\text{out},j}\neq\mathbf{D}_l(g)\mathbf{W}^{l k}(\mathbf{x}_{ij})\mathbf{f}^k_{\text{in,}j}
\end{align}

We can visualize this idea with an example: consider rotating the vector field $f(\mathbf{x})$, which assigns a 3-dimensional vector to every point in 3D space, by \textbf{90 degrees counterclockwise}.

\begin{align}
    f(\mathbf{x}):\mathbb{R}^3\to \mathbb{R}^3
\end{align}

\begin{figure}[h!]
    \centering
    \includegraphics[width=0.8\linewidth]{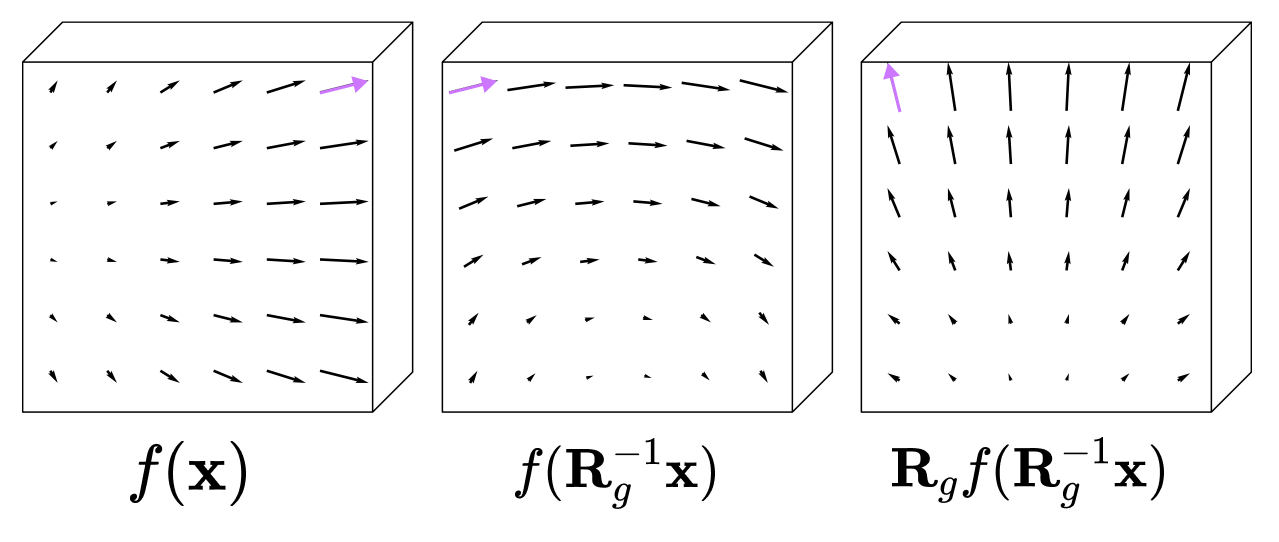}
    \caption{Diagram showing the rotation of a vector field $f$ that maps points in 3D to 3-dimensional vectors by a counter-clockwise rotation of 90 degrees. First, the vector at point $\mathbf{R}^{-1}\mathbf{x}$ (highlighted in pink) is assigned to a new rotated point without changing its orientation. Then, the vectors assigned to the rotated point $\mathbf{x}$ are themselves rotated by the rotation matrix $\mathbf{R}$. (Adapted from \citet{weiler20183d})}
    \label{fig:tensor-fields}
\end{figure}

To rotate the vector field, we must perform \textbf{two operations} (visualized in Figure \ref{fig:tensor-fields}):

\begin{enumerate}
    \item First, we \textbf{shift} the vector assigned to the point $\mathbf{R}^{-1}\mathbf{x}$ to a new, rotated point 90 degrees counterclockwise from the original point $\mathbf{x}$ without changing its orientation. This is done by applying the inverse rotation matrix to the point $\mathbf{x}$ so that the vector field $f(\mathbf{x})$ outputs the same vector as it would at the unrotated point. 
    
    \item Then, we \textbf{rotate} the vector at point $\mathbf{x}$ itself using the $3 \times 3$ rotation matrix $\mathbf{R}$. 
\end{enumerate}

The rotated tensor field can be written with the expression below, where $g$ denotes the 90-degree counterclockwise rotation:

\begin{align}
    f(\mathbf{x})\xrightarrow{\text{rotate by g}}\mathbf{R}_gf(\mathbf{R}_g^{-1}\mathbf{x})
\end{align}

\textbf{Let’s apply the same idea to rotate the tensor field expression by $g$.}

First, we shift the tensor assigned to the unrotated point $\mathbf{R}^{-1}\mathbf{x}$ to a new point rotated by $g$ without changing its orientation by applying the inverse $3 \times 3$ rotation matrix to the point $\mathbf{x}$. This sets the tensor assigned to the rotated point to be the same as the unrotated point. Then, we rotate the output type-$l$ feature of the expression by the type-$l$ Wigner-D matrix for $g$.

\begin{align}
    \mathbf{W}^{l k}(\mathbf{x}_{ij})\mathbf{f}^k_{\text{in,}j}\xrightarrow{\text{rotate by g}}\mathbf{D}_l(g)\big(\mathbf{W}^{l k}(\mathbf{R}_g^{-1}\mathbf{x}_{ij})\mathbf{f}^k_{\text{in,}j}\big)
\end{align}

This gives us the definition of rotating the type-$l$ output of the kernel by $g$ in terms of the type-$k$ input and the displacement vector:

\begin{align}
    \mathbf{D}_l(g)\mathbf{f}^l_{\text{out},j}=\mathbf{D}_l(g)\mathbf{W}^{l k}(\mathbf{R}_g^{-1}\mathbf{x}_{ij})\mathbf{f}^k_{\text{in,}j}
\end{align}

Intuitively, when we rotate the point cloud, the kernel also changes since it is dependent on the displacement vector. However, we can only apply the equivariant condition on a ‘\textit{fixed}’ function. This means that when \textbf{applied in the same way to both an unrotated and rotated point cloud, the function can recognize rotated input features and operate on them equivariantly}, such that the output is rotated accordingly. So we have to rotate the displacement vector back to its original frame to ensure that the kernel is defined the same way when operating on the rotated feature.

Now that we have defined how to rotate the entire tensor field expression by g, we can use it to \textbf{rewrite the equivariance constraint} defined earlier using our new definition for the rotated type-$l$ output feature.

\begin{align}
\mathbf{D}_l(g)\mathbf{f}^l_{\text{out},j}&=\mathbf{W}^{l k}(\mathbf{x}_{ij})\mathbf{D}_k(g)\mathbf{f}^k_{\text{in,}j}\\
\mathbf{D}_l(g)\mathbf{W}^{l k}(\mathbf{R}_g^{-1}\mathbf{x}_{ij})\mathbf{f}^k_{\text{in,}j}
&=\mathbf{W}^{l k}(\mathbf{x}_{ij})\mathbf{D}_k(g)\mathbf{f}^k_{\text{in,}j}\\
\mathbf{D}_l(g)\mathbf{W}^{l k}(\mathbf{R}_g^{-1}\mathbf{x}_{ij})
&=\mathbf{W}^{l k}(\mathbf{x}_{ij})\mathbf{D}_k(g)
\end{align}

We can substitute $\mathbf{x}$ with the vector rotated by $g$:

\begin{align}
    \mathbf{x}_{ij}\mapsto\mathbf{R}_g\mathbf{x}_{ij}
\end{align}

Then, we multiply the inverse of the type-$k$ Wigner-D matrix on both sides to get:

\begin{align}
\mathbf{D}_l(g)\mathbf{W}^{l k}(\mathbf{R}_g^{-1}\mathbf{R}_g\mathbf{x}_{ij})\mathbf{D}_k(g)^{-1}&=\mathbf{W}^{l k}(\mathbf{R}_g\mathbf{x}_{ij})\mathbf{D}_k(g)\mathbf{D}_k(g)^{-1}\\\mathbf{D}_l(g)\mathbf{W}^{l k}(\mathbf{x}_{ij})\mathbf{D}_k(g)^{-1}
&=\mathbf{W}^{l k}(\mathbf{R}_g\mathbf{x}_{ij})
\end{align}

The last line of the derivation above is called the \textbf{kernel constraint} because a kernel is SO(3)-equivariant if and only if it is a solution to the constraint.

\begin{align}
    \mathbf{W}^{l k}(\mathbf{R}_g\mathbf{x}_{ij})=\mathbf{D}_l(g)\mathbf{W}^{l k}(\mathbf{x}_{ij})\mathbf{D}_k(g)^{-1}\tag{kernel constraint}
\end{align}

We can convert the constraint into an \textbf{equivalent matrix-vector form} by vectorizing both sides and using the tensor product identity defined earlier, with the property that Wigner-D matrices are orthogonal (their inverse is equal to their transpose).

\begin{align}
    \text{vec}\big(\mathbf{W}^{l k}(\mathbf{R}\mathbf{x}_{ij})\big)=\big(\mathbf{D}_k(g)\otimes\mathbf{D}_l(g)\big)\text{vec}(\mathbf{W}^{l k}(\mathbf{x}_{ij}))
\end{align}

As discussed in Section \ref{subsec:tensor-product}, the Kronecker product of the type-$k$ and type-$l$ Wigner-D matrices (in this order) is a reproducible representation of SO(3) that acts on $(2l+1)(2k+1)$-dimensional tensors in a way that is \textbf{equivalent} to \textbf{(1)} applying a change-of-basis matrix Q that converts the tensor into the direct sum of spherical tensors of degree ranging from $|k - l|$ to $|k + l|$, \textbf{(2)} applying the block diagonal matrix of the Wigner-D \textit{irreps} corresponding to every degree ranging from $|k - l|$ to $|k + l|$, and \textbf{(3)} changing the tensor back to its original basis. $\mathbf{Q}$ is an orthogonal $(2l+1)(2k+1) \times (2l+1)(2k+1)$ matrix that is composed of Clebsch-Gordan coefficients.

\begin{align}
    \text{vec}(\mathbf{W}^{l k}(\mathbf{R}\mathbf{x}_{ij}))=\mathbf{Q}^{l k\top}\left[\bigoplus_{J=|k-l|}^{k+l}\mathbf{D}_J(\mathbf{R})\right]\mathbf{Q}^{l k}\text{vec}(\mathbf{W}^{l k}(\mathbf{x}_{ij}))
\end{align}

By multiplying both sides by $\mathbf{Q}$ and denoting the Clebsch-Gordan decomposition of the vectorized kernel with $\eta$, we can rewrite the equation as:

\begin{align}
    \mathbf{\eta}^{l k}(\mathbf{R}_g\mathbf{x}_{ij})=\left[\bigoplus_{J=|k-l|}^{k+l}\mathbf{D}_J(\mathbf{R})\right]\mathbf{\eta}^{l k}(\mathbf{x}_{ij}), \quad \text{where}\quad \eta^{l k}(\mathbf{x}_{ij})=\mathbf{Q}^{l k}\text{vec}(\mathbf{W}^{l k}(\mathbf{x}_{ij}))
\end{align}

Since $\eta$ can be directly transformed by the block diagonal matrix of type $|k - l|$ to $|k + l|$ Wigner-D \textit{irreps} without a change-of-basis, we know it must be the \textbf{direct sum of spherical tensors of degrees ranging from $J = |k - l|$ to $|k + l|$} that rotate independently under the corresponding type-$J$ Wigner-D block:

\begin{align}
    \eta^{l k}(\mathbf{x}_{ij})=\bigoplus_{J=|k-l|}^{k+l}\eta^{l k}_J(\mathbf{x}_{ij}),\;\;\;\;\eta^{l k}_J(\mathbf{R}_g\mathbf{x}_{ij})=\mathbf{D}_J(g)\eta^{l k}_J(\mathbf{x}_{ij})
\end{align}

These properties precisely define the \textbf{spherical harmonic projections} of the angular unit vector onto spherical tensors that directly rotate under Wigner-D matrices. This means we can set $\eta$ equal to the direct sum of spherical tensors of types ranging from $|k-l|$ to $|k+l|$ derived from evaluating the type-$J$ spherical harmonic functions on the unit angular displacement vector.

\begin{align}
    \eta^{l k}(\mathbf{x}_{ij})=\bigoplus_{J=|k-l|}^{k+l}\mathbf{Y}^{(J)}(\hat{\mathbf{x}}_{ij}),\;\;\;\;\eta^{l k}_J(\mathbf{x}_{ij})=\mathbf{Y}^{(J)}(\hat{\mathbf{x}}_{ij})
\end{align}

Since the spherical harmonics only restrict the orientation of the angular component of the displacement vector, we can modulate the radial distance without breaking equivariance. This allows us to incorporate a \textbf{uniquely defined radial function} (which we will define in the next section) for each value of $J$ that \textbf{maps the radial distance to a weight that scales the independently equivariant type-$J$ spherical harmonic}.

\begin{align}
    \eta^{l k}_J(\mathbf{x}_{ij})=\varphi^{l k}_J(\|\mathbf{x}_{ij}\|)\mathbf{Y}^{(J)}(\hat{\mathbf{x}}_{ij}),\;\;\;\varphi^{l k}_J:\mathbb{R}\to\mathbb{R}
\end{align}

Now, we can set the two expressions for $\eta$ equal to each other to \textbf{derive an expression for the equivariant kernel}:

\begin{align}
\mathbf{Q}^{l k}\text{vec}(\mathbf{W}^{l k}(\mathbf{x}_{ij}))&=\bigoplus_{J=|k-l|}^{k+l}\varphi^{l k}_J(\|\mathbf{x}_{ij}\|)\mathbf{Y}^{(J)}(\hat{\mathbf{x}}_{ij})\\
\text{vec}(\mathbf{W}^{l k}(\mathbf{x}_{ij}))
&=\mathbf{Q}^{l k\top}\bigoplus_{J=|k-l|}^{k+l}\varphi^{l k}_J(\|\mathbf{x}_{ij}\|)\mathbf{Y}^{(J)}(\hat{\mathbf{x}}_{ij})\\
\mathbf{W}^{l k}(\mathbf{x}_{ij})
&=\text{unvec}\left(\mathbf{Q}^{l k\top}\bigoplus_{J=|k-l|}^{k+l}\varphi^{l k}_J(\|\mathbf{x}_{ij}\|)\mathbf{Y}^{(J)}(\hat{\mathbf{x}}_{ij})\right)
\end{align}

\begin{figure}[h!]
    \centering
    \includegraphics[width=\linewidth]{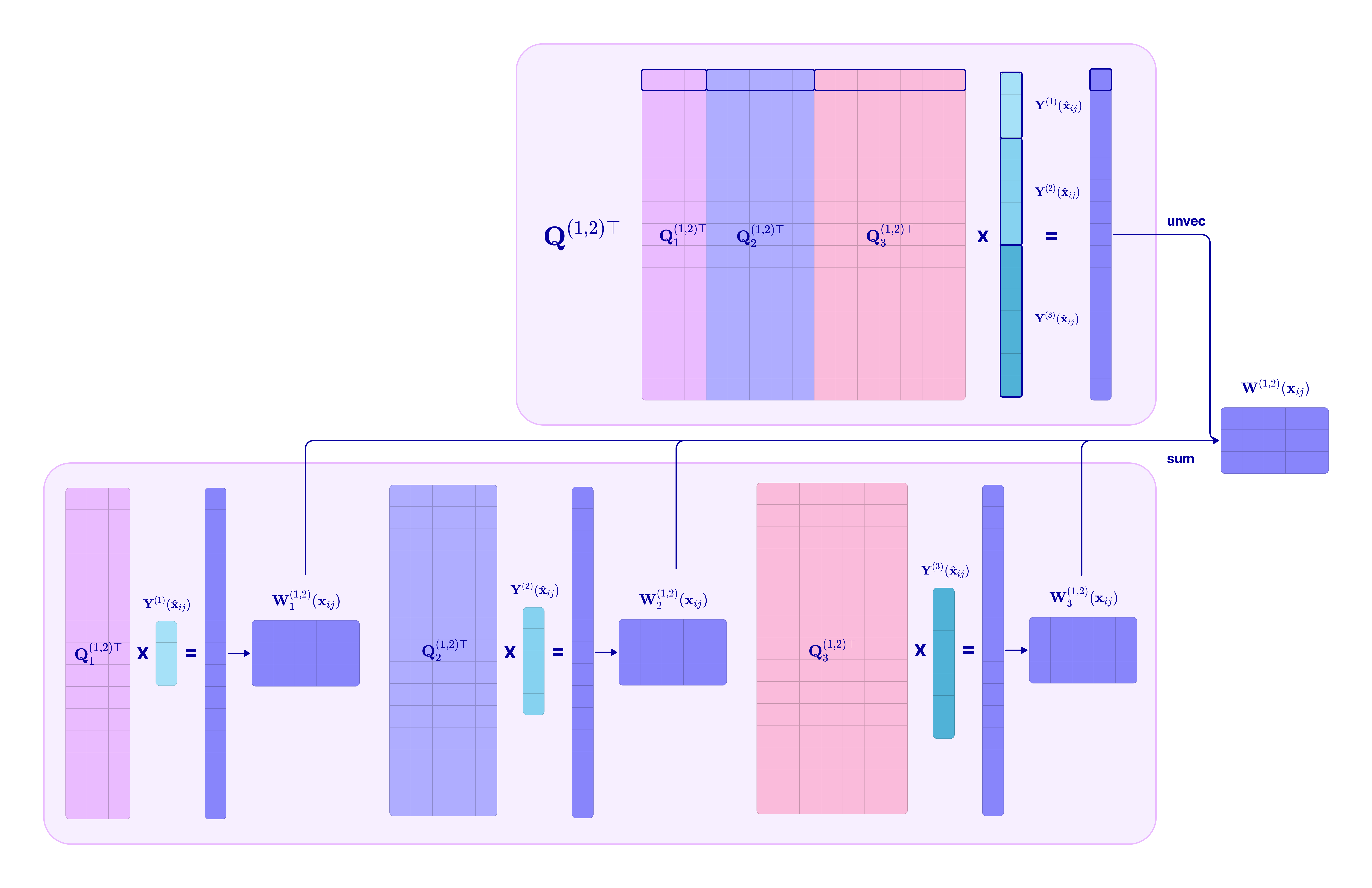}
    \caption{Multiplying the transpose of the full $(2l+1)(2k+1) \times (2l +1)(2k+1)$ Clebsch-Gordan change-of-basis matrix with the direct sum of all the types of spherical harmonics (red vectors) and unvectorizing the product (top) is equivalent to taking the matrix-vector product of each $(2l +1)(2k+1) \times (2J+1)$ type-$J$ slice of the full CG matrix with the type-$J$ component, unvectorizing the product, and taking the sum across all values of $J$ (bottom).}
    \label{fig:clebsch-gordan}
\end{figure}

Multiplying the transpose of the full $(2l+1)(2k+1) \times (2l+1)(2k+1)$ Clebsch-Gordan change-of-basis matrix with the direct sum of all the types of spherical harmonics and unvectorizing the product is \textbf{equivalent} to taking the matrix-vector product of each transposed $(2l+1)(2k+1) \times (2J+1)$ type-$J$ slice of the full CG matrix with the type-$J$ spherical tensor component, unvectorizing the product, and taking the sum across all values of $J$ (Figure \ref{fig:clebsch-gordan}). So, we can rewrite the equation for the equivariant kernel as:

\begin{align}
    \mathbf{W}^{l k}(\mathbf{x}_{ij})=\sum_{J=|k-l|}^{k+l}\varphi^{l k}_J(\|\mathbf{x}_{ij}\|)\text{unvec}\left(\mathbf{Q}^{l k\top}_J\mathbf{Y}^{(J)}(\hat{\mathbf{x}}_{ij}))\right)
\end{align}

Now, we denote the expression containing the unvectorization operation as the \textbf{type-$J$ basis kernel} that transforms the input tensor with the type-$J$ projection of the angular displacement vector and projects the output back to its original basis through the type-$J$ slice of the CG change-of-basis matrix. 

\begin{align}
    \mathbf{W}_J^{l k}(\mathbf{x}_{ij})=\text{unvec}(\mathbf{Q}^{l k\top}_{J}\mathbf{Y}^{(J)}(\hat{\mathbf{x}}_{ij})),\;\;\;\;\mathbf{Q}^{l k\top}_J\in \mathbb{R}^{(2l+1)(2k+1)\times(2J+1)}
\end{align}

The $J$th slice of the transpose of $\mathbf{Q}$ corresponds to the Clebsch-Gordan coefficients that project the orthogonal basis $J$ back to the coupled basis of type-$l$ and type-$k$ spherical tensors.

The $J$th basis kernel can also be written as a linear combination of $(2l  + 1) \times (2k + 1)$ Clebsch-Gordan matrices corresponding to fixed values of $m_l$ and $m_k$ and all values of m between $-J$ and $J$, scaled by the spherical harmonic function with degree $J$ and order m evaluated on the angular unit vector.

\begin{align}
\mathbf{W}_J^{l k}(\mathbf{x})&=\text{unvec}(\mathbf{Q}^{l k\top}_{J}\mathbf{Y}^{J}(\hat{\mathbf{x}}_{ij}))\\
&=\sum_{m=-J}^{J}\mathbf{Q}^{l k\top}_{Jm}Y^{(J)}_m(\hat{\mathbf{x}}_{ij})
\end{align}

\textbf{We have shown that every equivariant kernel lies in the orthogonal basis spanned by the basis kernels} defined for each type of spherical harmonic, ranging from $|k - l|$ to $k+l$, and can be constructed by taking the linear combination of the basis kernels.

\begin{align}
    \mathbf{W}^{lk}(\mathbf{x}_{ij})=\sum_{J=|k-l|}^{k+l}\varphi_J^{lk}(\|\mathbf{x}_{ij}\|)\mathbf{W}_J^{lk}(\|\mathbf{x}_{ij}\|)
\end{align}

Equivariant kernels are also called \textbf{intertwiners} in the literature, which refer to functions that are \textbf{linear and equivariant}. We can think of the equivariant kernel as transforming the type-$k$ input tensor in orthogonal type-$J$ spherical harmonics bases of varying degrees and transforming it back to its original basis via the transposed CG matrices. 

The equivariant kernel can \textbf{detect rotationally symmetric patterns of varying frequencies} from the input feature relevant to the prediction task, which is analogous to how a set of convolutional filters in a CNN is applied to generate multiple feature maps that enhance the signal of various patterns in the input images for object detection \citep{o2015introduction}. The signals across the different frequencies are then reduced to a single type of output tensor that can be aggregated with other messages, similar to how the set of feature maps produced by convolutional filters is aggregated into a single feature map for further processing.

\subsection{Computing The Basis Kernels}
\label{subsec:basis-kernels}
\purple[]{
Since each basis kernel is used to construct every equivariant kernel transforming between types $k$ and $l$ for a given edge across all SE(3)-equivariant layers in a model, it is convenient to calculate and store them for repeated use. This involves two steps: precomputing the change-of-basis matrices and the spherical harmonic projections of the angular unit vector for every edge up to a maximum spherical tensor degree $J$.
}

First, we want to calculate the $(2l+1)(2k+1) \times (2J+1)$ change-of-basis matrix $\mathbf{Q}$ transpose for all values of $J$, such that the following equation holds:

\begin{align}
    \mathbf{D}_k(g)\otimes\mathbf{D}_l(g)=\mathbf{Q}^{l k\top}\left[\bigoplus_{J=|l-k|}^{l+k}\mathbf{D}_J(g)\right]\mathbf{Q}^{l k}
\end{align}

Instead of constructing $\mathbf{Q}$ from the Clebsch-Gordan coefficients directly, the implementation of the 3D-steerable CNN \citep{weiler20183d} and the SE(3)-Transformer \citep{fuchs2020se} compute $\mathbf{Q}$ by solving the Sylvester equation, which we will decompose below.

$\mathbf{Q}$ converts a tensor into the direct product of spherical tensor components that each transform \textit{independently} in orthogonal subspaces via multiplication with the corresponding Wigner-D block in the block diagonal matrix. Then, the transpose of $\mathbf{Q}$ projects each of the rotated spherical tensors back to the original coupled basis. Since the operation of the \textbf{left-hand side for each value of $J$ is independent}, they must all satisfy the following equation:

\begin{align}
\mathbf{D}_k(g)\otimes\mathbf{D}_l(g)=\mathbf{Q}^{l k\top}_J\mathbf{D}_J(g)\mathbf{Q}^{l k}_J
\end{align}

Since $\mathbf{Q}$ is an orthogonal matrix where the inverse is equal to its transpose, we can rewrite the above equation as:

\begin{align}(\mathbf{D}_k(g)\otimes\mathbf{D}_l(g))\mathbf{Q}^{l k\top}_J&=\mathbf{Q}^{l k\top}_J\mathbf{D}_J(g)\\(\mathbf{D}_k(g)\otimes\mathbf{D}_l(g))\mathbf{Q}^{l k\top}_J-\mathbf{Q}^{l k\top}_J\mathbf{D}_J(g)&=0
\end{align}

This is in the form of a \textbf{homogeneous Sylvester equation} \citep{sylvester1884equation}:

\begin{align}
\mathbf{AX}-\mathbf{XB}&=0
\end{align}

In this equation, $\mathbf{A}$ and $\mathbf{B}$ are matrices, and we want to solve for the matrix $\mathbf{X}$. In our case, these matrices are defined below:

\begin{align}
    \mathbf{A}=(\mathbf{D}_k(g)\otimes\mathbf{D}_l(g))^{(J)},\;\;\;\;\mathbf{B}=\mathbf{D}_J(g),\;\;\;\;\mathbf{X}=\mathbf{Q}^{l k\top}_J
\end{align}

To simplify the computation of the matrix $\mathbf{X}$, we can convert it into a standard linear algebra problem in the form of the homogeneous equation $\mathbf{Mx} = 0$, where we can solve for the vector $\mathbf{x}$ by finding the null space of the matrix $\mathbf{M}$. To do this, we vectorize the matrix $\mathbf{X}$ and write the matrix-matrix product into an equivalent matrix-vector product.

The matrix product $\mathbf{AX}$ is equivalent to the matrix-vector product below:

\begin{align}
    \mathbf{AX}\mapsto \mathbf(\mathbf{I}\otimes \mathbf{A})\text{vec}(\mathbf{X})
\end{align}

\begin{figure}[h!]
    \centering
    \includegraphics[width=\linewidth]{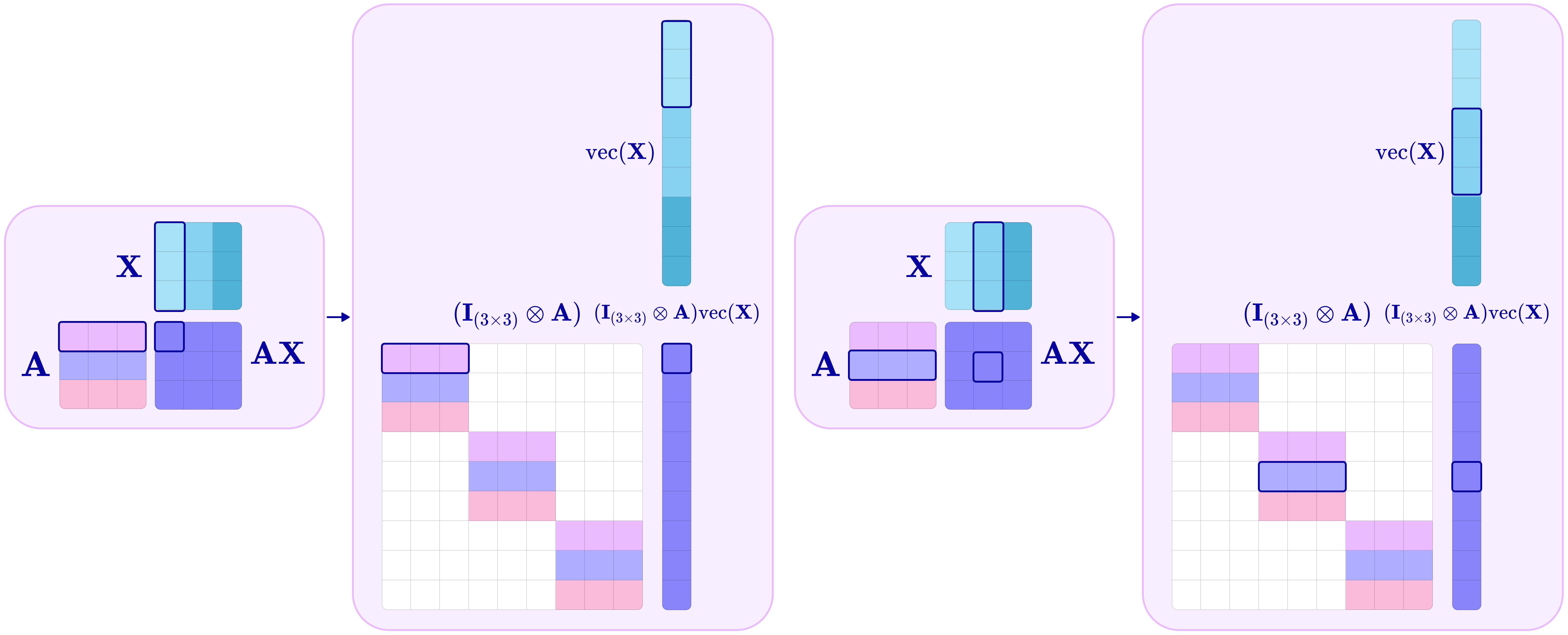}
    \caption{Taking the matrix-matrix product $\mathbf{AX}$ is equivalent to taking the matrix-vector product of the Kronecker product between the identity matrix and the matrix $\mathbf{A}$ with the vectorized form of $\mathbf{X}$.}
    \label{fig:syl1}
\end{figure}

The Kronecker product between the identity matrix $\mathbf{I}$ and the matrix $\mathbf{A}$ produces a square block diagonal matrix where every block along the diagonal is the matrix A repeated by the dimension of $\mathbf{I}$. Taking its matrix-vector product with the matrix $\mathbf{X}$ stacked into a vector produces the vectorized equivalent to the matrix product $\mathbf{AX}$ (Figure \ref{fig:syl1}). 

The matrix product $\mathbf{XB}$ is equivalent to the following matrix-vector product:

\begin{align}
    \mathbf{XB}\mapsto (\mathbf{B}^{\top}\otimes\mathbf{I})\text{vec}(\mathbf{X})
\end{align}

\begin{figure}[h!]
    \centering
    \includegraphics[width=\linewidth]{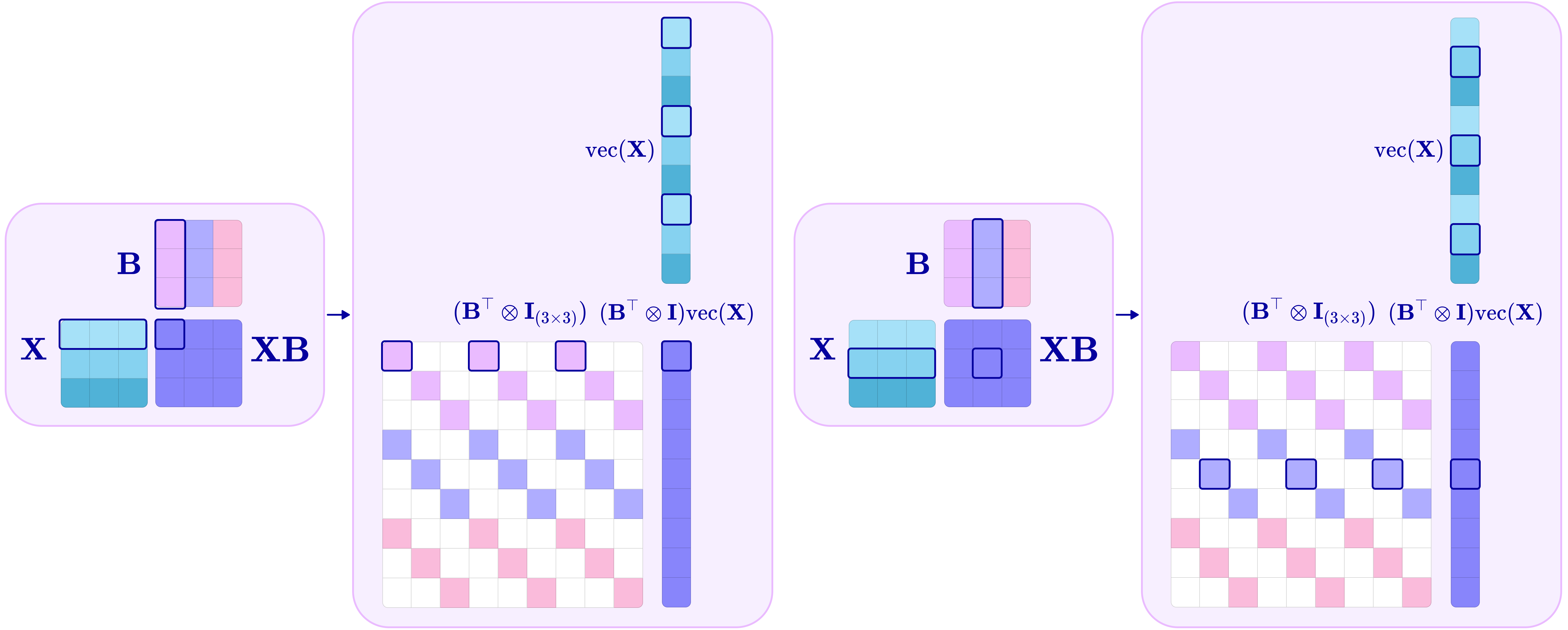}
    \caption{Taking the matrix-matrix product $\mathbf{XB}$ is equivalent to taking the matrix-vector product of the Kronecker product between the transpose of $\mathbf{B}$ and the identity matrix with the vectorized form of $\mathbf{X}$.}
    \label{fig:syl2}
\end{figure}

The Kronecker product between the transpose of $\mathbf{B}$ and the identity matrix $\mathbf{I}$ produces a \textit{matrix of matrices} where every inner matrix has the element of $\mathbf{B}$ transpose corresponding to its position in the outer matrix repeated along the diagonal with the same dimension as $\mathbf{I}$. Taking its matrix-vector product with the vectorized matrix $\mathbf{X}$ gives the vectorized equivalent to the matrix product $\mathbf{XB}$ (Figure \ref{fig:syl2}). 

Making these substitutions allows us to isolate the vectorized matrix $\mathbf{X}$:

\begin{align}
\mathbf{AX}-\mathbf{XB}&=0\\
(\mathbf{I}\otimes \mathbf{A})\text{vec}(\mathbf{X})-(\mathbf{B}^{\top}\otimes\mathbf{I})\text{vec}(\mathbf{X})&=0\\\big((\mathbf{I}\otimes \mathbf{A})-(\mathbf{B}^{\top}\otimes\mathbf{I})\big)\text{vec}(\mathbf{X})&=0
\end{align}

This is in the form of the homogenous matrix-vector product $\mathbf{Mx} = 0$, where we can solve for the vector $\mathbf{x}$ by finding the non-zero vectors in the null space of $\mathbf{M}$. In the equation above, we can solve for the vectorized matrix $\mathbf{X}$ by finding the null space of the following matrix:

\begin{align}
    (\mathbf{I}\otimes \mathbf{A})-(\mathbf{B}^{\top}\otimes\mathbf{I})
\end{align}

In our case, we want to solve the following homogeneous equation:

\begin{align}
    \left((\mathbf{I}\otimes (\mathbf{D}_k(g)\otimes\mathbf{D}_l(g)))-(\mathbf{D}_J(g)^{\top}\otimes\mathbf{I})\right)\text{vec}(\mathbf{Q}^{l k\top}_J)=0
\end{align}

Let's break down how the type-$J$ slice of the change-of-basis matrix $\mathbf{Q}^\top$ is computed in the code implementation:

\begin{enumerate}
    \item First, we calculate the type-$l$, type-$k$, and type-$J$ Wigner-D matrices on randomly defined Euler angles alpha ($\alpha$), beta ($\beta$), and gamma ($\gamma$). For the type-$l$ and type-$k$ matrices, we take their Kronecker product, producing a $(2l+1)(2k+1) \times (2l+1)(2k+1)$ matrix. Note that the implementation of the \texttt{kron} function takes the second matrix in the Kronecker product as the first input and the first matrix as the second input.

\begin{lstlisting}[language=Python]
# helper function that calculates the Kronecker product between type-l and type-k Wigner-D matrices for rotation angles a, b, c
def _R_tensor(a, b, c): 
    # kron calculates the Kronecker product between two matrices
    # irr_repr returns the irrep from (type, alpha, beta, gamma)
    return kron(irr_repr(l, a, b, c), irr_repr(k, a, b, c))

def _sylvester_submatrix(J, a, b, c):
    # calculate the Kronecker product between type-l and type-k Wigner-D matrices for rotation angles a, b, c
    R_tensor = _R_tensor(a, b, c)  # [(2l+1)*(2k+1), (2l+1)*(2k+1)]
    # calculate type-J Wigner-D matrix for same rotation
    R_irrep_J = irr_repr(J, a, b, c)  # [2J+1, 2J+1]
    return kron(R_tensor, torch.eye(R_irrep_J.size(0))) -   kron(torch.eye(R_tensor.size(0)), R_irrep_J.t())
\end{lstlisting}

    \item Now, we take the \textbf{Kronecker product} of the $(2J+1) \times (2J+1)$ identity matrix with the $(2l+1)(2k+1) \times (2l+1)(2k+1)$ Kronecker product of types $l$ and $k$ Wigner-D matrices and subtract the Kronecker product between the $(2J+1) \times (2J+1)$ transposed type-$J$ Wigner-D matrix and the $(2l+1)(2k+1) \times (2l+1)(2k+1)$ identity matrix. The complementary dimensions of the identity matrices turn $\mathbf{A}$ and $\mathbf{B}$ into $(2l+1)(2k+1)(2J+1) \times (2l+1)(2k+1)(2J+1)$ matrices that can be subtracted element-wise from each other and used to solve the Sylvester equation.

    \begin{align}
        (\mathbf{I}_{(2J+1)\times(2J+1)}\otimes (\mathbf{D}_k(g)\otimes\mathbf{D}_l(g)))-(\mathbf{D}_J(g)^{\top}\otimes\mathbf{I}_{(2l+1)(2k+1)\times (2l+1)(2k+1)})
    \end{align}

\begin{lstlisting}[language=Python]
# output of _sylvester_submatrix function 
# torch.eye returns the identity matrix with the input dimension
return kron(R_tensor, torch.eye(R_irrep_J.size(0))) -   kron(torch.eye(R_tensor.size(0)), R_irrep_J.t())
\end{lstlisting}
    
    \item Next, we call the function defined in steps 1 and 2 to generate the Kronecker product matrix for five sets of random angles and calculate the vectors in the null space of all five matrices. Using multiple sets of random angles ensures the solution is unique and can be used as the change-of-basis $\mathbf{Q}$ across all rotations in SO(3).
\begin{lstlisting}[language=Python]
# random angles that define the Wigner-D matrices
random_angles = [
    [4.41301023, 5.56684102, 4.59384642],
    [4.93325116, 6.12697327, 4.14574096],
    [0.53878964, 4.09050444, 5.36539036],
    [2.16017393, 3.48835314, 5.55174441],
    [2.52385107, 0.2908958, 3.90040975]
]
# calculate the vector that is the solution of the homogeneous equation for all sets of angles 
null_space = get_matrices_kernel([_sylvester_submatrix(J, a, b, c) for a, b, c in random_angles])
# confirm that the solution is unique
assert null_space.size(0) == 1, null_space.size()
\end{lstlisting}

    \item Finally, we reshape the vector into the matrix $\mathbf{Q}^\top$ with dimensions $(2l+1)(2k+1) \times (2J+1)$ and verify that it satisfies the homogeneous equation below for randomly generated angles.
    \begin{align}
        (\mathbf{D}_k(g)\otimes\mathbf{D}_l(g))\mathbf{Q}^{l k\top}_J-\mathbf{Q}^{l k\top}_J\mathbf{D}_J(g)=0
    \end{align}
\begin{lstlisting}[language=Python]
Q_J = Q_J.view((2*l+1)*(2*k+1), 2*J+1) #reshape
# verify that AX=XB holds for random angles a, b, c
assert all(torch.allclose(_R_tensor(a, b, c) @ Q_J, Q_J @ irr_repr(J, a, b, c)) for a, b, c in torch.rand(4, 3))

# return value of _basis_transformation_Q_J function
return Q_J
\end{lstlisting}
\end{enumerate}

Now, we must compute the spherical harmonics projections of the angular unit displacement vector. Given that the spherical harmonic function must be computed for all $2J+1$ values of $m$ corresponding to all $2\min(l,k)+1$ values of $J$ needed for all transformations between pairs of input and output degrees for every edge in the graph, the number of computations increases quickly with larger and more complex graphs.

To speed up computation, the SE(3)-Transformer precomputes the spherical harmonic projections for values of $J$ up to double the maximum feature degree (since $J$ has a maximum value of $k + l$) for every edge in the graph using \textbf{recursive relations of the associated Legendre polynomials (ALPs)}.

\begin{align}
    P^m_l(x)=(-1)^m(1-x^2)^{\frac{m}{2}}\underbrace{\frac{d^m}{dx^m}(\underbrace{P_J(x)}_{\text{degree-}J \text{ Legendre polynomial}})}_{m\text{th derivative}}
\end{align}

The ALP term of the spherical harmonic equation is the most computationally intensive and needs to be recomputed for every edge of the graph since it is dependent on $x=\cos(\theta)$, where $\theta$ is the polar angle of the angular unit vector. Using recursive relations, we only need to compute the ALP for boundary values of $m=J$, and the remaining polynomials can be derived from recursively combining the previously computed polynomials and storing them to compute the next set of polynomials.

\begin{figure}
    \centering
    \includegraphics[width=\linewidth]{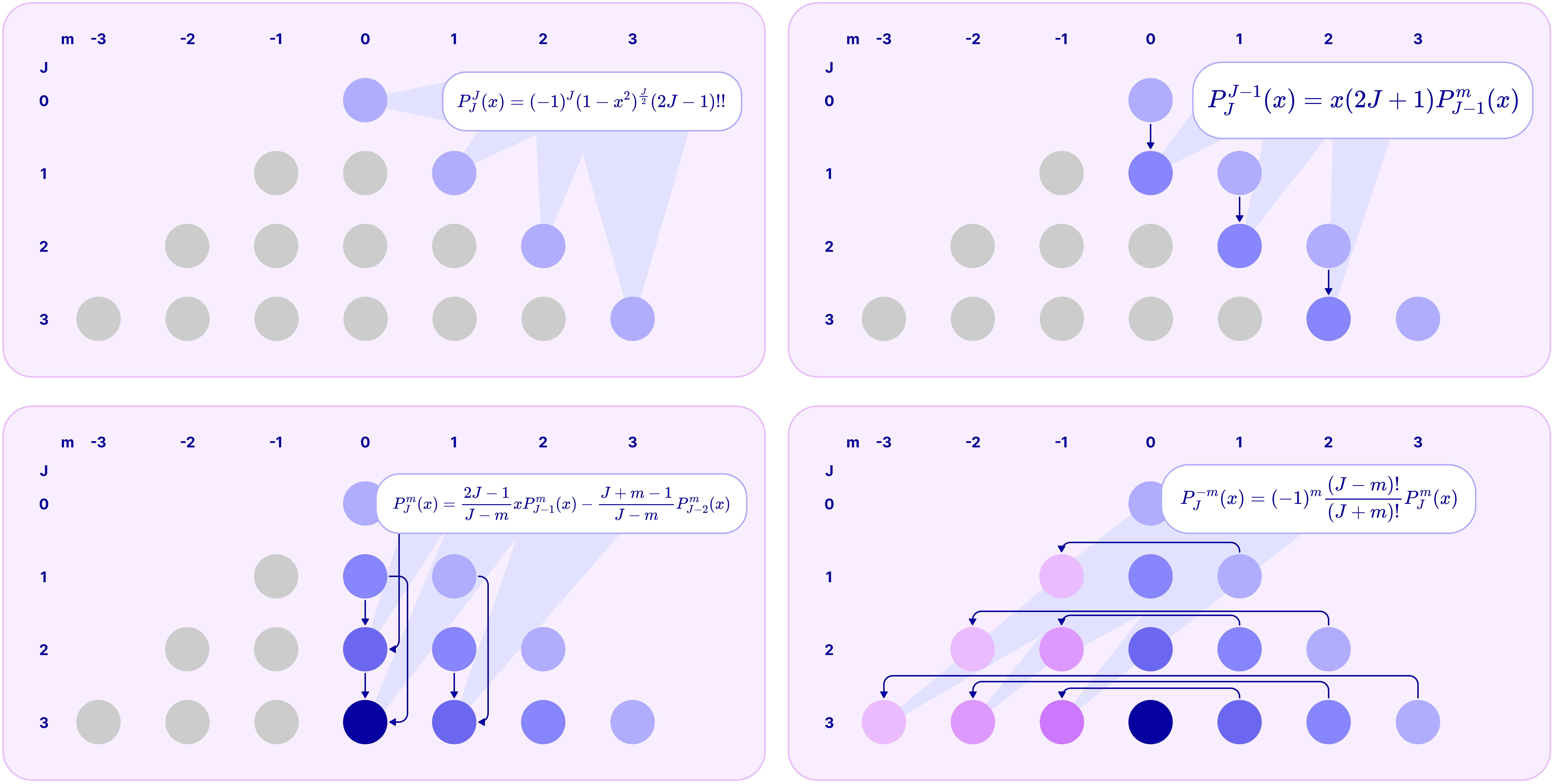}
    \caption{The sequence of steps to calculate the associated Legendre polynomials for all values of $J$ and $m$. First, we use the boundary equation to calculate the ALPs when $m=J$ (top-left). Next, we use a recursive relation to calculate the ALPs for $m=J-1$ from the boundary ALPs (top-right). Then, we use the ALPs for $J-2$ and $J-1$ in the recursive relation to calculate all remaining ALPs with non-negative values of $m$ (bottom-left). Finally, we can calculate the ALPs for all $m < 0$ by multiplying their positive counterparts by a coefficient (bottom-right).}
    \label{fig:recursive}
\end{figure}

The recursive computation of all non-zero ALPs for all values of $J$ and $m$ involves only three equations applied in the following sequence of steps (Figure \ref{fig:recursive}):

\begin{enumerate}
    \item When $m=J$, which is the maximum value of m where the ALP is non-zero, we can calculate it directly with the following equation:
    \begin{align}
        P^J_J(x)=(-1)^J(1-x^2)^\frac{J}{2}(2J-1)!!\tag{$m=J$}
    \end{align}
    where $x!!$ is the \textbf{semi-factorial} given by:
    \begin{align}
        x!! = x(x-2)(x-4)\dots
    \end{align}
    Notice that the Legendre polynomial in this equation is a constant since taking the $m$th derivative reduces the degree to 1.
    \begin{figure}[h!]
        \centering
        \includegraphics[width=0.6\linewidth]{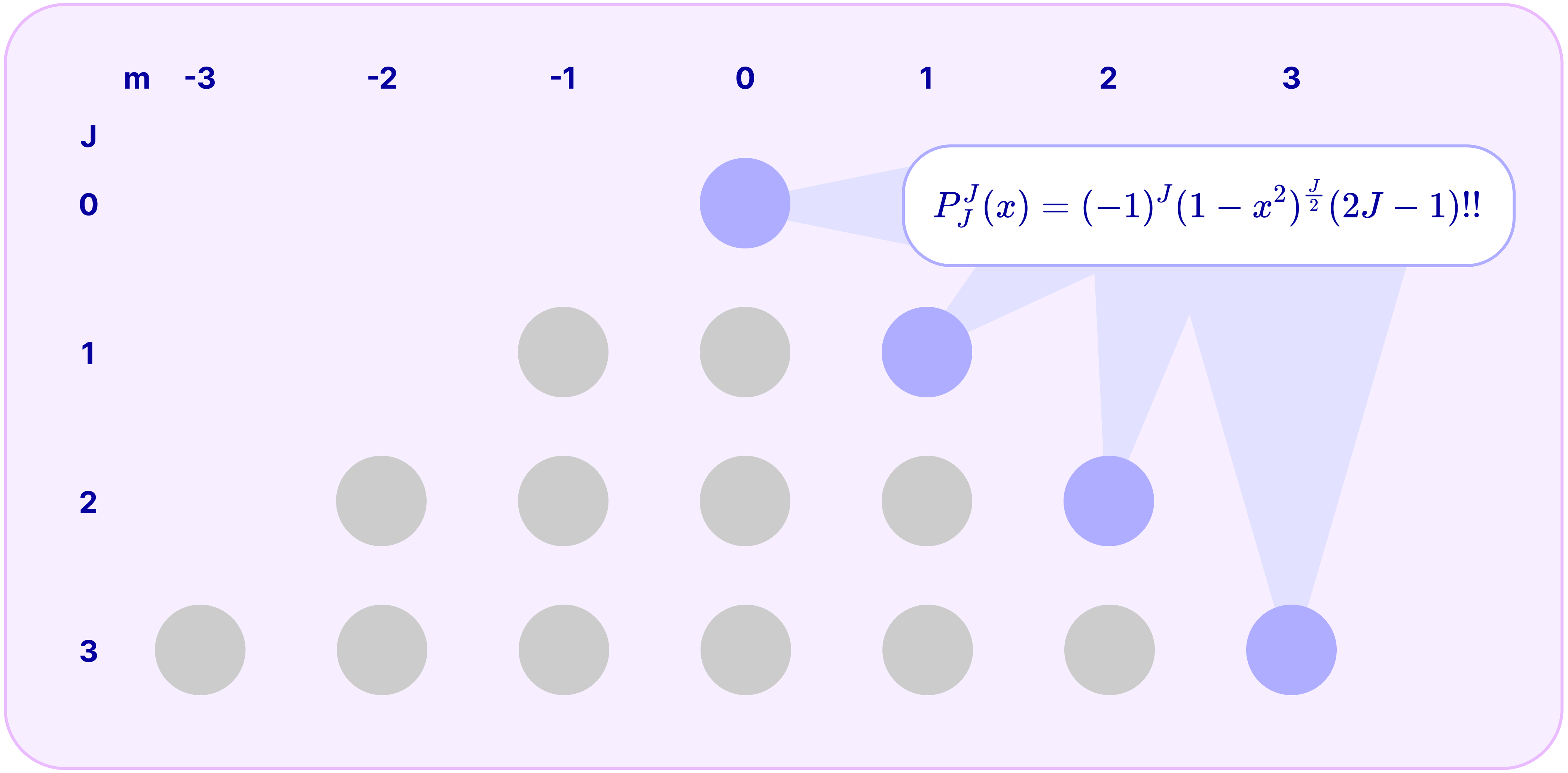}
        \caption{Calculating the boundary ALPs for $m=J$.}
        \label{fig:step1}
    \end{figure}
    
    \item Then, we can compute the polynomials for when $m=J-1$ from the polynomials stored from step 1 using the following recursive relation:
    \begin{align}
        P^{J-1}_{J}(x)=x(2J+1)P^m_{J-1}(x)\tag{$m=J-1$}
    \end{align}
    \begin{figure}[h!]
        \centering
        \includegraphics[width=0.6\linewidth]{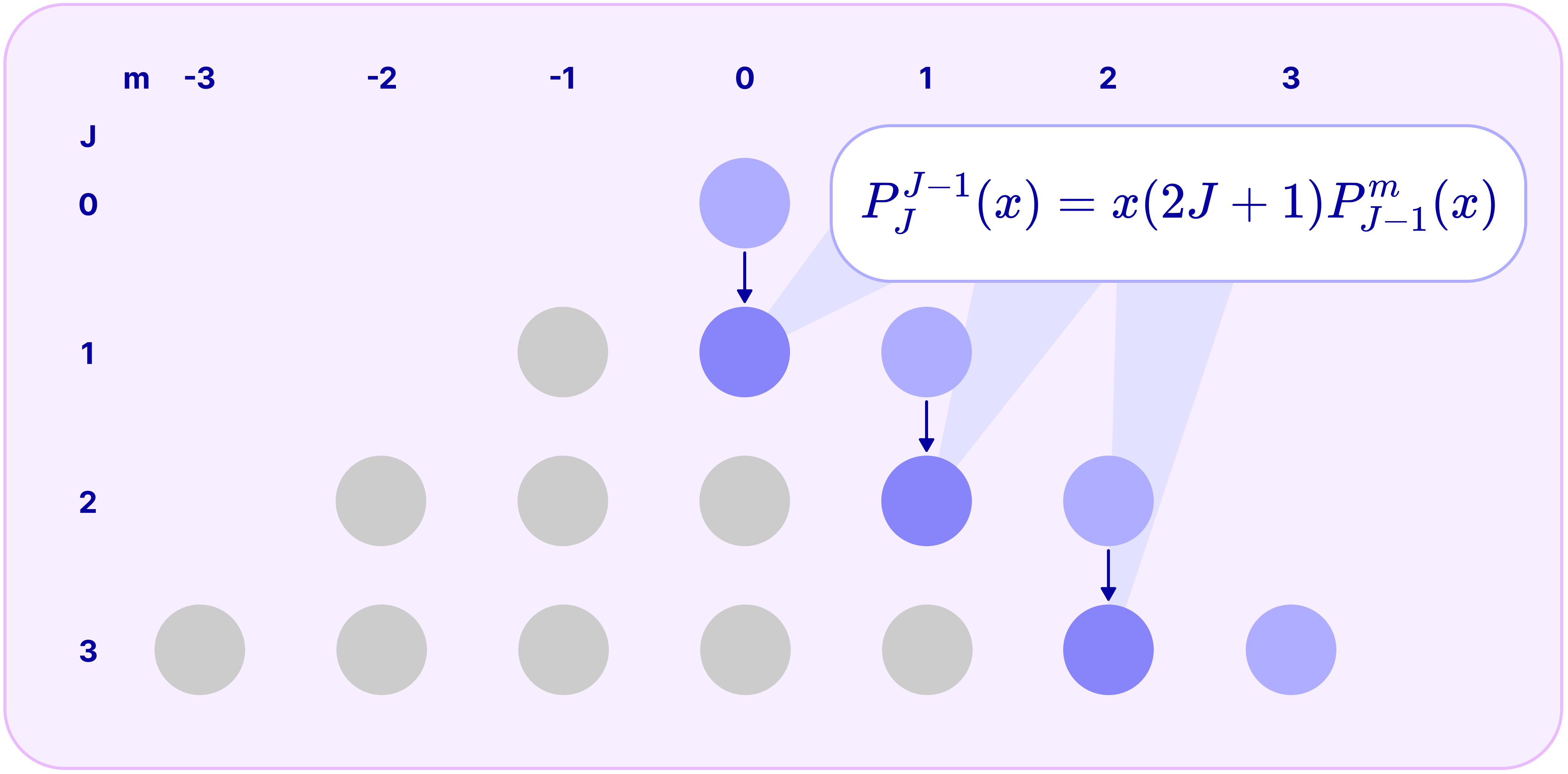}
        \caption{Calculating the ALPs for $m=J-1$ from the boundary ALPs.}
        \label{fig:step2}
    \end{figure}
    
    \item For the remaining ALPs with non-negative values of $m$, we can compute it recursively from the two preceding ALPs with order $m$ and degrees $J-1$ and $J-2$ using the recursive relation below:
    \begin{align}
        P^m_{J}(x)=\frac{2J-1}{J-m}xP^m_{J-1}(x)-{\frac{J+m-1}{J-m}P^m_{J-2}(x)}\tag{$m\geq 0$}
    \end{align}
    Notice that the recursive relation for step 2 is just the first term of this equation, simplified for $m=J-1$.
    \begin{figure}[h!]
        \centering
        \includegraphics[width=0.6\linewidth]{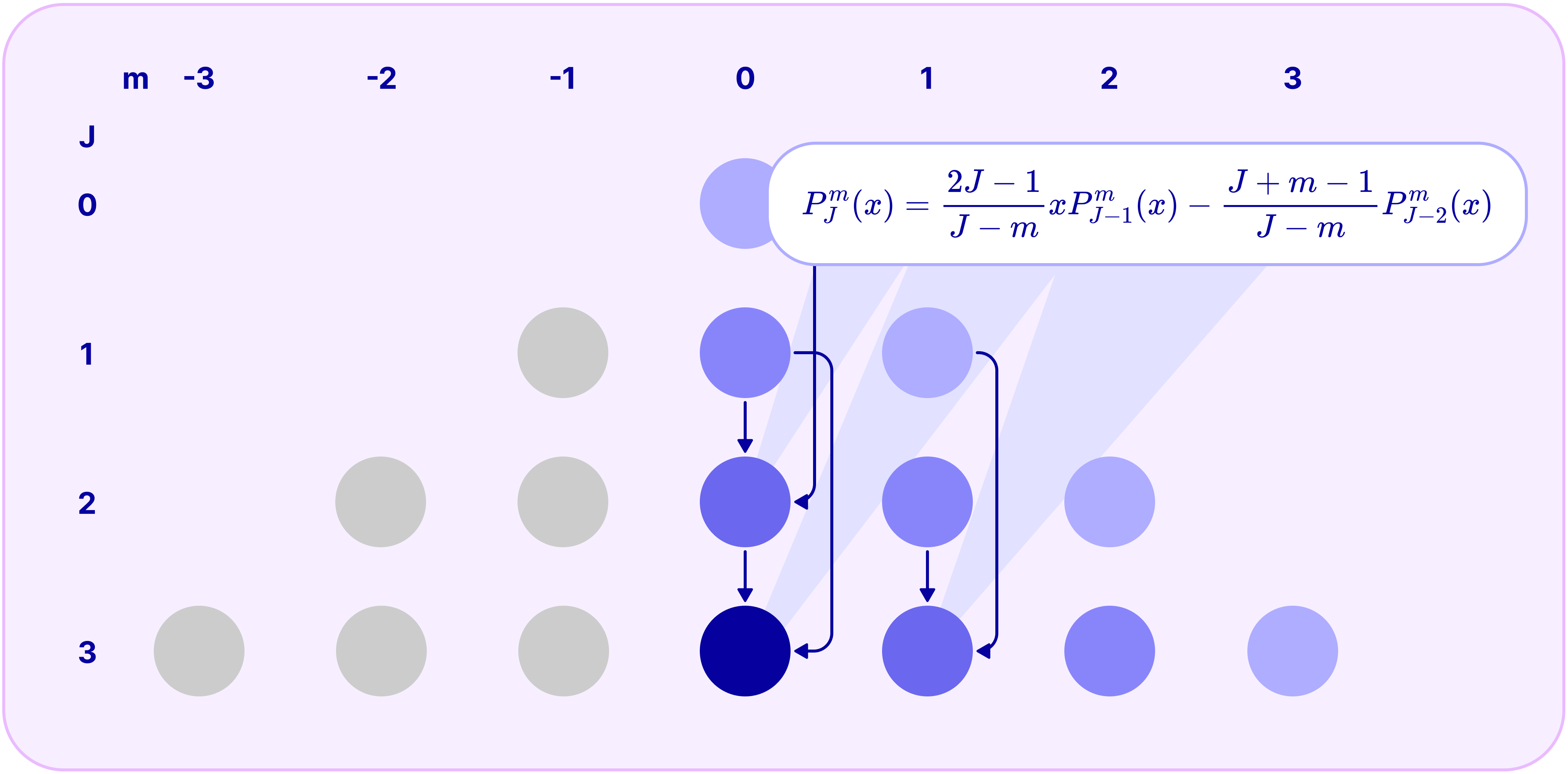}
        \caption{Calculating all ALPs for non-negative $m$ with previously computed ALPs.}
        \label{fig:step3}
    \end{figure}
    
    \item Finally, we can compute the ALPs for all negative values of $m$ from the ALP of its positive counterpart using the following relationship:
    \begin{align}
        P^{-m}_J(x)=(-1)^m\frac{(J-m)!}{(J+m)!}P^m_J(x)
    \end{align}
    \begin{figure}[h!]
        \centering
        \includegraphics[width=0.6\linewidth]{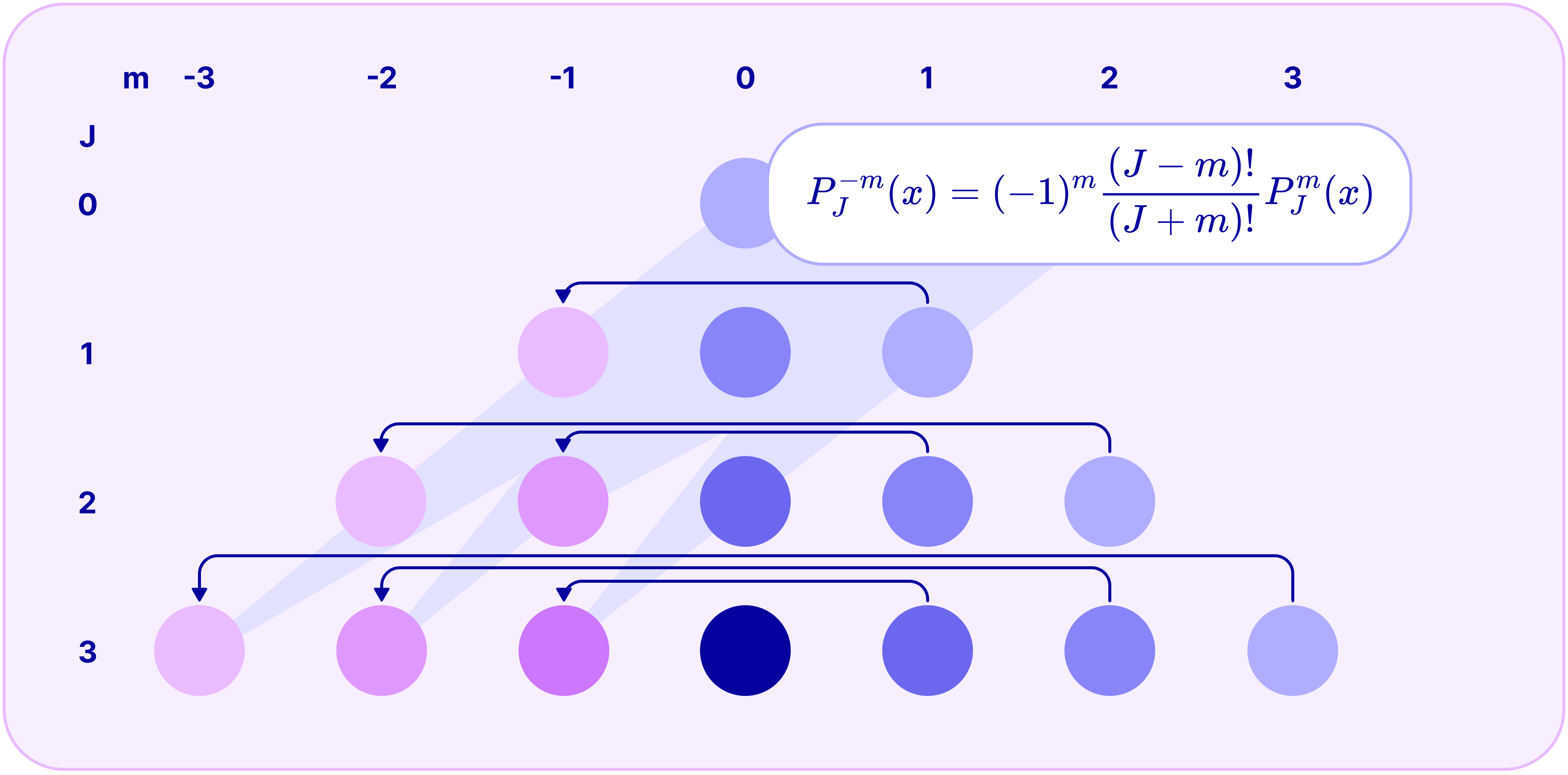}
        \caption{Calculating all ALPs for negative orders $m$ using its positive counterpart.}
        \label{fig:step4}
    \end{figure}
    
    In this equation, the fraction term can be written in terms of the inverse of a \textbf{falling factorial} from $(J+m)$ to $(J-m+1)$:
    \begin{align}\frac{(J-m)!}{(J+m)!}&=\frac{(J-m)\cdot (J-m-1)\cdot \ldots \cdot 1}{(J+m)\cdot(J+m-1)\cdot \ldots\cdot (J-m)\cdot (J-m-1)\cdot \ldots\cdot 1}\\
    &=\frac{1}{(J+m)\cdot (J+m-1)\cdot\ldots \cdot(J-m+1)}
    \end{align}
    
    We can calculate the falling factorial for a given value of $J$ and $m$ using the following helper function:
\begin{lstlisting}[language=Python]
# this is analogous to the pochammer function in the SE(3)-Transformer modified for clarity
def falling_factorial(J, m):
    # computes (J+m)*(J+m-1)*...*(J-m+1)
    f = 1.
    for n in range(J+m, J-m, -1):
      f *= n
    return f
\end{lstlisting}
    
    If $m$ is negative, the function below takes the polynomial for the absolute value of $m$ and returns the polynomial scaled by the negative coefficient:
\begin{lstlisting}[language=Python]
# y: Legendre polynomial for the absolute value of m
def negative_lpmv(self, J, m, y):
    # check if m is negative
    if m < 0:
        # multiply y by the coefficient containing the falling factorial
        y *= ((-1)**m / falling_factorial(J, m))
    return y
\end{lstlisting}
\end{enumerate}

Now that we understand how the recursive relations work, we can implement the code that returns the ALP for a given $m$ and $J$ either by applying the recursive relations using previously stored ALPs or making a recursive call to compute the ALP for $m$ and $J-1$ until the boundary where $m=J$:

\begin{lstlisting}[language=Python]
def lpmv(self, J, m, x):
    # get the absolute value of m
    m_abs = abs(m)
    # check if the polynomial has already been computed
    if (J,m) in self.leg:
        return self.leg[(J,m)]
    # check if m is out of range -J to J
    elif m_abs > J:
        return None
    # if J=0, the associated Legendre polynomial is equal to 1
    elif J == 0: # return tensor of 1s with the same shape as x
        self.leg[(l,m)] = torch.ones_like(x)
        return self.leg[(l,m)]
    
    # if |m|=J, compute the polynomial using the equation from step 1
    if m_abs == J: # calculate coefficient term
        y = (-1)**J * semifactorial(2*J-1)
        # multiply by the term dependent on x
        y *= torch.pow(1-x*x, m_abs/2)
        # negative_lpmv returns y if m is positive and y multiplied by the negative coefficient defined in step 4 if m is negative
        self.leg[(l,m)] = self.negative_lpmv(l, m, y)
        return self.leg[(l,m)]
    else:
        # recusive call to compute lower degree polynomials up to boundary m=J
        self.lpmv(J-1, m, x)

    # if m is not on the boundary, first compute the first term of the relation defined in step 3
    # if m_abs=J-1, then this calculates the relation defined in step 2
    y = ((2*J-1) / (J-m_abs)) * x * self.lpmv(J-1, m_abs, x)
    # check if m_abs!=J-1, then add the second term defined in step 3
    if l - m_abs > 1:
        y -= ((l+m_abs-1)/(l-m_abs)) * self.leg[(l-2, m_abs)]
        
    # if m is negative, return the polynomial for m_abs scaled by the negative coefficient
    if m < 0:
        y = self.negative_lpmv(l, m, y)
   
    self.leg[(l,m)] = y

    return self.leg[(l,m)]
\end{lstlisting}

After computing the ALP, we can calculate the \textbf{real spherical harmonics from the associated Legendre polynomials} with the following equations, depending on the sign of $m$:

\begin{align}
    Y^{(J)}_m(\theta,\phi)=\begin{cases}\sqrt2\sqrt{\frac{(2J+1)}{4\pi}\frac{(J-m)!}{(J+m)!}}P_J^{m}(\cos\theta)\cdot\sin(|m|\phi)&m<0\\\\\sqrt{\frac{2J+1}{4\pi}}P_J^{m}(\cos\theta)&m=0\\\\\sqrt2\sqrt{\frac{(2J+1)}{4\pi}\frac{(J-m)!}{(J+m)!}}P_J^{m}(\cos\theta)\cdot\cos(m\phi)&m>0\end{cases}
\end{align}

This can be converted into the following function that returns the spherical harmonic given values of $J$, $m$, $\theta$ (theta), and $\phi$ (phi):

\begin{lstlisting}[language=Python]
def get_element(self, J, m, theta, phi):
    assert abs(m) <= J, "m must be in the range -J to J"
  	
    # calculates the first fraction in the square root
    N = np.sqrt((2*J+1) / (4*np.pi))
    # stores the ALP term in leg
    leg = self.lpmv(J, abs(m), torch.cos(theta))
    
    # multiply by the phi-dependent term depending on the value of m
    if m == 0: # when m=0 the other fraction in the square root cancels, and the phi-dependent term is 1
        return N*leg
    elif m > 0:
        Y = torch.cos(m*phi) * leg
    else:
        Y = torch.sin(abs(m)*phi) * leg
        
    # multiplies the square root of the inverse falling factorial, which is the same as in the ALP
    # sqrt(2) is included when m!=0 to account for splitting the complex harmonic into sine and cosine
    N *= np.sqrt(2. / falling_factorial(J, abs(m)))
    # multiplies the coefficient with the angle-dependent term
    Y *= N
    return Y
\end{lstlisting}

Now, we can generate the tensor of \textbf{spherical harmonics for all values of $m$ corresponding to a given $J$}, which is a type-$J$ spherical tensor:

\begin{lstlisting}[language=Python]
def get(self, J, theta, phi, refresh=True):
    # initialize tensor
    results = []
    
    # loop over all possible values of m from -J to J and add the computed spherical harmonic to results
    for m in range(-J, J+1):
        results.append(self.get_element(J, m, theta, phi))
    return torch.stack(results, -1)
\end{lstlisting}

The function below is called to calculate the spherical harmonics given the relative displacement vector between nodes in spherical coordinates, using the angle conventions from the implementation of the 3D steerable CNN \citep{weiler20183d}, where each edge is represented by the radius, beta (angle from south pole), and alpha (same as phi in spherical harmonics):

\begin{lstlisting}[language=Python]
# r_ij: the relative displacement between nodes in spherical coordinates [radius, alpha, beta]
# beta = pi - theta (beta is 0 at south pole and pi at north pole; supplementary to theta)
# alpha = phi (ranges from 0 to 2*pi)
# r_ij: shape (batch size, nodes, neighbors, 3 (r_ij))
def precompute_sh(r_ij, max_J):
    # initialize dictionary where keys correspond to J and values are tensors with shape (batch size, nodes, neighbors, 2J+1)
    Y_Js = {}
    # initialize an instance of the SphericalHarmonics object
    sh = SphericalHarmonics()
  		
    # calculate (2J+1)-dimensional spherical harmonics tensors for degrees up to max_J
    for J in range(max_J+1):
        # r_ij[...,2] extracts the values for beta for every edge in the graph 
        # r_ij[...,1] extracts the values alpha for every edge in the graph
        Y_Js[J] = sh.get(J, theta=math.pi-r_ij[...,2], phi=r_ij[...,1], refresh=False)
  
    sh.clear()
    return Y_Js
\end{lstlisting}

Finally, the basis kernels for all values of $J$ up to a maximum feature degree are computed and stored so that the kernels across every layer in the model transforming from type-$l$ to type-$k$ features can be derived simply by taking a linear combination of stored basis kernels for $J$ from $|k-l|$ to $k+l$.

The shape of the set of basis kernels for a given transformation between types $k$ and $l$ is $(1, 2l +1, 1, 2k+1, \text{num bases})$ where the singleton dimensions are \textbf{broadcast into the number of input and output channels} when taking the weighted sum to generate a unique kernel that transforms from a specific type-$k$ input channel to a specific type-$l$ output channel.

\begin{lstlisting}[language=Python]
# compute all spherical harmonics for every edge up to 2*maximum feature type
Y = utils_steerable.precompute_sh(x_ij, 2 * max_degree)
# define the device where the spherical harmonics are stored
device = Y[0].device

# initialize dictionary where the key is the input and output degree pair and the values are all the basis kernels stored in an array of shape (edges, 1, 2l+1, 1, 2k+1, 2min(l,k)+1)
basis = {}
# loop through all input and output degree pairs up to max_degree
for di in range(max_degree + 1):
    for do in range(max_degree + 1):
        K_Js = [] # initialize set of basis kernels
        # loop through all values of J from |k-l| to k+l
        for J in range(abs(di-do), di+do+1):
            # get change-of-basis matrices with shape ((2l+1)*(2k+1), 2J+1) that transforms the (2J+1)-dim spherical tensor back to its original basis
            Q_J = utils_steerable._basis_transformation_Q_J(J, di, do)
            Q_J = Q_J.float().to(device)
            # Y[J] has shape (edges, 2J+1)
            # Q_J has shape ((2l+1)*(2k+1), 2J+1)
            # matrix-vector multiplication to get K_J with shape (edges, (2l+1)*(2k+1)) of the vectorized type-J basis kernels
            K_J = torch.matmul(Q_J, Y[J])
            # append to list of bases with shape (2min(l,k)+1, edges, (2l+1)*(2k+1))
            K_Js.append(K_J)

        # reshape for dot product with radial weights
        size = (-1, 1, 2*do+1, 1, 2*di+1, 2*min(di, do)+1)
        # stack reshapes to (edges, (2l+1)*(2k+1), 2min(l,k)+1)
        # view reshapes to match size
        basis[f'{di},{do}'] = torch.stack(K_Js, -1).view(*size)
return basis
\end{lstlisting}

\subsection{Constructing the Radial Function}
\purple[]{
We can construct equivariant kernels by scaling the basis kernels with learned radial functions that transform the radial distance between nodes with a set of weights. This function is a feed-forward network (FFN) that is effective in learning complex dependencies between the basis kernels and the distance between nodes.
}

\begin{figure}[h!]
\centering
\includegraphics[width=\linewidth]{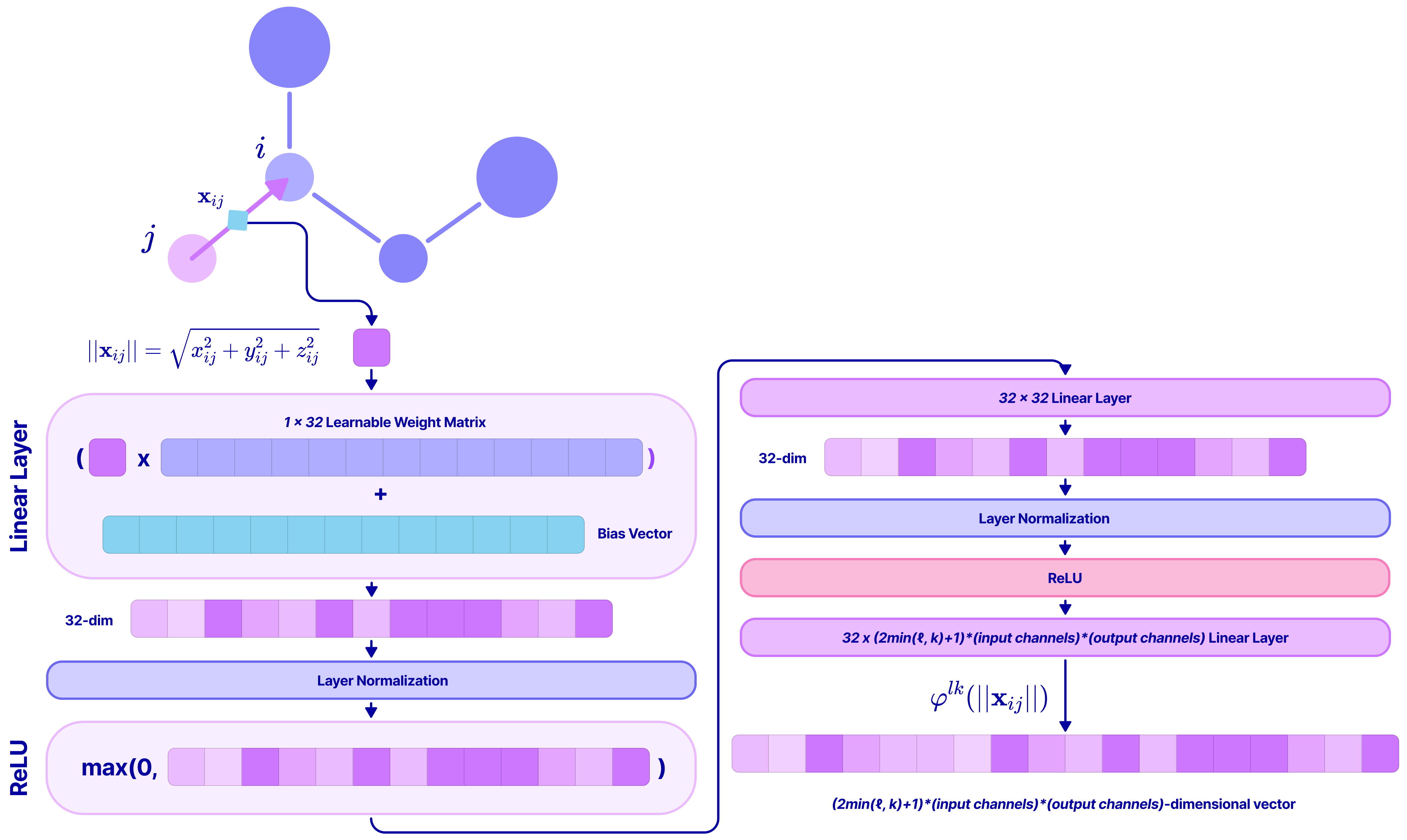}
\caption{The architecture of the feed-forward network (FFN) that takes the radial distance as input and outputs a set of (num bases)(input channels)(output channels)-dimensional vector of weights to scale the basis kernel to generate the transformation kernels from all input channels of a single type to all output channels of a single type.}
\label{fig:radial}
\end{figure}

Since each basis kernel represents a \textbf{unique function that changes with a defined pattern on the unit sphere} (defined by the spherical harmonic), we need to incorporate learnable parameters when constructing the equivariant kernels so that they can learn to detect specific feature motifs represented by distinct weighted combinations of the basis kernels.

The \textbf{radial function} not only allows us to incorporate distance dependence, which contributes to the strength of relationships between nodes, but it also allows the model to learn more complex dependencies on the node and edge features through backpropagation.

To construct the \textbf{kernel that transforms type-$k$ to type-$l$ features}, we define a radial function that takes the scalar distance as input and returns a weight for each basis kernel for values of $J$ from $|k-l|$ to $|k+l|$. For multi-channel features, the radial function not only generates a \textbf{unique weight} for each value of $J$ but also for every input-to-output channel pair. This means that the output of the function is a (num bases)(input channels)(output channels)-dimensional vector of weights.

\begin{align}
    \varphi^{l k}(\|\mathbf{x}_{ij}\|): \mathbb{R}_{\geq 0}\to \mathbb{R}^{(2min(l,k)+1)(\text{mi})(\text{mo})} 
\end{align}

We can denote a single weight produced by the function with the following notation:

\begin{align}
    \varphi^{l k}_{(J,c_l,c_k)}(\|\mathbf{x}_{ij}\|)\in\mathbb{R}
\end{align}

Since rotating a vector does not change its length, the radial distance is SO(3)-invariant. This means the output of the radial function is invariant to rotations of the input graph for \textit{any function definition}:

\begin{align}
    \varphi^{l k}_{(J,c_l,c_k)}(\|\mathbf{R}_g\mathbf{x}_{ij}\|)=\varphi^{l k}_{(J,c_l,c_k)}(\|\mathbf{x}_{ij}\|)
\end{align}

Thus, we define the radial function as a \textbf{feed-forward network (FFN) with multiple linear layers interspersed with nonlinear activation functions} to maximize the model’s ability to learn and detect complex feature motifs. The linear layers of an FFN can transform a vector from one dimension to another via multiplication with a (output dimension) $\times$ (input dimension) matrix of learnable weights and the addition of a bias vector with the same dimension as the output. Linear layers are followed by nonlinear activation functions like ReLU or LeakyReLU that capture more complex dependencies across weights.

\begin{align}
    \text{FFN}: \mathbb{R}_{\geq 0}\to\mathbb{R}^{(2min(l,k)+1)(\text{mi})(\text{mo})} 
\end{align}

A new network is constructed for every ordered pair $(k, l)$ of input and output feature types and every equivariant layer or type of embedding (attention layers have different networks for generating key and value embeddings). Each network is used across all nodes in the graph.

The FFN used in the SE(3)-Transformer consists of the following layers:

\begin{enumerate}
    \item The first linear layer transforms the input into a 32-dimensional vector by multiplication with a $32 \times 1$ learnable weight matrix and addition with a 32-dimensional bias vector.
    \begin{align}
        \mathbf{W}_1\|\mathbf{x}_{ij}\| + \mathbf{b}_1\tag{$\mathbf{W}_1\in \mathbb{R}^{32\times 1}$}
    \end{align}
    \item A layer normalization step transforms the mean to 0 and the variance to 1 across the values in the 32-dimensional vector.
    \item A non-linear ReLU activation function is applied element-wise. For each element, the ReLU function outputs 0 for negative values or itself for positive values.
    \begin{align}
        \mathbf{x}'=\max\left(0,\text{LayerNorm}(\mathbf{W}_1\|\mathbf{x}_{ij}\| + \mathbf{b}_1)\right)
    \end{align}
    \item Another FFN block with a linear layer (hidden dimension of 32), normalization step, and ReLU activation function is applied.
    \item A final linear layer transforms the 32-dimensional vector output of the second block into a ($2\min(l, k)+1$)(mi)(mo)-dimensional vector by multiplication with a $32\times$ ($2\min(l, k)+1$)(mi)(mo) matrix of learnable weights and addition with a ($2\min(l, k)+1$)(mi)(mo)-dimensional bias vector.
    \begin{align}
    \varphi^{l k}(\|\mathbf{x}_{ij}\|)=\mathbf{W}_2\mathbf{x}'' + \mathbf{b}_2\;\;\;\;(\mathbf{W}_2\in \mathbb{R}^{32\times (2min(l,k)+1)(\text{mi})(\text{mo})})
    \end{align}
    
\end{enumerate}
The FNN can be constructed in PyTorch with the following code:

\begin{lstlisting}[language=Python]
# num_bases = 2min(l,k)+1
# mid_dim = 32
self.net = nn.Sequential(
    # FFN transforms from 1 to mid_dim
    nn.Linear(1, self.mid_dim), 
    # layer norm over entire vector 
    BN(self.mid_dim), 
    # relu activation 
    nn.ReLU(), 
    # FFN that does not change dim
    nn.Linear(self.mid_dim, self.mid_dim), 
    # layer norm over entire vector
    BN(self.mid_dim), 
    # relu activation 
    nn.ReLU(), 
    # FFN transforms from mid_dim to (2min(l,k)+1)*mi*mo
    nn.Linear(self.mid_dim, self.num_bases * mi * mo))
\end{lstlisting}

Then, we can generate the radial weights for a given pair of input and output types $k$, $l$:

\begin{lstlisting}[language=Python]
def forward(self, x):
    # calculates a single vector of radial weights given the distance between nodes with the FFN
    y = self.net(x)
    # reshapes to separate the radial weights by output channel, input channel, and degree J to prepare for broadcasting and element-wise multiplication with the array of basis kernels of shape (-1, 1, 2l+1, 1, 2k+1, num_bases)
    return y.view(-1, self.mo, 1, self.mi, 1, self.num_bases)
\end{lstlisting}

\subsection{Mechanism of the Equivariant Kernel}
\purple[]{
To understand how the equivariant kernel can capture high-frequency rotationally symmetric patterns in a single matrix-vector multiplication step, we will break down how to interpret the mechanism of the kernel in terms of tensor products.
}

Given that each input-to-output channel pair has a unique radial weight, from now on, we will denote the kernel that transforms the type-$k$ feature at channel $c_k$ in node $j$ for message passing to the type-$l$ feature at channel $c_l$ in node $i$ as:

\begin{align}
    \mathbf{W}^{l k}_{(c_l,c_k)}(\mathbf{x})\mathbf{f}^k_{\text{in,}j,c_k}
\end{align}

The \textbf{weighted kernel} acts as an \textbf{aggregated function that takes multiple tensor products} between the type-$k$ input feature tensor and all the type-$J$ projections of the displacement vector.

\begin{figure}[h!]
    \centering
    \includegraphics[width=\linewidth]{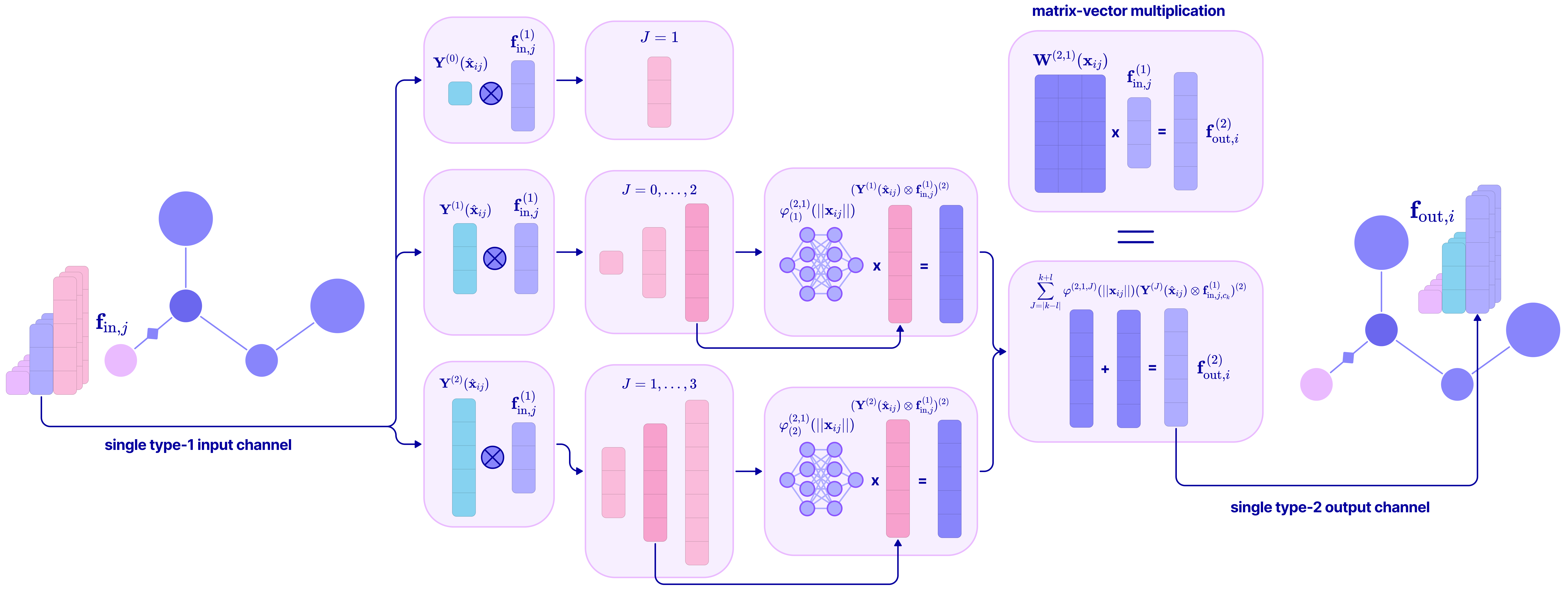}
    \caption{A visual breakdown of the mechanism of the equivariant kernel that transforms a single type-1 input channel to a single type-2 output channel. This mechanism involves taking the tensor product with all the type-$J$ projections from $J=0$ to 2, extracting the type-2 component from each tensor product decomposition, multiplying it with a weight generated from the radial function, and taking the sum of the resulting type-2 tensors. This mechanism is equivalent to multiplying the kernel $\mathbf{W}$ with the type-1 tensor to get a type-2 output tensor.}
    \label{fig:mechanism}
\end{figure}

Intuitively, we can think of the kernel transformation as performing the following operations:

\begin{enumerate}
    \item \textbf{Extracting the type-$l$ tensor component} of the tensor product between the type-$k$ feature tensor with the type-$J$ spherical tensor projection of the displacement vector via the Clebsch-Gordan coefficients. Each dimension (indexed by the magnetic quantum number $m_l$) of the type-$l$ tensor component can be written as:
    \begin{align}
        (\mathbf{Y}^{(J)}(\hat{\mathbf{x}}_{ij})\otimes\mathbf{f}^{k}_{\text{in,}j,c_k})^{(l)}_{m_l}=\sum_{m=-J}^J\sum_{m_k=-k}^kC^{(l, m_l)}_{(J, m)(k, m_k)}Y^{(J)}_{m}f^{(k)}_{m_k}
    \end{align}
    We can write the full type-$l$ tensor as a $(2l + 1)$-dimensional vector by concatenating all dimensions indexed by $m$ from $-l $ to $l $ :
    \begin{align}
        (\mathbf{Y}^{(J)}(\hat{\mathbf{x}}_{ij})\otimes\mathbf{f}^{k}_{\text{in,}j,c_k})^{(l)}=\begin{bmatrix}(\mathbf{Y}^{(J)}(\hat{\mathbf{x}}_{ij})\otimes\mathbf{f}^{k}_{\text{in,}j,c_k})^{(l)}_{-l} \\\\ (\mathbf{Y}^{(J)}(\hat{\mathbf{x}}_{ij})\otimes\mathbf{f}^{k}_{\text{in,}j,c_k})^{(l)}_{-l+1} \\ \vdots \\(\mathbf{Y}^{(J)}(\hat{\mathbf{x}}_{ij})\otimes\mathbf{f}^{k}_{\text{in,}j,c_k})^{(l)}_{l} \end{bmatrix}
    \end{align}

    \item \textbf{Scaling the type-$l$ tensor} component by the weight calculated by a learnable function on the radial component of the displacement vector.
    \begin{align}
        \varphi^{l k}_{(J,c_l,c_k)}(\|\mathbf{x}_{ij}\|)(\mathbf{Y}^{(J)}(\hat{\mathbf{x}}_{ij})\otimes\mathbf{f}^{k}_{\text{in,}j,c_k})^{(l)}
    \end{align}
    \item Repeating Steps 1 and 2 for \textit{all} types-$J$ projections of the angular unit vector ranging from $|k  l|$ to $|k + l|$ and \textbf{scaling the type-$l$ output by a unique learnable weight}. Then, taking the \textbf{sum} of all the weighted type-$l$ components of the tensor products to get the \textbf{output type-$l$ message for channel $c_l$ from the type-$k$ feature at channel $c_k$:}
    \begin{align}
        \mathbf{W}^{l k}_{(c_l,c_k)}(\mathbf{x}_{ij})\mathbf{f}^k_{\text{in,}j,c_k}=\sum_{J=|k-l|}^{k+l}\varphi^{l k}_{(J,c_l,c_k)}(\|\mathbf{x}_{ij}\|)(\mathbf{Y}^{(J)}(\hat{\mathbf{x}}_{ij})\otimes\mathbf{f}^k_{\text{in,}j,c_k})^{(l)}
    \end{align}
\end{enumerate}

Instead of performing all these steps and taking the tensor product with every type-$J$ projection, the \textbf{equivariant kernel aggregates all of these operations into a single kernel matrix-vector multiplication step.}

\subsection{Rules of Equivariant Layers}
\label{subsec:rules-equiv}
Now, let’s quickly solidify some general rules when constructing equivariant layers:
\begin{enumerate}
    \item To transform between spherical tensors of different types, we must take the tensor product, which generates a higher-dimensional tensor that can be decomposed into its spherical tensor components. Then, we can extract the target feature type from the decomposition to combine it with other tensors of the same type. This process can be condensed into a single matrix-vector multiplication step with an equivariant kernel that combines \textbf{spherical harmonics and Clebsch-Gordan coefficients such that it satisfies the kernel constraint.}
    
    \item Applying non-linear functions element-wise to spherical tensors with degrees greater than 0 breaks equivariance. To introduce non-linearities, we can incorporate nonlinearities into learnable functions that transform scalar or higher-dimensional spherical tensors into scalar weights that can be multiplied across all elements of a spherical tensor.
    
    \item All group representations, including Wigner-D matrices, are \textbf{linear transformations} from a tensor space to itself that act exclusively on a \textit{type} of spherical tensor. This means that Wigner-D matrices \textbf{preserve addition and scalar multiplication of tensors with the same type}, such that we can combine multiple features across channels of the same type by taking weighted sums.
    \begin{align}
        \mathbf{D}_l(g)(a\mathbf{s}^l+b\mathbf{t}^l)=a\mathbf{D}_l(g)\mathbf{s}^l+b\mathbf{D}_l(g)\mathbf{t}^l\tag{$a,b\in \mathbb{R}$}
    \end{align}
    This makes intuitive sense when considering scaling and adding vectors in 3D space. Rotating the sum of two 3D vectors is equivalent to rotating the two vectors and taking their sum. This generalizes to higher-degree spherical tensors, since by definition, tensors of the same type transform under the same set of \textit{irreps} that \textbf{preserve lengths and angles between tensors}.
    
    \item Since graphs are considered unordered sets of nodes, operators used to aggregate messages must be \textbf{permutation-invariant} such that the order in which the messages are aggregated does not change the output. The operators must also be \textbf{equivariant to rotation}, which includes taking the weighted sum or average (which is just the sum scaled by the fraction with the total number of messages in the denominator) of messages of the same type.
\end{enumerate}

With these rules in mind, let’s break down how to build one of the first fully SO(3)-equivariant modules for geometric graphs: the \textbf{Tensor Field Network (TFN)} \citep{thomas2018tensor}.

\newpage

\section{Tensor Field Network Module}
\label{sec:4}
\purple[]{
The \textbf{Tensor Field Network (TFN)} introduced by \citet{thomas2018tensor} is an SE(3)-equivariant model that takes a point cloud and applies an equivariant kernel that outputs a geometric tensor at every point in space (definition of a tensor field) using spherical harmonics and radial weights. The TFN layer in the SE(3)-Transformer is used to reduce high-degree attention embeddings into lower-degree tensors before generating a prediction.
}

\begin{figure}[h!]
    \centering
    \includegraphics[width=\linewidth]{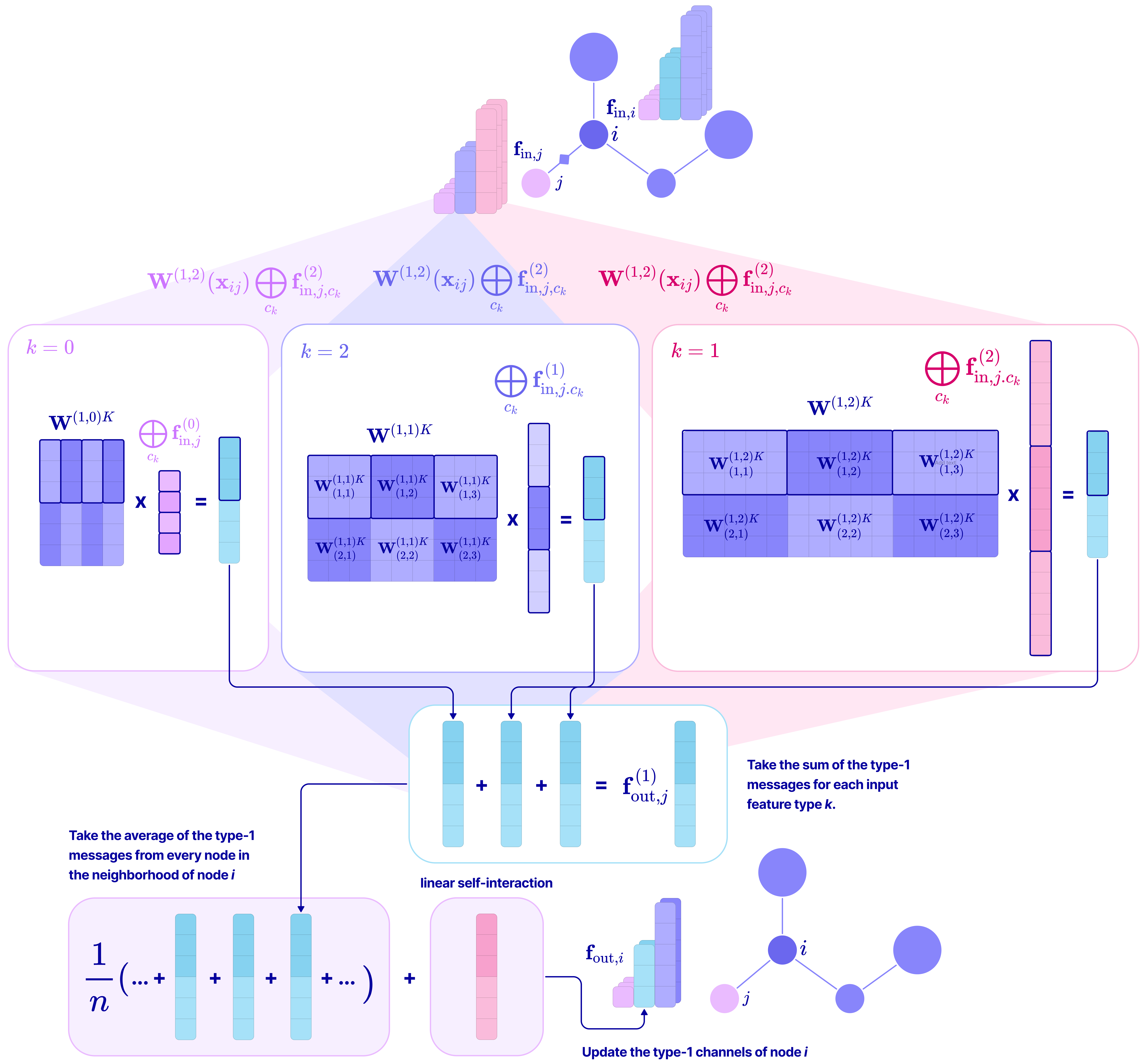}
    \caption{A visual breakdown of how the TFN layer generates the updated type-1 output features that are used to update the center node $i$. For each input degree, the direct sum of all the channels is multiplied by a combined equivariant kernel to generate a message for each type-$l$ output channel. Then, we take the sum across the type-$l$ messages from each input feature type to get the type-$l$ message from a single neighborhood node $j$. Next, we reduce the type-$l$ messages from all nodes in the neighborhood of $i$ by taking the average. Finally, we add a linear self-interaction term to get the updated type-$l$ feature tensor at node $i$ that is outputted by the TFN layer.}
    \label{fig:tfn}
\end{figure}

A Tensor Field Network (TFN) layer generates a message for every channel of every feature type for every node in the graph with the equation below, which is the focus of this section:

\begin{align}
\mathbf{f}^l_{\text{out},i,c_l}=\underbrace{\sum_{c_l'}w^{l l}_{(c_l,c_l')}\mathbf{f}^l_{\text{in},i,c_l'}}_{\text{linear self-interaction}}+\underbrace{
    \frac{1}{n} \sum_{j \neq i}^n 
    \underbrace{
        \sum_{k \geq 0} 
        \underbrace{
            \sum_{c_k} 
            \underbrace{
                \mathbf{W}^{l k}_{(c_l,c_k)}(\mathbf{x}_{ij}) \mathbf{f}^k_{\text{in},j,c_k}
            }_{\text{channels }c_k\to c_l\text{ message}}
        }_{\text{message from type-}k\text{ input channels}}}_{\text{type-}l\text{ message from node }j}
}_{\text{average message over all } n \text{ adjacent nodes } j}
\end{align}

This message generated above is used to update the type-$l$ feature at channel $c_l$ at node $i$, denoted $\mathbf{f}^l_{\text{in},i,c_l}\mapsto\mathbf{f}^l_{\text{out},i,c_l}$.

\subsection{Equivariant Message-Passing}

Since we have already described the function of the equivariant kernel, this section will focus on breaking down how the messages are aggregated and implementing it in code.

For every node in the graph, the \textbf{TFN layer updates the input feature tensor with an updated output feature tensor generated equivariantly from the input features of adjacent nodes}. This process involves the following steps:

\begin{enumerate}
    \item First, we construct a \textbf{new radial network for all $k \to l $ transformations} with an output dimension of $(2\min(l,k)+1)(\text{type-$k$ input channels, mi})(\text{type-$l$ output channels, mo})$ which are used to calculate unique radial weights for every basis kernel and every possible path from a type-$k$ input channel to a type-$l$ output channel.
\begin{lstlisting}[language=Python]
# initialize the radial network for type-$k$ inputs and type-l outputs 
# the value of edge_dim has a default value of 1 which determines the dimension of the input to the radial function
self.rp = RadialFunc(self.num_bases, mi, mo, self.edge_dim)
\end{lstlisting}
    
    \item The radial distance corresponding to every edge is fed into the $(k, l )$ radial function to obtain an array of \textbf{radial weights} with shape (mo, 1, mi, 1, num bases). We will discuss how other scalar edge features can be incorporated in this step in Section \ref{subsec:edge-features}.
\begin{lstlisting}[language=Python]
# calls forward method of RadialFunc class which feeds the relative distance into radial network
R = self.rp(feat)
\end{lstlisting}
    
    \item For a given edge, we obtain the \textbf{kernel} of the transformation of features from an input type-$k$ channel $c_k$ to an output type-$l$ channel $c_l$ by taking a \textbf{weighted sum of all basis kernels scaled by their corresponding radial weight}.
    \begin{align}
        \mathbf{W}^{l k}_{(c_l,c_k)}(\|\mathbf{x}_{ij}\|)=\sum_{J=|k-l|}^{k+l}\varphi^{l k}_{(J,c_l,c_k)}(\|\mathbf{x}_{ij}\|)\mathbf{W}^{l k}_{J}(\|\mathbf{x}_{ij}\|)
    \end{align}
    In the code implementation, the singleton dimensions of the array of radial weights with shape (mo, 1, mi, num bases) and the array of basis kernels with shape (1, $2l+1$, 1, $2k+1$, num bases) are broadcast to match each other, and the element-wise product of the two tensors is calculated. The resulting product is a set of weighted basis kernels for every input and output channel pair with shape $(mo, 2l + 1, mi, 2k + 1, num\_bases)$. Finally, we take the sum over the weighted basis kernels (last dimension) to get an array with shape $(mo, 2l + 1, mi, 2k + 1)$, containing the final kernel for every pair of channels $(c_l, c_k)$.
\begin{lstlisting}[language=Python]
# R: radial weights with shape (batch size, mo, 1, mi, 1, 2min(di, do)+1)
# basis[f'{self.di},{self.do}']: tensor of basis kernels for input deg di and output deg do
kernel = torch.sum(R * basis[f'{self.di},{self.do}'], -1)
\end{lstlisting}
    
    \item The kernel array is \textbf{reshaped} to have dimensions $(\text{mo})(2l + 1) \times (\text{mi})(2k + 1)$, where each block of the matrix is the $(2l + 1) \times (2k + 1)$ kernel corresponding to the transformation from input channel $c_k$ to output channel $c_l$, denoted by the subscript $(c_l, c_k)$.
    \begin{align}
        \mathbf{W}^{l k}(\mathbf{x}_{ij})=\begin{bmatrix}\mathbf{W}^{l k}_{(1,1)}&\mathbf{W}^{l k}_{(1,2)}&\dots&\mathbf{W}^{l k}_{(1,\text{mi})}\\\\\mathbf{W}^{l k}_{(2,1)}&\ddots&\dots &\vdots \\\\\vdots &\dots &\ddots& \vdots\\\\\mathbf{W}^{l k}_{(\text{mo},1)}&\dots &\dots &\mathbf{W}^{l k}_{(\text{mo},\text{mi})}\end{bmatrix}_{(\text{mo}))(2l+1)\times(\text{mi})(2k+1)}
    \end{align}
\begin{lstlisting}[language=Python]
# reshape kernel to (mo*(2*do+1), mi*(2*di+1)) to prepare for matrix-vector multiplication with the concatenated input channels of type di
return kernel.view(kernel.shape[0], (2*self.do+1) * self.out_channels, -1)
\end{lstlisting}
    
    \item Now we calculate and store a kernel for every $k \to  l$ transformation by looping through all (multiplicity, degree) tuples in the input and output fiber structures. The code for generating the kernel is implemented in the \texttt{PairwiseConv} class.

\begin{lstlisting}[language=Python]
# loop over (multiplicity, degree) tuples in input fiber
for (mi, di) in self.f_in.structure:
    # loop over (multiplicity, degree) tuples in output fiber
    for (mo, do) in self.f_out.structure:
        # generate a (mi * mo) unique kernels corresponding to every input and output channel pair
        # store in dictionary with key f'({di}{do})'
        self.kernel_unary[f'({di},{do})'] = PairwiseConv(di, mi, do, mo, edge_dim=edge_dim)
\end{lstlisting}
    
    \item For every feature type in the output fiber, we can compute the message for a single type-$l$ output channel $c_l$ at the center node $i$ from all the features at node $j$ by \textbf{(1)} transforming the type-$k$ input channel $c_k$ into a type-$l$ message by matrix multiplication with the uniquely defined kernel for output channel $c_l$, \textbf{(2)} taking the sum over channel $c_l$ messages from all type-$k$ input channels, and \textbf{(3)} iterating for all feature types $k$ defined in the input fiber and adding the type-$l$ message from type-$k$ input channels to the total type-$l$ message with each iteration.
    
    \begin{align}
        \underbrace{
        \sum_{k \geq 0} 
        \underbrace{
            \sum_{c_k} 
            \underbrace{
                \mathbf{W}^{l k}_{(c_l,c_k)}(\mathbf{x}_{ij}) \mathbf{f}^k_{\text{in},j,c_k}
            }_{\text{(a) channels }c_k\to c_l\text{ message}}
        }_{\text{(b) message from type-}k\text{ input channels}}}_{\text{(c) type-}l\text{ channel }c_l\text{ message from node }j}
    \end{align}
    In the code implementation, the messages for all type-$l$ channels from a single input type $k$ are calculated in parallel with a single matrix-vector multiplication step by concatenating all the type-$k$ input features into a $(\text{mi})(2k+1)$-dimensional vector and multiplying it with the $(\text{mo})(2l + 1) \times (\text{mi})(2k + 1)$ block kernel defined above. The output of the kernel contains the messages for each type-$l$ output channel stacked in a single $(\text{mo})(2l+1)$-dimensional vector, which is then summed together with the messages from each input degree $k$.
    \begin{align}
        \underbrace{\sum_{k\geq0}\underbrace{\mathbf{W}^{l k}(\mathbf{x}_{ij})\bigoplus_{c_k}\mathbf{f}^k_{\text{in},j,c_k}}_{\text{message from type-}k\text{ input channels}}}_{\text{direct sum of type-}l\text{ messages from node }j}
    \end{align}
    The code implementation calculates the type-$l$ messages for every edge in the graph in parallel and reshapes the vectorized output messages into an array with shape (edges, mo, $2l+1$).
    
\begin{lstlisting}[language=Python]
# calculate neighbor -> center messages for type single output feature type do
msg = 0
# loop over input (multiplicity, degree) pairs 
for mi, di in self.f_in.structure:
    # extract all feature channels of type di from the neighborhood nodes and condense into a single vector
    # src has shape (edges, mi*(2*di+1), 1)
    src = edges.src[f'{di}'].view(-1, mi*(2*di+1), 1) 
    
    # extract kernel for input type di and output type do
    edge = edges.data[f'({di},{do})']
    # matrix multiplication to get (mo*(2*do+1))-dimensional vector and add to total msg
    msg = msg + torch.matmul(edge, src)
# reshape message to separate output channels 
msg = msg.view(msg.shape[0], -1, 2*do+1)
\end{lstlisting}
\end{enumerate}

We repeat step 6 for every output feature type and \textbf{every directional edge in the graph}. The neighbor-to-center message for a single output type is stored as an array with shape (edges, mo, $2l+1$) and modified with a \textbf{self-interaction step}, which we will break down in the next two sections. 

\begin{align}
    \text{Self-Interaction}\left(\sum_{k\geq0}\mathbf{W}^{l k}(\mathbf{x}_{ij})\bigoplus_{c_k}\mathbf{f}^k_{\text{in},j,c_k}\right)
\end{align}

Since each of the $n$ edges (from every neighborhood node) pointing to the center node $i$ generates a mo $\times  (2l+1)$ type-$l$ message, we need to apply a \textbf{permutation-invariant reduction operation} to reduce the messages across edges into a single output type-$l$ tensor. We achieve this by taking the \textbf{mean across the type-$l$ messages from all $n$ incoming edges}, which becomes the final type-$l$ neighbor-to-center message to node $i$. 

\begin{align}
    \mathbf{f}^l_{\text{out},i}=\frac{1}{n}\sum_{j\neq i}^n\left(\text{Self-Interaction}\left(\sum_{k\geq0}\mathbf{W}^{l k}(\mathbf{x}_{ij})\bigoplus_{c_k}\mathbf{f}^k_{\text{in},j,c_k}\right)\right)
\end{align}

In the next two sections, we will break down the two types of self-interaction that can be applied to the neighbor-to-center message before aggregation: \textbf{linear self-interaction} (Section \ref{subsec:linear-self-interact}) and \textbf{channel-mixing} (Section \ref{subsec:channel-mixing}). At the end of the TFN section, we will put all the components together in a user-defined function and see how to generate an updated graph representation with the output messages.

\subsection{Linear Self-Interaction}
\label{subsec:linear-self-interact}
The output of the TFN layer also contains a \textbf{linear self-interaction term added to the message-passing term} that incorporates the features from the node itself and acts as an \textbf{equivariant skip connection} or residual connection.

\begin{wrapfigure}{r}{0.5\linewidth} % or {l} instead of {r}
\centering
\includegraphics[width=\linewidth]{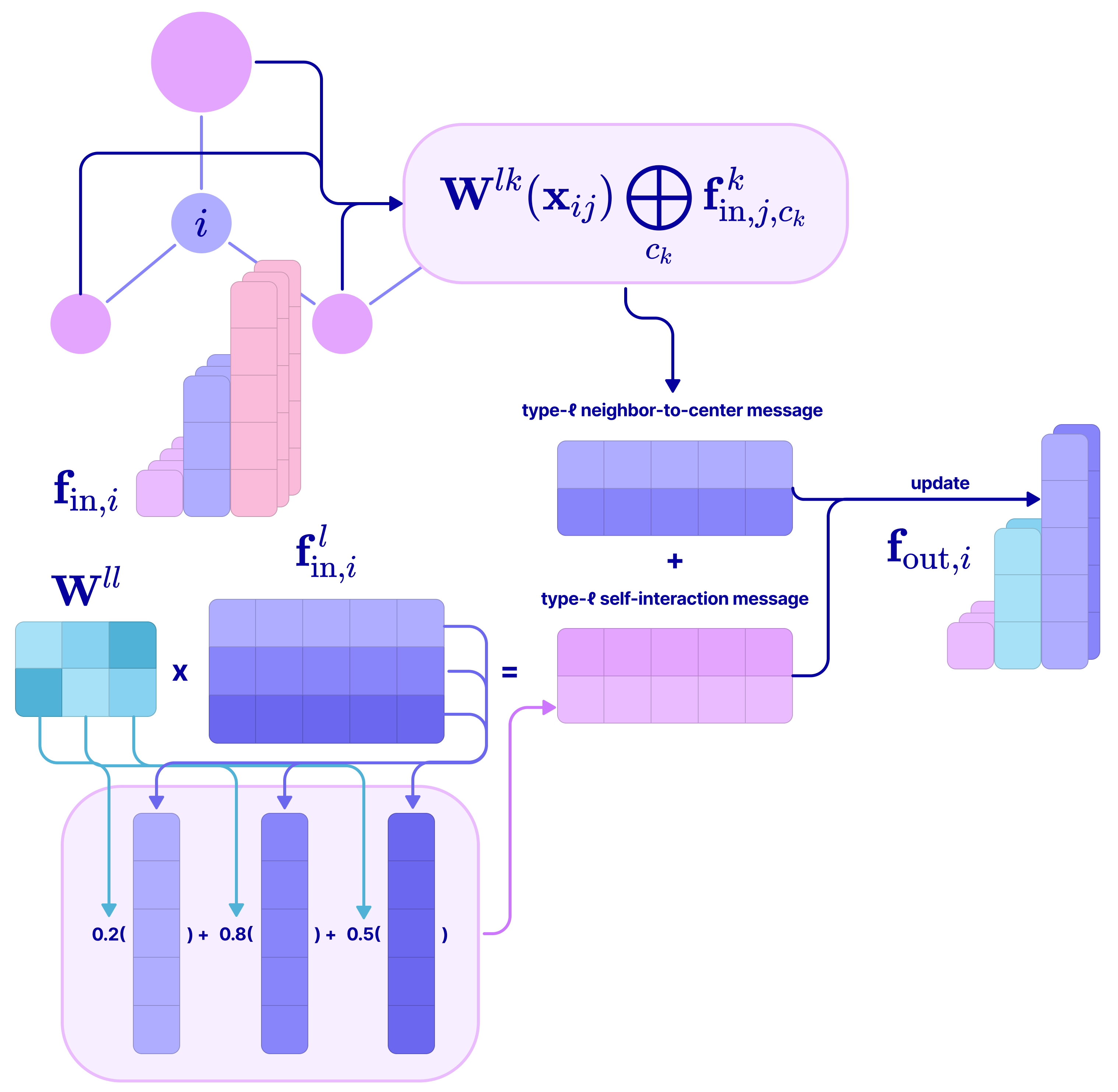}
\caption{For each input feature type $l$, linear self-interaction takes the linear combination of type-$l$ input channels to generate a matrix of self-interaction messages that matches the number of type-$l$ output channels. The matrix of type-$l$ self-interaction messages is then added to the matrix of type-$l$ neighbor-to-center messages to get the output type-$l$ feature tensor used to update the center node.}
\label{fig:linear-self-interact}
\end{wrapfigure}

\textbf{Skip connections} in deep learning are a technique used to prevent vanishing gradients by adding or concatenating input data to the transformed output data between layers, which allows the gradient to skip past gradient-diminishing layer operations during backpropagation. In the TFN module, \textbf{linear self-interaction introduces a skip connection by adding linear combinations of a node’s input features into the output feature tensor}. Since the neighbor-to-center message is dominated by features from adjacent nodes, this step also prevents the independent properties of each node from being lost after each feature update.

We cannot transform between different feature types in a single node without a displacement vector, and directly combining different types breaks equivariance. To preserve equivariance, \textbf{linear self-interaction} takes the learnable weighted sum of \textbf{input channels with the same degree} and adds it to the output neighbor-to-center message of that degree.

\begin{align}
    \sum_{c_l'}w^{l l }_{(c_l,c_l')}\mathbf{f}^l_{\text{in},i,c_l'}
\end{align}

This expression calculates a type-$l$ self-interaction spherical tensor for the type-$l$ output channel $c_l$ of node $i$ by taking the \textbf{weighted sum of all the type-$l$ input channels} (denoted by $c_l’$) using a set of learnable weights.

Each unique weight is denoted by a subscript $(c_l, c_l’)$ that indicates the input channel $c_l’$ that is being scaled by the weight and the output channel $c_l$ for which the self-interaction is being calculated, and this weight is used \textbf{across every node in the graph}.

The self-interaction weights for all type-$l$ channels are stored in a matrix with dimensions (output type-$l$ channels, mo) $\times$ (input type-$l$ channels, mi), where each row corresponds to all the weights used to calculate self-interaction for a single output channel.

\begin{align}
    \mathbf{W}^{l l}=\begin{bmatrix}w^{l l}_{(1,1)}&w^{l l}_{(1,2)}&\dots&w^{l l}_{(1,\text{mi})}\\\\w^{l l}_{(2,1)}&\ddots&&\vdots\\\\\vdots&&\ddots&\vdots\\\\w^{l l}_{(\text{mo},1)}&\dots&\dots&w^{l l}_{(\text{mo,mi})}\end{bmatrix}_{\text{mo}\times\text{mi}}
\end{align}

The implementation of self-interaction proceeds as follows:

\begin{enumerate}
    \item For each output feature type, a \textbf{weight array} of shape (1, mo, mi) is initialized with random integers and scaled down by the square root of the number of input channels to reduce the variance of outputs.
\begin{lstlisting}[language=Python]
# loop through all input types
for mi, di in self.f_in.structure:
    # proceed if input type is also an output type
    if di in self.f_out.degrees: 
        # extract num of output channels of the type
        m_out = self.f_out.structure_dict[di]
        # initialize learnable mi x mo weight matrix with random integers and scale down
        # singleton dimension used to broadcast across nodes
        W = nn.Parameter(torch.randn(1, mo, mi) / np.sqrt(mi))
        self.kernel_self[f'{di}'] = W
\end{lstlisting}
    \item Then, we generate the \textbf{self-interaction tensor} for all type-$l$ output features at node $i$ by multiplying the weight array with the matrix formed by stacking each type-$l$ input feature as rows.
    \begin{align}
        \mathbf{W}^{l l}\mathbf{f}^l_{\text{in},i}
    \end{align}
    \begin{figure}[h!]
        \centering
        \includegraphics[width=0.6\linewidth]{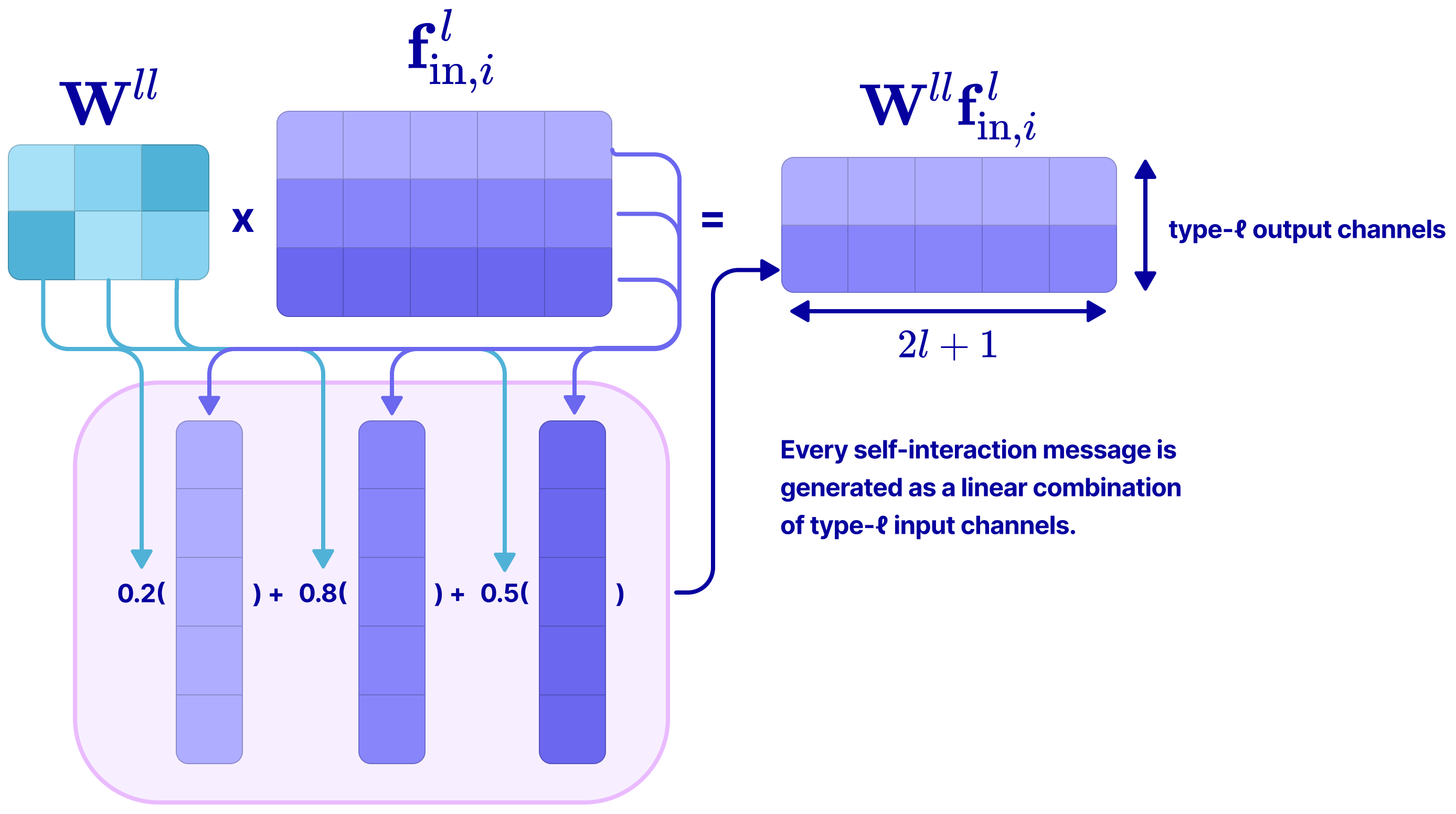}
        \caption{Applying the linear self-interaction weight matrix to the matrix of all type-$l$ input channels stacked into the rows outputs the matrix of self-interaction messages for all type-$l$ output channels. Each row of the output matrix is equivalent to taking the sum of all the input channels scaled by the weights in the corresponding row of the weight matrix.}
        \label{fig:self-interact2}
    \end{figure}
    
    This operation condenses the process of calculating the weighted sum for each type-$l$ channel into a single matrix multiplication step that calculates self-interaction for all type-$l$ channels simultaneously.
    
    In the code implementation, this step is performed for every node in parallel by multiplying the weight array with shape (1, mo, mi) with the type-$l$ input feature array with shape (nodes, mi, $2l+1$). Based on broadcasting rules, this will produce a tensor with shape (nodes, mo, $2l+1$).
\begin{lstlisting}[language=Python]
    # extract all input features of type do from all nodes
    dst = edges.dst[f'{do}']
    # extract self-interaction weights for type do channels
    W = self.kernel_self[f'{do}']
    # calculate the array of self-interaction tensors with shape (nodes, mo, 2*do+1)
    self_int = torch.matmul(W, dst)
\end{lstlisting}
    
    \item Lastly, the self-interaction array is added to the neighbor-to-center message with the same shape (nodes, mo, $2l+1$) for every output degree.
    \begin{lstlisting}[language=Python]
    # add to neighbor-to-center message
    msg = msg + self_int
    \end{lstlisting}
\end{enumerate}

Since the \textbf{same linear self-interaction message is added to the neighbor-to-center message from every incoming edge to the center node $i$}, we can write the full TFN layer with the equation below:

\begin{align}
    \mathbf{f}^l_{\text{out},i}=\underbrace{\mathbf{W}^{l l}\mathbf{f}^l_{\text{in},i}}_{\text{linear self-interaction}}+\underbrace{\frac{1}{n}\sum_{j\neq i}^n\sum_{k\geq0}\mathbf{W}^{l k}(\mathbf{x}_{ij})\bigoplus_{c_k}\mathbf{f}^k_{\text{in},j,c_k}}_{\text{neighbor }\to\text{ center message}}
\end{align}

We will also revisit the linear self-interaction mechanism in Section \ref{subsec:query} when we discuss how to generate the query embeddings for self-attention.

\subsection{Channel Mixing}
\label{subsec:channel-mixing}
Another way to incorporate equivariant self-interaction in a TFN layer is through \textbf{channel mixing}. Since a unique neighbor-to-center message is generated for every channel of every type, instead of simply updating each channel with its corresponding message, channel mixing updates each channel with the \textbf{weighted sum of the neighbor-to-center messages for all other type-$l$ output channels of the same type}.

\begin{align}
    \mathbf{f}^l_{\text{out},i,c_l}=\sum_{c_l'}w^{l l}_{(c_l,c_l')}\mathbf{f}^l_{\text{out},i,c_l'}
\end{align}

where the output feature for channel $c_l’$ inside the summation refers to the neighbor-to-center message before applying channel mixing.

\begin{figure}
\centering
\includegraphics[width=0.8\linewidth]{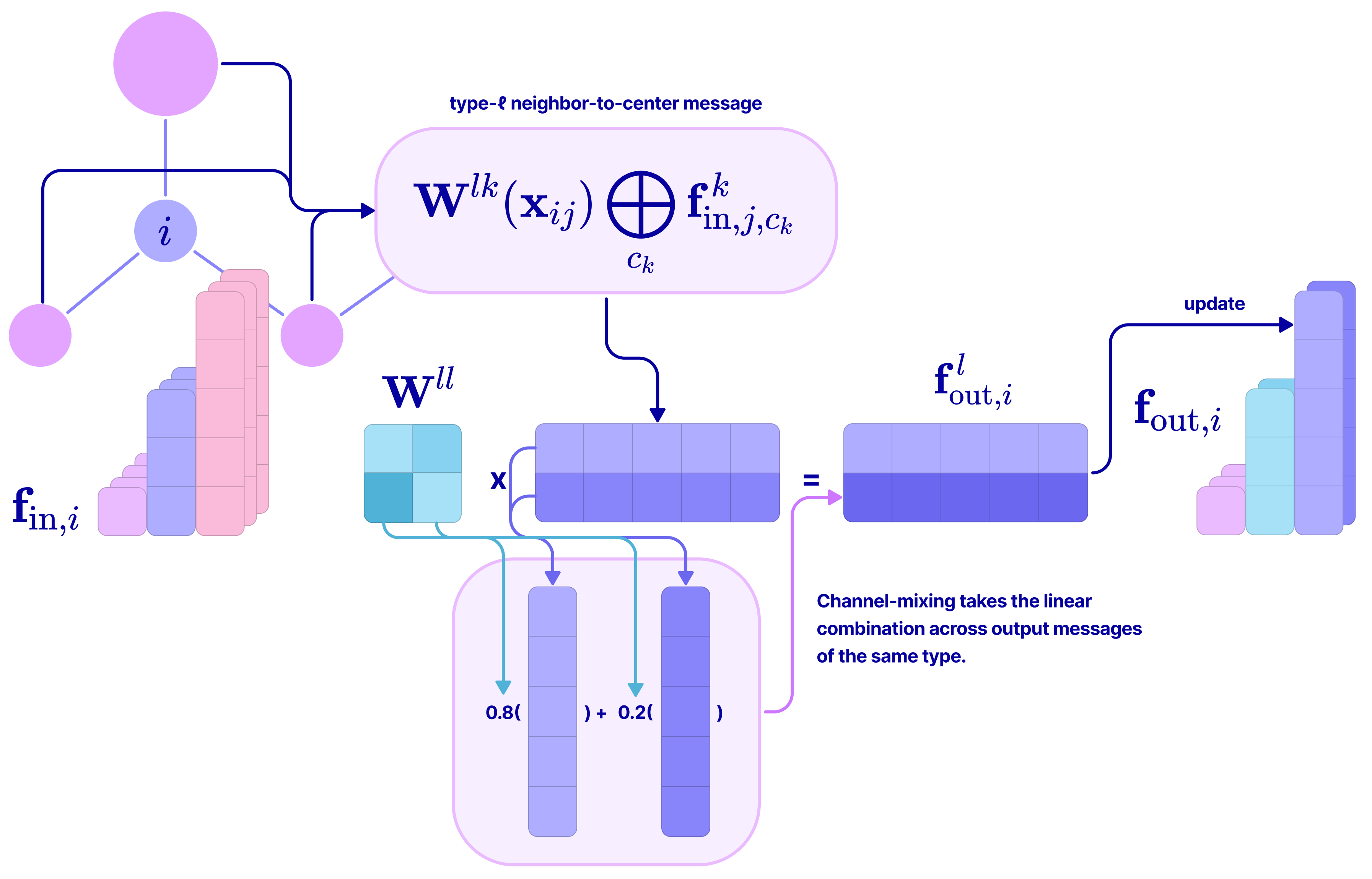}
\caption{For each output type $l$, channel mixing multiplies a matrix of weights to the matrix of type-$l$ neighbor-to-center messages stacked into rows to get the output type-$l$ feature tensor used to update the center node.}
\label{fig:channel-mixing}
\end{figure}

For every output type, there is a corresponding square matrix of learnable weights with dimensions $\text{mi} \times  \text{mo}$, where each row corresponds to the weights used to perform channel mixing for updates to a single output channel $c_l$.

\begin{align}
    \mathbf{W}^{l l}=\begin{bmatrix}w^{l l}_{(1,1)}&w^{l l}_{(1,2)}&\dots&w^{l l}_{(1,\text{mo})}\\\\w^{l l}_{(2,1)}&\ddots&&\vdots\\\\\vdots&&\ddots&\vdots\\\\w^{l l}_{(\text{mo},1)}&\dots&\dots&w^{l l}_{(\text{mo,mo})}\end{bmatrix}_{\text{mo}\times\text{mo}}
\end{align}

Instead of multiplying by the input type-$l$ channels like in linear self-interaction, channel-mixing is implemented by \textbf{multiplying the weight matrix with the matrix formed by the output type-$l$ channels as rows}, which outputs a matrix of type-$l$ output messages from node $j$ where each channel is a ‘\textit{mix}’ of the other channels.

\begin{align}
    \mathbf{f}^l_{\text{out},i}=\underbrace{\frac{1}{n}\sum_{j\neq i}^n\underbrace{\mathbf{W}^{l l}\underbrace{\text{unvec}\left(\sum_{k\geq0}\mathbf{W}^{l k}(\mathbf{x}_{ij})\bigoplus_{c_k}\mathbf{f}^k_{\text{in},j,c_k}\right)}_{\text{matrix of output type-}l\text{ messages}}}_{\text{channel mixing}}}_{\text{average of all type-}l\text{ messages from node }j}
\end{align}

In the code implementation, the weight array of shape (1, mo, mo) is broadcast and multiplied by the array of all output type-$l$ tensors generated in the message-passing step of shape (nodes, mo, $2l+1$) to produce the \textbf{mixed type-$l$ output message with the same shape}.

\begin{lstlisting}[language=Python]
# in constructor
# loop over all output types
for mo, do in self.f_out.structure:
    # initialize square learnable weight matrix of random integers and scale down
    W = nn.Parameter(torch.randn(1, mo, mo) / np.sqrt(m_out))
    self.kernel_self[f'{do}'] = W

# in user-defined DGL edge -> node function
# extract weight array of shape (1, mo, mo)
W = self.kernel_self[f'{do}']
# matrix multiplication that generates output feature tensor for  degree do
msg = torch.matmul(W, msg)
\end{lstlisting}

To encapsulate the code for the neighbor-to-center message and self-interaction message, we define an edge UDF in DGL called \texttt{udf\_u\_mul\_e} that computes the messages for a single output feature type for each edge with either linear self-interaction or channel mixing. The messages are stored in a dictionary where the label ‘\texttt{msg}’ is linked to an array with shape (edges, type-$l$ output channels, $2l+1$). 

\begin{lstlisting}[language=Python]
# edge user-defined function in DGL that computes the messages for a single output feature type do 
def udf_u_mul_e(self, do):
    # takes the batch of edges in the graph
    def fnc(edges):
        # neighbor -> center messages
        msg = 0
        for mi, di in self.f_in.structure:
            src = edges.src[f'{di}'].view(-1, mi*(2*di+1), 1)
            edge = edges.data[f'({di},{do})']
            msg = msg + torch.matmul(edge, src)
        msg = msg.view(msg.shape[0], -1, 2*d_out+1)

        # self-interaction message
        if self.self_interaction:
            # checks if there is a self-interaction weight matrix for output degree
            if f'{do}' in self.kernel_self.keys():
                # channel mixing
                if self.flavor == 'TFN':
                    W = self.kernel_self[f'{do}']
                    # multiply weight matrix with output msg
                    msg = torch.matmul(W, msg)
                # linear self-interaction
                if self.flavor == 'skip':
                    # get the input features from the center node
                    dst = edges.dst[f'{di}']
                    # get the self-int weight matrix 
                    W = self.kernel_self[f'{do}']
                    # add the neighbor -> center message to the linear combinations of input type do features
                    msg = msg + torch.matmul(W, dst)
            # return a dictionary where an array with shape (edges, output channels, 2*do+1) with the label 'msg'
            return {'msg': msg.view(msg.shape[0], -1, 2*do+1)}
    # returns the edge -> node function handle that takes the edges in the graph and returns a dictionary of messages for every edge
    return fnc
\end{lstlisting}

Then, we can loop through every output degree type and call the built-in \texttt{update\_all} function in DGL which executes the \texttt{udf\_u\_mul\_e} function and reduces the output messages using the built-in mean function that extracts the array labeled ‘\texttt{msg}’ generated from the \texttt{udf\_u\_mul\_e} function, takes the average of the messages of a given type across all edges that share a destination node, and stores them in the node features of the graph with the label \texttt{f`out\{do\}`}.

\begin{lstlisting}[language=Python]
# loop through all output feature types
for do in self.f_out.degrees:
    # call update_all function that takes (message_func, reduce_func as input
    G.update_all(self.udf_u_mul_e(do), fn.mean('msg', f'out{do}'))

# return a dictionary of the output node features where every degree is linked to an array with shape (edges, mo, 2*do+1) by extracting the node data stored from calling update_all
return {f'{do}': G.ndata[f'out{do}'] for do in self.f_out.degrees}
\end{lstlisting}

\textbf{Passing the graph through the full TFN module generates a fully updated graph representation that integrates neighbor-to-center and self-interaction messages.}

Now that we understand how equivariant message passing is performed in a TFN, we can extend the core mechanisms to construct the SE(3)-Transformer module that incorporates the attention mechanism. 

\newpage
\section{SE(3)-Transformer Module}
\label{sec:5}
\purple[]{
The Transformer model introduced in \citet{vaswani2017attention} uses the attention mechanism to learn dependencies between distant tokens in sequence data. The \textbf{SE(3)-Transformer} \citet{fuchs2020se} leverages the attention mechanism to weight certain messages between nodes with greater importance than others with scalar attention scores. Attention scores are calculated from node-based query embeddings generated via linear self-interaction and edge-based key embeddings generated via the message-passing mechanism of TFNs. Instead of aggregating messages uniformly, attention scores introduce an added degree of learnable angular dependence and nonlinearity that increases model performance.
}

\subsection{Primer on Self-Attention for Sequences}
Before diving into how to assemble an equivariant self-attention layer for geometric graphs, let’s refresh how \textbf{self-attention is applied to sequence data} (e.g., amino acid sequence, small molecule drug SMILES string).

\begin{figure}[h!]
\centering
\includegraphics[width=\linewidth]{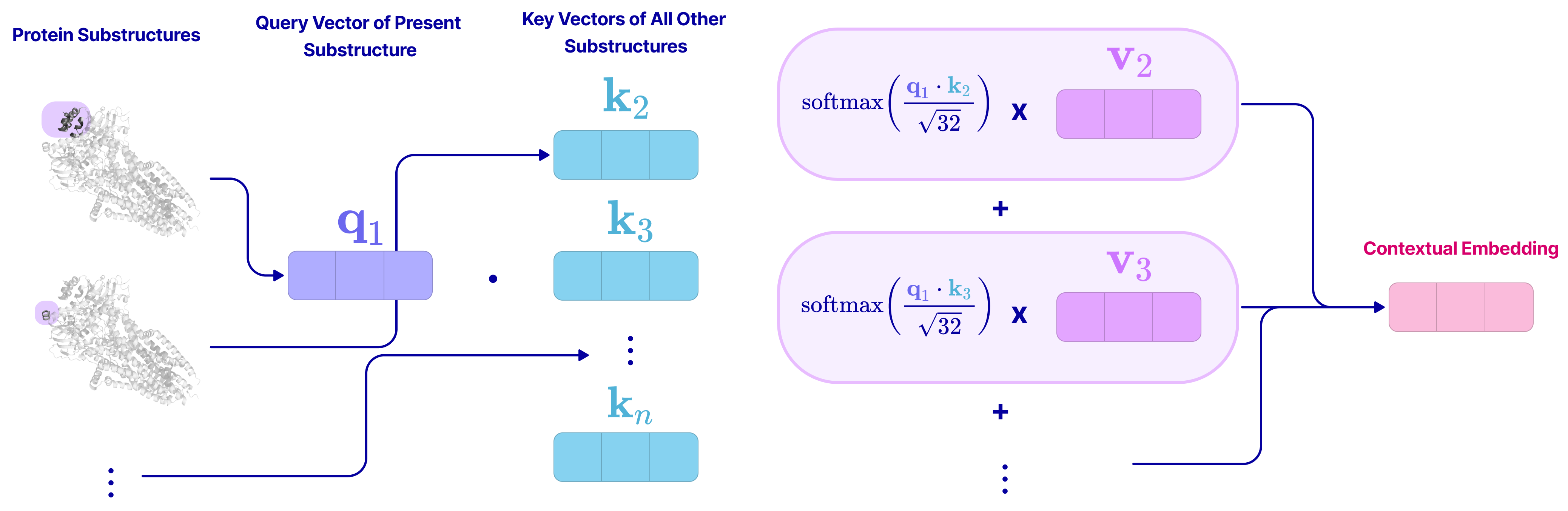}
\caption{Diagram of the attention mechanism on a sequence of protein substructures. Each substructure is represented by a feature vector, which is transformed to query, key, and value embeddings. The dot product between the query embedding of a single substructure (analogous to the center node) and the key embeddings of all other substructures in the sequence generates a set of raw attention scores. These attention scores are scaled down and fed into a softmax function before being used to scale the value embeddings. The weighted sum of every value embedding is a contextual embedding of the substructure corresponding to the query embedding.}
\label{fig:self-att}
\end{figure}

\begin{enumerate}
    \item First, the initial embedding of each element in the sequence—containing information on its independent properties and positional information—is converted into a \textbf{query, key, and value embedding} via matrix-vector multiplication with a learnable query, key, and value weight matrix. The dimension of the embedding is generally reduced to facilitate efficient multi-head attention.
    
    \item To calculate attention on a given element $i$ in the sequence, we take the \textbf{dot product of the query vector of $i$ with the key vectors of all the other elements in the sequence $j \neq i$} and normalize with the softmax function to generate a score between 0 and 1 corresponding to each $j$ such that the sum across all elements in the sequence is equal to 1. Higher scores represent stronger relationships (or a higher degree of dependency) between the element $i$ with the element $j$ in the representation subspace defined by the set of query, key, and value matrices.
    
    \item Then, the \textbf{value vectors} for all the elements in the sequence $j \neq i$ are scaled by their corresponding score and added to generate a learned internal representation or contextual embedding of the element $i$. This contextual embedding not only holds information on the element itself but also describes its relationship with every other element in the sequence.
    
    \item \textbf{Multi-head attention} generates \textbf{multiple attention scores} by multiplying the initial embedding with multiple sets of learned query, key, and value matrices, each of which projects the embeddings to a new representation subspace that captures a particular type of relationship between elements in the sequence.
    
    \item The contextual embeddings from each attention head are \textbf{concatenated and multiplied with an additional learned weight matrix} that captures the relative importance of each representation subspace in describing the element, aggregating all the representations into a single embedding.
    
    \item Finally, the contextual embeddings for all elements in the sequence are passed into a conventional \textbf{neural network} that generates a prediction for a classification or regression task.
\end{enumerate}

\subsection{Self-Attention for Geometric Graphs}

Due to the equivariance constraint, there are no learnable parameters with angular dependence in the TFN module. The kernel generated from the angular unit vector has \textbf{fixed angular dependence} and cannot incorporate dynamic learnable parameters without breaking equivariance, limiting the ability of the kernels to learn complex angular dependencies between nodes.

The SE(3)-Transformer overcomes this challenge by incorporating \textbf{SE(3)-invariant attention scores} computed from query and key embeddings dependent on both learnable parameters and fixed angular basis kernels. These scores are then used to scale the value embeddings for each edge, adding an extra degree of freedom for capturing angular relationships between nodes.

\begin{figure}[h!]
\centering
\includegraphics[width=0.8\linewidth]{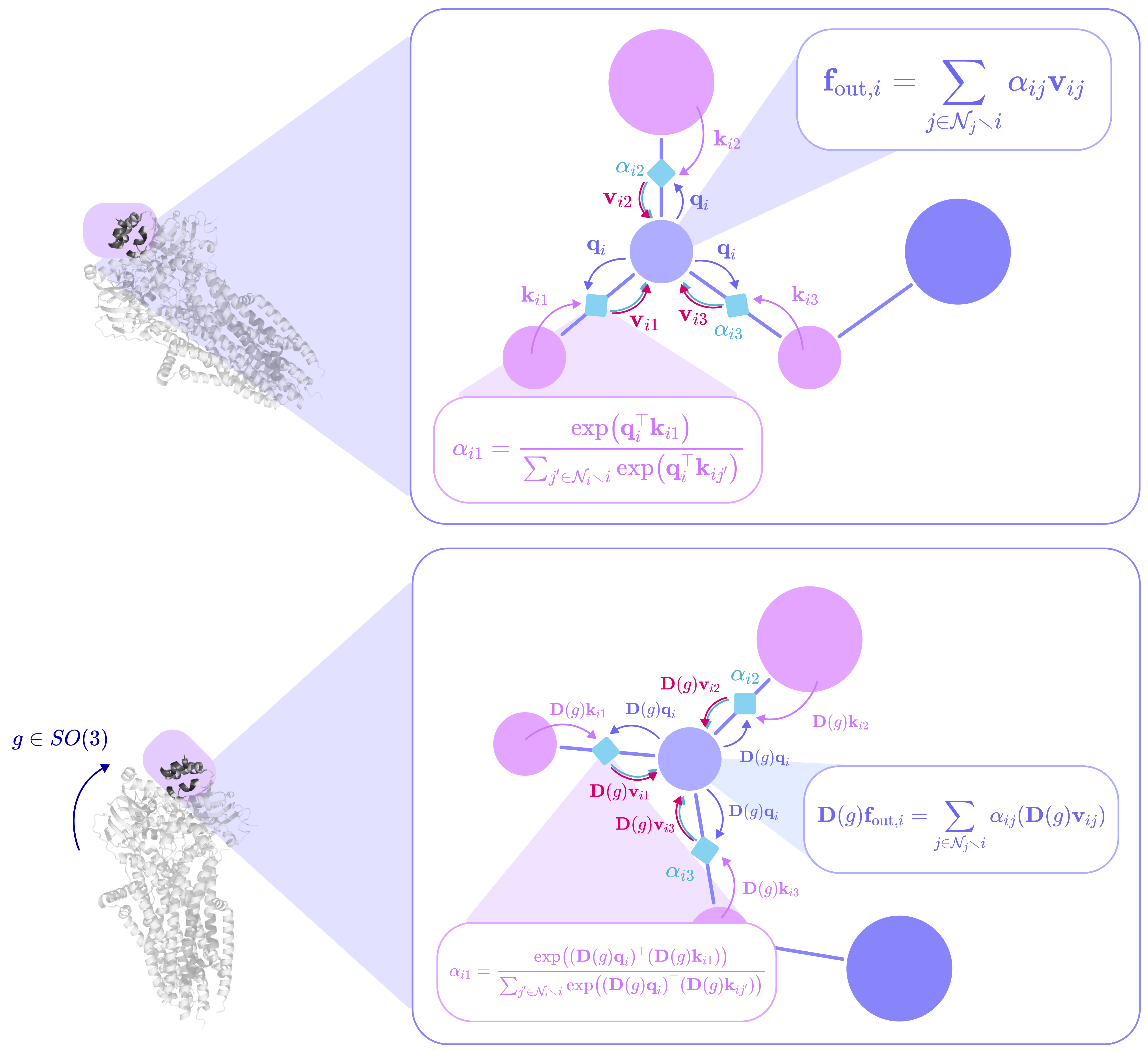}
\caption{The query, key, and value embeddings are generated such that they transform equivariantly under rotation using equivariant kernels. This means the dot product between the query embedding of the center node and the key embeddings of all neighboring nodes is invariant under rotation. The attention scores generated from the dot product are also invariant and are used to scale the equivariant value messages from each neighborhood node to get the output neighbor-to-center message for the center node.}
\label{fig:se(3)}
\end{figure}

\textbf{The equation below describes the full mechanism of the SE(3)-Transformer module:}

\begin{align}
    \mathbf{f}^l_{\text{out},i,c_l}=\underbrace{\sum_{c_l'}w^{l l}_{(i,c_l,c_l')}\mathbf{f}^l_{\text{in},i,c_l'}}_{\text{attentive self-interaction}}+\underbrace{\sum_{j\neq i}^n\underbrace{\alpha_{ij}}_{\text{attention score}}\underbrace{\sum_{k\geq 0}\underbrace{\sum_{c_k}\underbrace{\mathbf{W}^{l kV}_{(c_l,c_k)}(\mathbf{x}_{ij})\mathbf{f}^k_{\text{in},j,c_k}}_{\text{channel }c_k\to c_l\text{ value message}}}_{\text{message from type-}k\text{ input channels}}}_{\text{type-}l\text{ message from node }j}}_{\text{weighted sum over all }n \text{ adjacent nodes}}
\end{align}

In summary, self-attention can be applied to generate a message for a node $i$ with the following steps:

\begin{enumerate}
    \item The features at each node are transformed into a \textbf{node-based query embedding} by applying \textbf{linear self-interaction across same-degree channels} with a set of learnable weight matrices, generating each output query tensor channel as a linear combination of all input channels of the same type. The query tensors for every channel of every degree are concatenated into a single query embedding associated with the center node.

    \item An \textbf{edge-based key embedding} with the same dimension as the query embedding is generated for every directional edge in the graph in the same way as the neighbor-to-center messages of the TFN layer.
    
    \item The dot product between the query embedding of the center node and the key embedding for each incoming edge computes the \textbf{raw attention scores} associated with each incoming edge. 
    
    \item The raw attention scores across edges that share the same destination node are fed into the softmax function, which \textbf{normalizes the scores into a set of probabilities that sum to 1}.
    
    \item Instead of taking the average message over all incoming edges to the center node, like the TFN module, the SE(3)-Transformer \textbf{scales a value message generated for each edge by the associated attention score} and takes the sum to get the aggregated neighbor-to-center message.
    
    \item Then, the input features from the center node are concatenated to the neighbor-to-center and fed into an \textbf{attentive self-interaction layer}, which projects the concatenated tensor to match the output fiber. A matrix of weights is uniquely computed for each node with the attention mechanism and multiplied by the concatenated tensor to generate the final output message.
\end{enumerate}

\textbf{Now, we will break down the implementation of these steps in detail.}

\subsection{Computing the Query Embedding}
\label{subsec:query}
\purple[]{
The query embedding serves as a representation of the center node that is used to look up specific interactions and motifs in the features of neighboring nodes that are relevant in updating its position and feature tensors. In the context of modeling proteins, the query embedding could contain data relevant to determining an amino acid’s interaction with other amino acids, such as hydrophobicity, charge, and the torsion angles between $N-C\alpha$ and $C\alpha-C$ bonds.
}

\begin{figure}[h!]
\centering
\includegraphics[width=\linewidth]{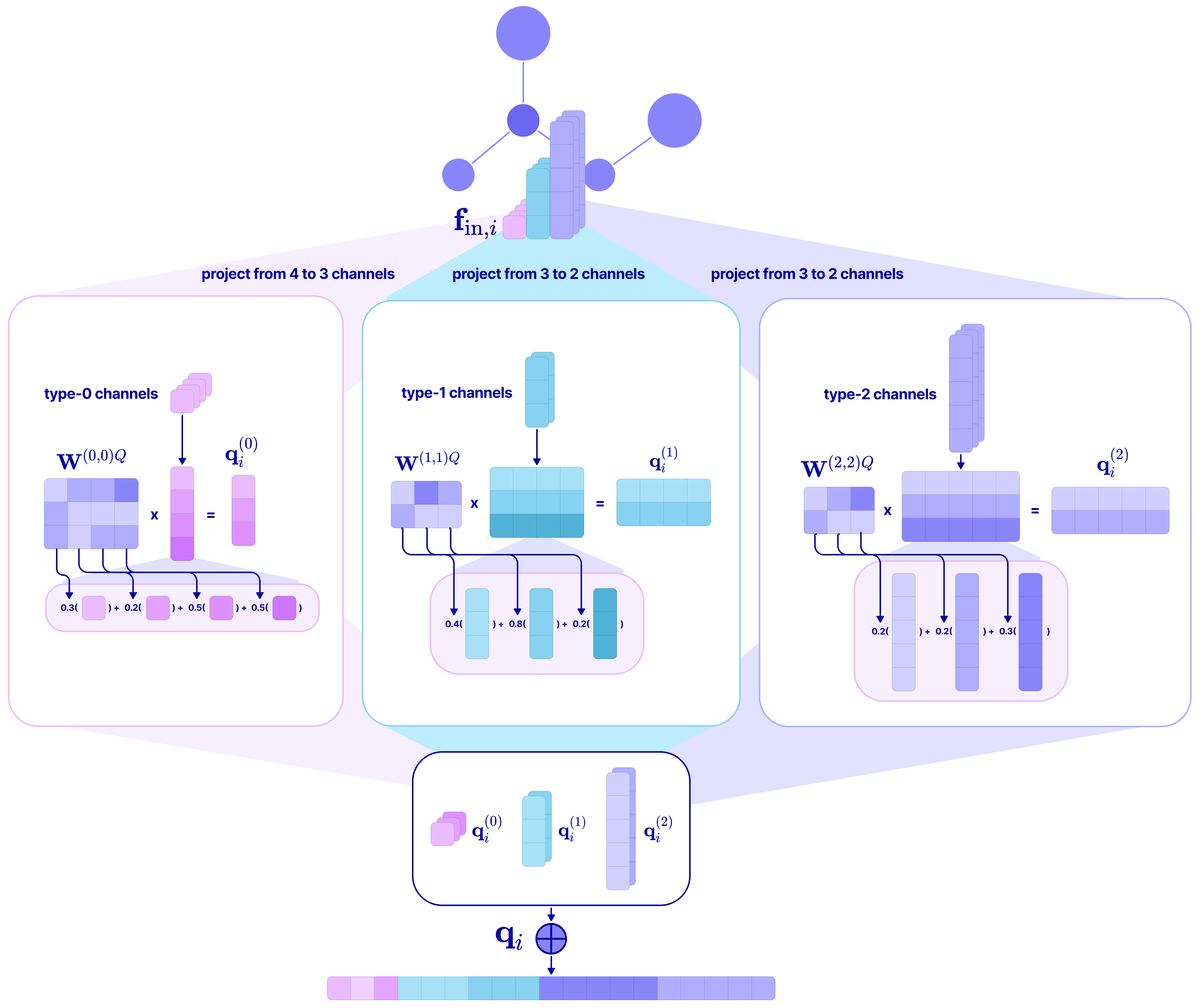}
\caption{The query embedding for the center node $i$ is generated using linear self-interaction across the channels of the same degree in the input feature tensor. The query embedding for every type-$l$ output channel is the weighted sum of all the type-$l$ input channels. The query embeddings for every type-$l$ output channel are computed in parallel by multiplying a learnable query weight matrix with the matrix formed by stacking the input channels into the rows. Finally, the query embeddings for every channel of every degree are concatenated into a single vector suitable for taking the dot product with the key embedding.}
\label{fig:query}
\end{figure}

The query embedding is a \textbf{node-based embedding}, which means it should only depend on the features of the node itself, because it is used to calculate the attention weights for all incoming edges.

Since we cannot transform between different feature types in a single node without the displacement vector, we use the \textbf{linear self-interaction mechanism} described in Section \ref{subsec:linear-self-interact} to \textbf{project each input feature type to its specified number of output channels}, where each output channel is the learnable weighted sum of all input channels of the same type.

To determine the dimensions of the query embedding, we can define a new fiber structure called \texttt{f\_mid\_in} that contains (multiplicity, degree) tuples corresponding only to degrees in \textbf{both the input and output fiber} but with the \textbf{same number of channels as the value messages} of the self-attention layer.

To illustrate, if the input fiber to the attention layer has structure $[(4, 0)]$ and the value fiber \texttt{f\_mid\_out} has structure $[(16, 0), (16, 1), (16, 2), (16, 3)]$, \texttt{f\_mid\_in} has structure $[(16, 0)]$, meaning the full query embedding will consist of 16 channels of type-0 tensors.

\begin{lstlisting}[language=Python]
# define fiber f_mid_in with same multiplicities as value msgs with structure f_mid_out, but for degrees in input f_in
f_mid_in = {d: m for d, m in f_mid_out.items() if d in self.f_in.degrees}
\end{lstlisting}

Now that we have defined its structure, we will break down how to calculate the query embedding for the center node $i$:

\begin{enumerate}
    \item The learnable weights are stored for each input degree as a \textbf{query weight matrix} with the following dimensions:
    \begin{align}
        \mathbf{W}^{l l Q}\in \mathbb{R}^{(\text{type-}l \text{  channels in value message})\times \text{(type-}l \text{ input channels})}
    \end{align}
    We define a dictionary of weight matrices where each degree $l$ is linked to a query weight matrix. The weights are initialized with random integers that are optimized with gradient descent and scaled down by the number of type-$l$ input channels, mi:
\begin{lstlisting}[language=Python]
self.transform = nn.ParameterDict()
# loop through all degrees in f_mid_in
for m_mid, d_mid in self.f_mid_in.structure:
    # extract number of input channels of degree d_mid to define dimensions of weight matrix
    mi = self.f_in.structure_dict[do]
    # initialize m_mid x mi weight matrix with random integers and scale down
    self.transform[str(do)] = nn.Parameter(torch.randn(m_mid, mi) / np.sqrt(mi), requires_grad=learnable)
\end{lstlisting}
    
    \item The query embedding for a single type-$l$ output channel $c_l$ is generated as a \textbf{weighted sum of all type-$l$ input channels} (indexed by $c_l’$) using the weights in the $c_l$th column of the query matrix:
    \begin{align}
        \mathbf{q}^l_{i,c_l}=\sum_{c_l'}w^{l l}_{c_l',c_l}\mathbf{f}^l_{\text{in},i,c_l'}
    \end{align}
    
    \item We can compute query embeddings for every type-$l$ output channel simultaneously by \textbf{multiplying the query matrix with all the type-$l$ features stacked horizontally} into a $(\text{input channels}) \times (2l+1)$ matrix, producing an $(\text{output channels}) \times (2l+1)$ matrix where \textbf{each row is the query embedding for a single output channel}. This operation is equivalent to separately computing the weighted sum for every type-$l$ output channel using the equation from step 2 and stacking the outputs into the rows of a matrix.
    \begin{align}
        \mathbf{q}^l_{i}=\text{vec}(\mathbf{W}^{l l Q}\mathbf{f}^l_{\text{in},i})
    \end{align}
    With the query matrices, we can implement the code that generates the matrix of type-$l$ output query embeddings for every node in the graph:
\begin{lstlisting}[language=Python]
output = {}
# loop through output degrees in features dictionary and extract features f
for do, f in features.items():
    # check if there is a query matrix corresponding to the degree do
    if str(do) in self.transform.keys():
        # calculate the query for every channel of degree do
        # output has shape (mo, 2*do+1)
        output[do] = torch.matmul(self.transform[str(do)], f)
return output
\end{lstlisting}
\end{enumerate}

Finally, we repeat this step for every degree defined in \texttt{f\_mid\_in}. The code above is implemented in the \texttt{G1x1SE3} module for linear self-interaction that is called in the \texttt{GSE3Res} module to generate a \textbf{list of query arrays} for every node in the graph, where the $l$th element of each list is an \textbf{array of type-$l$ queries with shape (type-$l$ channels in \texttt{f\_mid\_in}, $2l+1$)}:

\begin{lstlisting}[language=Python]
# initialize the function for generating the query that projects from f_in to f_mid_in
self.GMAB['q'] = G1x1SE3(f_in, self.f_mid_in)
# call function in forward pass with the features as input and store list in q
q = self.GMAB['q'](features, G=G)
\end{lstlisting}

Before we calculate the attention scores, we have to reshape the query embedding into a single vector that can be used to take the dot product with the key embeddings. We do this by \textbf{(1) concatenating the queries across all channels of each degree} by squeezing the last two dimensions corresponding to the channel and tensor-component axis of each array into a single dimension and \textbf{(2) concatenating the embeddings of all degrees} by squeezing along the last dimension of the array. This is implemented with the following helper function with the number of attention heads set to 1 for now:

\begin{lstlisting}[language=Python]
# F: list where each element is an array with shape (m, 2*d+1) for each feature degree d
# assume for now that H = 1 (single attention head)
def fiber2head(F, H, structure):
    # squeezes each array in the list into a m*(2*d+1)-dimensional vector of all channels concatenated along the last dimension
    fibers = [F[f'{d}'].view(*features[f'{d}'].shape[:-2], H, -1) for d in structure.degrees]
    # concatenate across the last dimension of every array in the list to get a single vector
    fibers = torch.cat(fibers, -1)
    return fibers
\end{lstlisting}

Now, we can call this function to concatenate all the query embeddings for all output type-$l$ channels $c_l$ in the value fiber and all input degrees $l$ to get the full query embedding for node $i$:

\begin{align}
    \mathbf{q}_i=\underbrace{\bigoplus_{l \geq 0}\underbrace{\bigoplus_{c_l}\sum_{c_l'}w^{l l}_{c_l',c_l}\mathbf{f}^l_{\text{in},i,c_l'}}_{\text{query emb for all type-}l\text{ channels}}}_{\text{query emb for all input degrees}}
\end{align}

\begin{lstlisting}[language=Python]
G.ndata['q'] = fiber2head(q, self.n_heads, self.f_mid_in)
\end{lstlisting}

\subsection{Computing the Key Embeddings}
\purple[]{
While the query embedding serves as a lookup tool that searches for relevant interactions and features from neighboring nodes, the key embedding serves as the dictionary of all interactions and feature motifs from neighborhood nodes that the query of the center node is compared to. If the query and key for a given edge are similar (large dot product), more weight is placed on the value message generated from that edge.
}

\begin{figure}[h!]
    \centering
    \includegraphics[width=\linewidth]{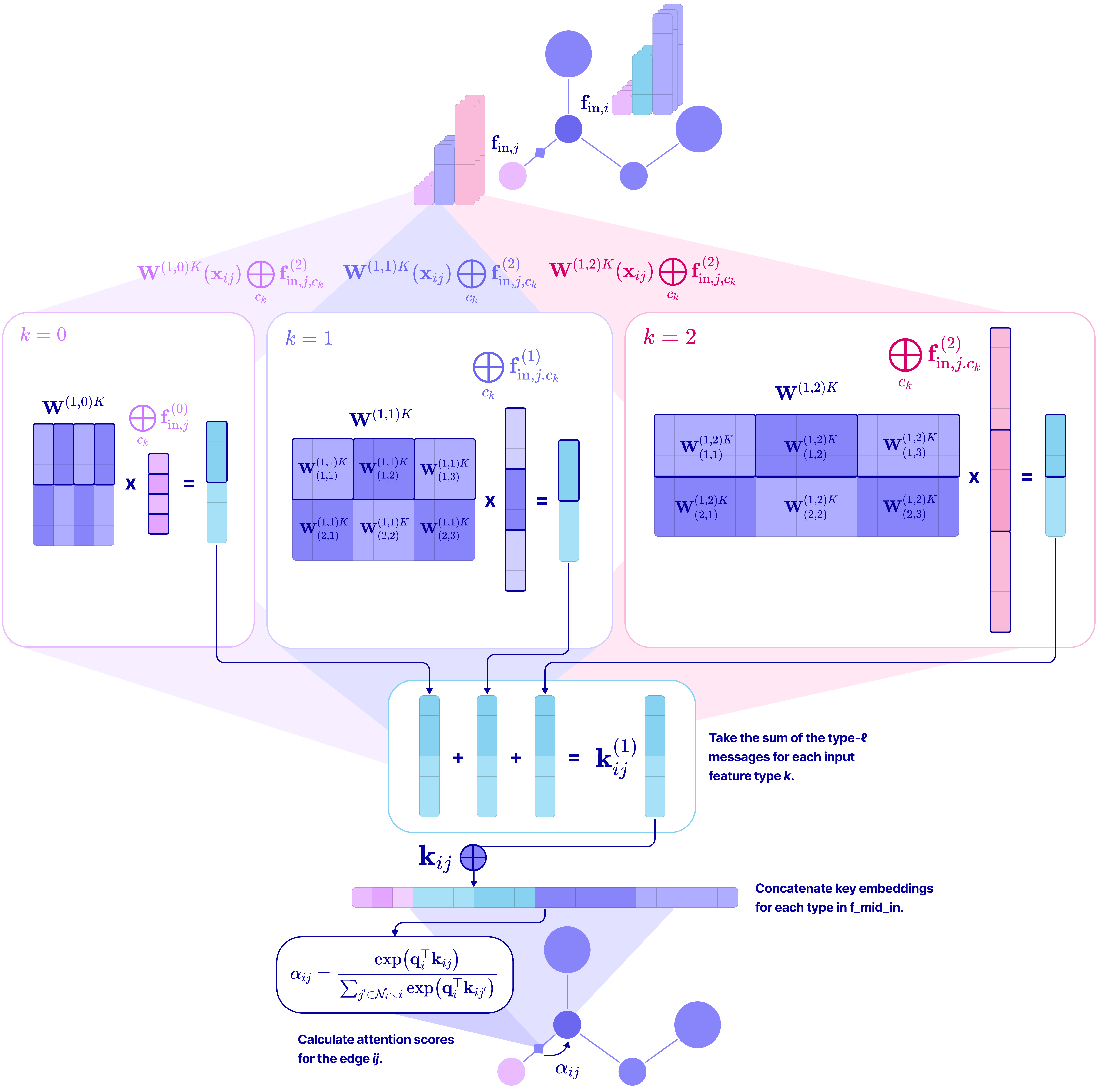}
    \caption{Diagram of how an input feature tensor from node $j$ is converted into a type-1 key embedding with 2 channels (this process is the same for each key embedding type defined in \texttt{f\_mid\_in}). First, the channels for a single input type $k$ are concatenated into a single feature vector and multiplied by the corresponding key weight matrix (where each block is a single transformation between two channels) that outputs the direct sum of the 2 type-$l$ key embedding channels. The 2 type-$l$ key embedding channels generated from the type-0, type-1, and type-2 input channels are summed and used to calculate the attention score for the edge $ij$.}
    \label{fig:key}
\end{figure}

The key embedding is an \textbf{edge-based embedding} generated in the same way as the neighbor-to-center message in the TFN module without self-interaction. For each edge, the key embedding is generated by transforming every feature in the source node with uniquely defined equivariant kernels constructed with learnable radial weights and spherical harmonics with angular dependence, such that the embedding contains information on both the source node and the edge features between the source and center node.

Note that the key embeddings for the edge from node $j$ to $i$ and the edge from node $i$ to $j$ are different, so each node has a unique set of key embeddings corresponding to every incoming edge from nodes in the neighborhood centered around it.

Since the \textbf{dimension key embedding must align with the query embedding to compute the element-wise dot product}, we use the \texttt{f\_mid\_in} fiber as the output fiber of the neighbor-to-center message computation to ensure that the multiplicity for each type of key embedding aligns with the query. From now on, we will refer to the \textbf{multiplicity for an arbitrary but particular degree $l$ in \texttt{f\_mid\_in} as mid}. 

The type-$l$ key embedding for the edge from node $j$ to node $i$ is generated as follows:

\begin{enumerate}
    \item For computing all key embeddings across every edge, a \textbf{set of key radial functions} (denoted by the superscript $K$) are defined for every ordered pair of input and output degrees $(k, l)$, each of which outputs a (mid)(type-$k$ input channels, mi)($2\min(l,k)+1$)-dimensional vector of weights for every input radial distance.
    \begin{align}
        \varphi^{l kK}(\|\mathbf{x}_{ij}\|):\mathbb{R}^3\to \mathbb{R}^{{\text{(mi)(mid)}(2\min(l,k)+1)}}
    \end{align}

    \item From the key radial function, we construct a $(\text{mid})(2l+1) \times (\text{mi})(2k+1)$ equivariant key kernel that \textbf{transforms all the type-$k$ input channels from node $j$ to the number of type-$l$ output key embeddings defined in \texttt{f\_mid\_in}}.
    \begin{align}
        \mathbf{W}^{l kK}(\mathbf{x}_{ij})=\begin{bmatrix}\mathbf{W}^{l kK}_{(1,1)}&\mathbf{W}^{l kK}_{(1,2)}&\dots&\mathbf{W}^{l kK}_{(1,\text{mi})}\\\\\mathbf{W}^{l kK}_{(2,1)}&\ddots&\dots &\vdots \\\\\vdots &\dots &\ddots& \vdots\\\\\mathbf{W}^{l kK}_{(\text{mid},1)}&\dots &\dots &\mathbf{W}^{l k}_{(\text{mid},\text{mi})}\end{bmatrix}_{(\text{mid}*(2l+1))\times(\text{mi}*(2k+1))}
    \end{align}
    Each block of the $l k$ key matrix is the $(2l+1) \times (2k+1)$ matrix that transforms a single type-$k$ input channel $c_k$ to a single type-$l$ key embedding for channel $c_l$:
    \begin{align}
        \mathbf{W}^{l kK}_{(c_l,c_k)}(\mathbf{x}_{ij})=\sum_{J=|k-l|}^{k+l}\varphi^{l kK}_{(J,c_l,c_k)}\mathbf{W}^{l k}_J(\mathbf{x}_{ij})
    \end{align}
    The key weight matrices are generated with the same code implementation as the kernels in the TFN layer described previously.
    
    \item Next, we can compute a corresponding key embedding for every type-$l$ channel in the query embedding by taking the \textbf{matrix-vector product} between the key kernel defined above and the direct sum of all type-$k$ input channels from node $j$.
    \begin{align}
        \mathbf{k}^l_{ij}=\bigoplus_{c_l}\mathbf{k}^l_{ij,c_l}=\underbrace{\sum_{k\geq0}\underbrace{\mathbf{W}^{l kK}(\mathbf{x}_{ij})\bigoplus_{c_k}\mathbf{f}^k_{\text{in},j,c_k}}_{\text{key emb from type-}k\text{ input channels}}}_{\text{sum over type-}l\text{ key embs for all input types }}
    \end{align}
    \begin{figure}[h!]
    \centering
    \includegraphics[width=0.6\linewidth]{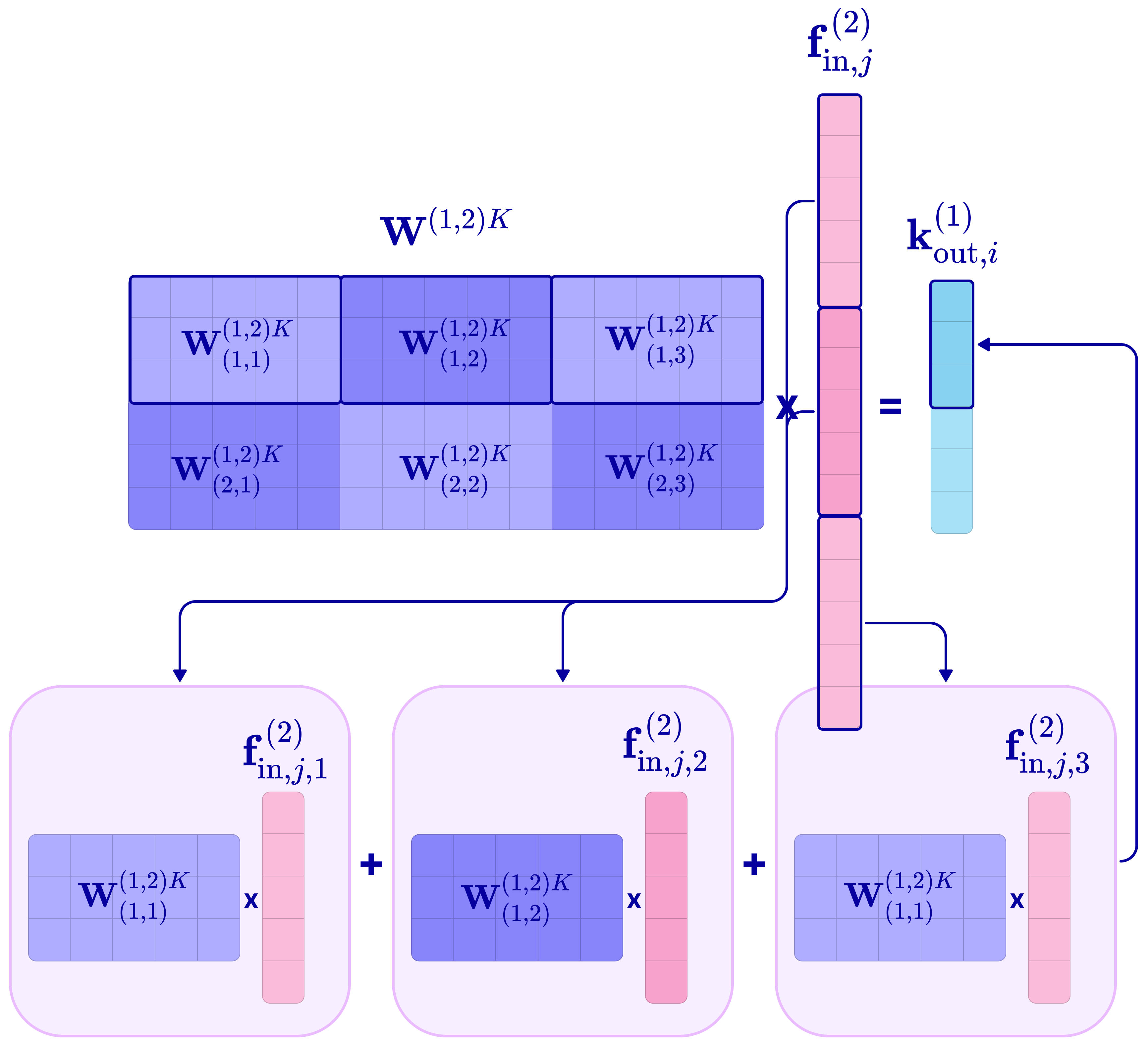}
    \caption{Taking the matrix-vector product between the combined key weight matrix transforming from types $k$ to $l$ and vector concatenation of every type-$k$ input channel is equivalent to separately multiplying a row of $(2l+1) \times (2k+1)$ blocks in the key kernel with the corresponding type-k input channel, and taking the sum of the type-$l$ outputs to get a single channel of the type-$l$ key embedding. Each row of kernels in the combined key kernel transforms the type-$k$ input channels into a single type-$l$ output channel.}
    \label{fig:key2}
    \end{figure}
    The code to implement this calculation is the same as step 6 in the section on equivariant message-passing, except with the incorporation of edge features, which we will discuss in Section \ref{subsec:edge-features}.
\end{enumerate}

The keys for every degree in \texttt{f\_mid\_in} are generated with the same process as step 3, only with a new set of key kernels.

In the code implementation, we loop through all the degrees in \texttt{f\_mid\_in} and call the \texttt{udf\_u\_mul\_e} UDF defined the same way as the TFN module which computes all the type-$l$ key embeddings for every edge in the graph and stores it in a dictionary with a single label \texttt{f`out\{l\}`} linked to an array with shape (edges, mid, $2l+1$). Then, the dictionary is stored in the edge data by calling the built-in \texttt{apply\_edges} function in DGL.

\begin{lstlisting}[language=Python]
for d_mid in self.f_mid_in.degrees:
    # udf_u_mul_e is a user-defined function that stores an array with shape (edges, type d_mid channels, 2*d_mid+1) in the edge data with the label f'out{d_mid}'
    G.apply_edges(self.udf_u_mul_e(d_mid))
\end{lstlisting}

Then, we return a dictionary where every degree is linked to a list of all the key embedding channels of that type for every edge.

\begin{lstlisting}[language=Python]
# return all key embeddings for all edges in a single dictionary 
# each degree is linked to an array with shape (edges, type-d_mid channels, 2*d_mid+1)
return {f'{d_mid}': G.edata[f'out{d_mid}'] for d_mid in self.f_mid_in.degrees}
\end{lstlisting}

The code above is implemented in the module called \texttt{GConvSE3Partial}, which is called in the \texttt{GSE3Res} module:

\begin{lstlisting}[language=Python]
# edge_dim: used to determine the dim of input (edge_dim+1) to the radial function
# if only radial distance, the edge_dim is 0
# x_ij: type-1 displacement vector used as edge feature
self.GMAB['k'] = GConvSE3Partial(f_in, self.f_mid_in, edge_dim=edge_dim, x_ij=x_ij)
\end{lstlisting}

Just like the query embedding, we concatenate the key embeddings for every channel of every type into a single vector with the same dimension as the query embedding for each edge.

\begin{align}
    \mathbf{k}_{ij}=\underbrace{\bigoplus_{l\geq0}\underbrace{\sum_{k\geq 0}\mathbf{W}^{l kK}(\mathbf{x}_{ij})\bigoplus_{c_k}\mathbf{f}^k_{\text{in},j,c_k}}_{\text{key emb for all type-}l\text{ channels}}}_{\text{key emb for all degrees in fiber}}
\end{align}

This is implemented with the \texttt{fiber2head} function defined earlier:

\begin{lstlisting}[language=Python]
# assume n_heads=1 for now
# squeezes the last two dimensions of the key embedding into an array with shape (edges, 1, dimension of query)
G.edata['k'] = fiber2head(k, self.n_heads, self.f_mid_in)
\end{lstlisting}

Now, we have a \textbf{single key embedding for every edge pointing to node $i$} that forms the complete set of embeddings used to generate the set of attention scores for node $i$, denoted $\{\mathbf{k}_{ij}\}_{j\in\mathcal{N}_i\setminus i}$.

\subsection{Calculating Attention Scores}

For each node in the graph, we compute a set of SE(3)-invariant attention scores by taking the dot product between the query embedding of the node with the set of key embeddings associated with each incoming edge.

Since the dot product measures the distance between tensors, the \textbf{raw attention scores indicate the similarity between the query representation of node $i$ and the key representations of each neighborhood node}. The key representations are encoded with learnable parameters, so they are trained to represent features from neighborhood nodes with strong interactions with the center node, such that they produce large attention scores with the query embedding, while representing features irrelevant to the center node, such that they produce low attention scores.

To illustrate, consider a model tasked to predict the interaction between amino acids in a protein. The parameters in the key matrix for a given edge can learn to transform the node features of the ‘\textit{source}’ residue carrying a positive charge into a key embedding that \textbf{produces a large dot product} with the query embedding of the ‘\textit{center}’ residue carrying a negative charge.

\begin{figure}[h!]
\centering
\includegraphics[width=\linewidth]{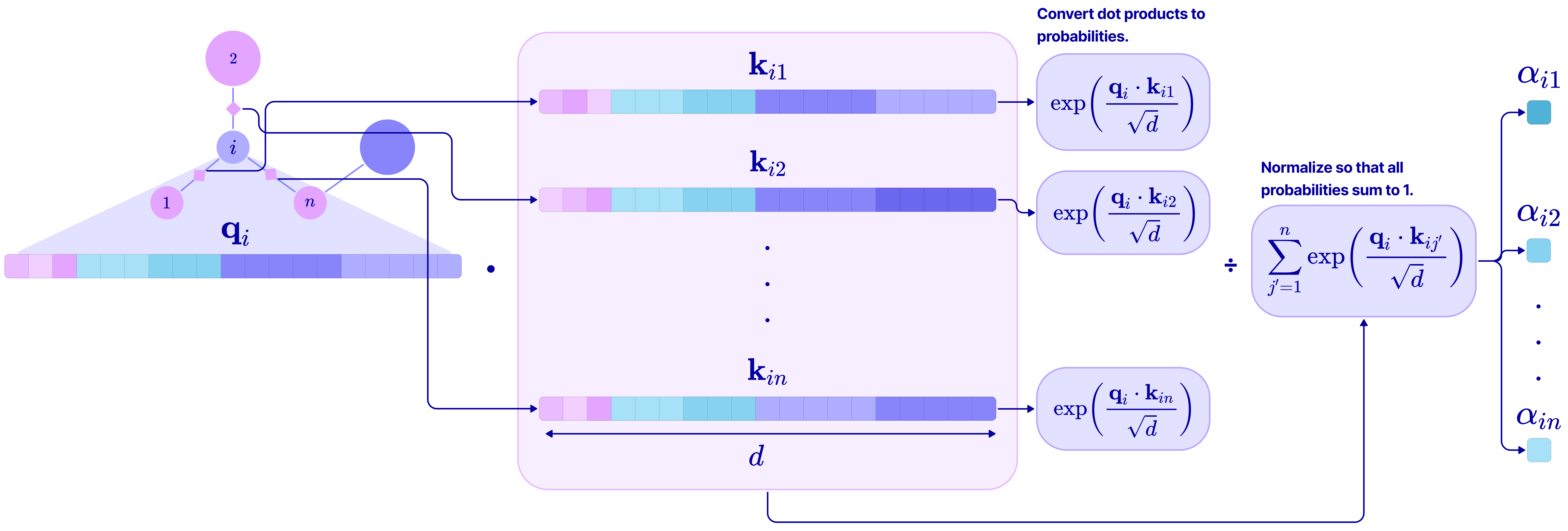}
\caption{The set of attention scores corresponding to a single node $i$ in the graph is computed by \textbf{(1)} taking the dot product between the query embedding at the center node and all the key embeddings of adjacent nodes, \textbf{(2)} scaling the scores down by the square root of the dimension of the key embeddings and \textbf{(3)} applying the softmax function to produce a set of attention scores that sum to 1.}
\label{fig:att}
\end{figure}

In single-head attention, a single attention score (denoted by $\alpha$) is computed for each edge $ij$ and used to \textbf{scale all channels and feature types} in the value message from the source node $j$. The set of all scores for the center node $i$ can be calculated as follows:

\begin{enumerate}
    \item First, we take the dot product between the query embedding of node $i$ and all the key embeddings of incoming edges to generate the set of raw attention scores:
    \begin{align}
        \mathbf{q}_i\cdot\mathbf{k}_{ij}=\mathbf{q}_i^{\top}\mathbf{k}_{ij}
    \end{align}
    The following code uses the built-in \texttt{e\_dot\_v} DGL function to calculate the set of \textbf{raw attention scores} for every node in the graph from the set of keys of incoming edges. In other words, \texttt{e\_dot\_v} computes a scalar value for every edge in the graph that is equal to the \textbf{dot product between the key of the source node and the query of the destination node of the edge}. Since the key array has shape (edges, heads=1, query dimension) and the query has dimension (nodes, heads=1, query dimension), the function first broadcasts the query array such that every key embedding of an edge pointing to the same node is multiplied with the same query embedding. The output of the function is an array with shape (edges, heads=1) that is \textbf{stored in the edge data} labeled \texttt{‘e’} by calling the \texttt{apply\_edges} function:

\begin{lstlisting}[language=Python]
# e_dot_v takes (first tensor, second tensor, output field in edge)
G.apply_edges(fn.e_dot_v('k', 'q', 'e'))
\end{lstlisting}

    \item Then, we divide each score by the \textbf{square root of the dimension of the query and key vector $d$}, which equals the sum of (mid)($2l+1$) for all values of $l$ in \texttt{f\_mid\_in}. This prevents the dot product from growing too large and pushing the gradient to zero after applying the softmax function.
    \begin{align}
        \frac{\mathbf{q}_i\cdot\mathbf{k}_{ij}}{\sqrt{d}}
    \end{align}
    The dimension $d$ of the key and query fiber is stored in \texttt{n\_features}, an instance variable of the fiber object that is calculated by multiplying the number of channels and the dimension of the tensor-component axis for each degree and summing over all degrees in the fiber:
\begin{lstlisting}[language=Python]
# self.structure is a list of (multiplicity, degree) tuples
self.n_features = np.sum([m * (2*d+1) for (m, d) in self.structure])
\end{lstlisting}

    \item The \textbf{softmax function} is applied separately to \textbf{each set of raw attention scores of edges that share a center node}. This converts the raw scores into a probability between 0 and 1 such that the \textbf{sum across the set of scores for any given node is 1}.
    \begin{align}
        \alpha_{ij}=\frac{\exp\left(\frac{\mathbf{q}_i\cdot \mathbf{k}_{ij}}{\sqrt{d}}\right)}{\sum_{j'\neq i}^n\exp\left(\frac{\mathbf{q}_i\cdot\mathbf{k}_{ij'}}{\sqrt{d}}\right)}
    \end{align}
    This is implemented with the \texttt{edge\_softmax} DGL function, which is designed to compute softmax across edges with the same destination node for all edges in the graph.
\begin{lstlisting}[language=Python]
from dgl.nn.pytorch.softmax import edge_softmax

# extract raw attention scores for all edges from edge data
e = G.edata.pop('e')
# scale down all scores by the dimension of the key embedding
e = e / np.sqrt(self.f_mid_out.n_features) 
# apply softmax for each set of edges and store attention scores in an array with shape (edges, heads=1) labeled 'a' in the edge data 
G.edata['a'] = edge_softmax(G, e)
\end{lstlisting}
\end{enumerate}

\textit{Before moving on, let’s make sure the dot product of the query and key embeddings is SO(3)-equivariant.}

We know that each type-$l$ component of the query and key embeddings is SO(3)-equivariant since we constructed the key and query kernels such that they transform under the Kronecker product of type-$l$ and type-$k$ \textit{irreps}.

Then, we concatenate them into a single embedding in the same order such that the type-$l$ component of the query embedding aligns with the type-$l$ component of the key embedding. This means they both rotate under the same representation formed by the direct sum of \textit{irreps} (block-diagonal of Wigner-D matrices).

\begin{align}
    \mathbf{q}\mapsto \mathbf{D}(g)\mathbf{q}\text{ and }\mathbf{k}\mapsto \mathbf{D}(g)\mathbf{k}
\end{align}

Since representations of SO(3) are orthonormal, meaning they preserve lengths and angles, the dot product (and attention scores) are \textbf{invariant under rotation}:

\begin{align}
(\mathbf{D}(g)\mathbf{q})^{\top}(\mathbf{D}(g)\mathbf{k})&=\mathbf{q}^{\top}\mathbf{D}(g)^{\top}\mathbf{D}(g)\mathbf{k}\\
&=\mathbf{q}^{\top}\mathbf{k}
\end{align}

\subsection{Computing the Value Messages}
\purple[]{
The \textbf{value embeddings} for each edge contain the neighbor-to-center messages that are scaled and added together into a complete message used to update the features at the center node. The contribution made by each value message is determined by its corresponding attention score, which indicates whether the source node from which the message is generated has relevant interactions with the center node.
}

In this section, we will break down how to generate the \textbf{edge-based message}, called the \textbf{value embedding} in self-attention, that is scaled by an attention weight and summed together with the weighted value embeddings from all other incoming edges to generate the updated output feature tensor for the center node $i$.

\begin{figure}[h!]
\centering
\includegraphics[width=\linewidth]{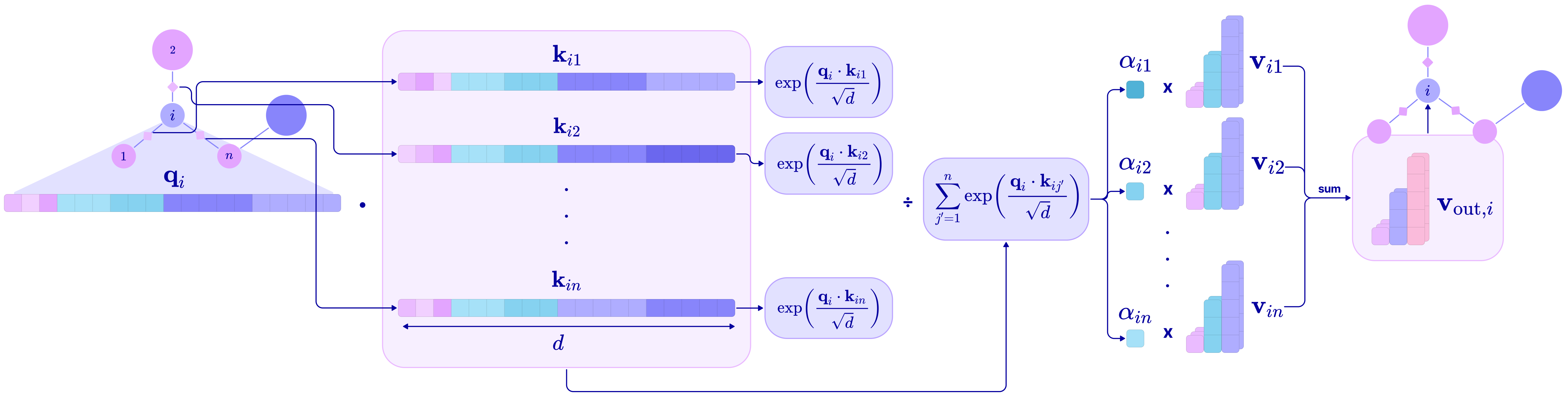}
\caption{After generating the set of attention scores corresponding to every incoming edge, they are used to scale the value message generated from that edge. The sum of all the scaled value messages is the aggregated neighbor-to-center message to node $i$.}
\label{fig:value}
\end{figure}

In the SE(3)-Transformer, the number of channels in the value message is scaled down from the structure of the output fiber. This means that the input feature fibers are projected to an \textbf{intermediate fiber structure}, called \texttt{f\_mid\_out}, where the channels for every feature type in the output fiber \texttt{f\_out} are scaled down by \texttt{div} before being projected into the total number of channels defined in \texttt{f\_out} using an attentive self-interaction layer. The intermediate fiber structure is initialized with the following code:

\begin{lstlisting}[language=Python]
# extract the multiplicity of each degree in f_out and divide by div
# // is the floor division operator
f_mid_out = {do: int(mo // div) for mo, do in self.f_out.structure_dict.items()}
self.f_mid_out = Fiber(dictionary=f_mid_out)
\end{lstlisting}

The steps involved in generating the value embedding are the same as the key embedding except instead of projecting the input features to match the structure of \texttt{f\_mid\_in}, the \textbf{value embeddings are generated from projecting the input features from the adjacent nodes to match the channels and degrees in \texttt{f\_mid\_out}} by multiplying with a unique set of value kernels generated with a unique set of radial functions. The value kernel that transforms all type-$k$ input channels into the number of type-$l$ value embeddings defined in \texttt{f\_mid\_out} has the following form:

\begin{align}
    \mathbf{W}^{l kV}(\mathbf{x}_{ij})=\begin{bmatrix}\mathbf{W}^{l kV}_{(1,1)}&\mathbf{W}^{l kV}_{(1,2)}&\dots&\mathbf{W}^{l kV}_{(1,\text{mi})}\\\\\mathbf{W}^{l kV}_{(2,1)}&\ddots&\dots &\vdots \\\\\vdots &\dots &\ddots& \vdots\\\\\mathbf{W}^{l kV}_{(\text{mval},1)}&\dots &\dots &\mathbf{W}^{l k}_{(\text{mval},\text{mi})}\end{bmatrix}_{(\text{mval}*(2l+1))\times(\text{mi}*(2k+1))}
\end{align}

where \texttt{mval} stands for the multiplicity of degree $l$ in \texttt{f\_mid\_out} and each block of the $l k$ key matrix is the $(2l+1) \times (2k+1)$ matrix that transforms a single type-$k$ input channel $c_k$ to a type-$l$ value embedding for channel $c_l$:

\begin{align}
    \mathbf{W}^{l kV}_{(c_l,c_k)}(\mathbf{x}_{ij})=\sum_{J=|k-l|}^{k+l}\varphi^{l kV}_{(J,c_l,c_k)}\mathbf{W}^{l k}_J(\mathbf{x}_{ij})
\end{align}

If the input and output fiber have structure $[(0, 32), (1, 32), (2, 32), (3, 32)]$ and \texttt{div} is set to 2, then \texttt{f\_mid\_out} has structure $[(0, 16), (1, 16), (2, 16), (3, 16)]$. When generating the value embeddings, the 32 input channels for a single type $k$ are projected down to 16 type-$l$ value embeddings with a $(16*(2l+1)) \times  (32*(2k+1))$ value kernel.

The \textbf{direct sum of the value embeddings for all type-$l$ channels} defined in \texttt{f\_mid\_out} is calculated by multiplying the combined value kernel with the direct sum of all type-$k$ input channels and summing over the products for all input types, just like the type-$l$ key embeddings:

\begin{align}
\mathbf{v}^l_{ij}=\bigoplus_{c_l}\mathbf{v}^l_{ij,c_l}=\underbrace{\sum_{k\geq0}\underbrace{\mathbf{W}^{l kV}(\mathbf{x}_{ij})\bigoplus_{c_k}\mathbf{f}^k_{\text{in},j,c_k}}_{\text{value msg from type-}k\text{ input channels}}}_{\text{sum over type-}l\text{ value msgs for all input types }}
\end{align}

For every output degree, we unvectorize the value embedding into an array with shape (edges, mval, $2l+1$). By calling the same \texttt{udf\_u\_mul\_e} UDF and \texttt{apply\_edges} function used to compute the keys, we compute the value arrays and store them in the edge data with the label \texttt{f`out\{dval\}`}.

\begin{lstlisting}[language=Python]
for dval in self.f_mid_out.degrees:
    # udf_u_mul_e is a user-defined function that stores an array with shape (edges, type dval channels, 2*dval+1) in the edge data with the label f'out{dval}'
    G.apply_edges(self.udf_u_mul_e(dval))
\end{lstlisting}

Then, we return all the value arrays in a single dictionary.

\begin{lstlisting}[language=Python]
# return all value embeddings for all edges in a single dictionary 
# each degree is linked to an array with shape (edges, type-dval channels, 2*dval+1)
return {f'{dval}': G.edata[f'out{dval}'] for dval in self.f_mid_out.degrees}
\end{lstlisting}

To calculate the output type-$l$ feature tensor for node $i$, we take the \textbf{weighted sum of the type-$l$ value messages from all $n$ edges} pointing to node $i$ and repeat for every degree in \texttt{f\_out}.

\begin{align}
    \underbrace{\sum_{j\neq i}^n\alpha_{ij}\mathbf{v}^l_{ij}}_{\text{neighbor }\to\text { center msg}}
\end{align}

In Section \ref{subsec:att-self-interact} on attentive self-interaction, we will project this tensor to match the output fiber structure and incorporate a skip connection. First, we will \textbf{extend the self-attention mechanism to multi-head attention}. Since the code implementation is written to generalize to multiple heads, we will also be breaking down the general implementation for single- and multi-head self-attention in the next section.

\subsection{Multi-Head Self-Attention}
\purple[]{
In sequence data, multi-head attention computes multiple sets of query, key, and value embeddings in parallel to capture distinct representation subspaces that each capture a particular type of relationship between elements in the sequence.
}

To illustrate the purpose of multi-head attention, \textbf{consider a node in a graph representing the amino acid cysteine}. Cysteine is non-polar, hydrophobic, and has the special property that its thiol group ($-SH$) reacts with the thiol groups of other cysteine residues to form a \textbf{disulfide bond} ($R-S-S-R$). Therefore, one feature channel for cysteine could indicate its hydrophobic property, and another could indicate the presence of the thiol group. A single attention head can detect the attractive forces between cysteine and other hydrophobic residues by generating a query embedding of the cysteine residue that \textbf{looks for key embeddings with motifs that indicate hydrophobicity}. This means that neighboring hydrophobic residues will have large attention scores, and the value messages corresponding to those residues will significantly contribute to the overall message. This set of attention scores can be considered a \textbf{single representation subspace} that uses the strength of hydrophobic interactions between residues to compute the overall message.

To capture a larger, more complex set of chemical relationships, like the formation of disulfide bonds, we can generate \textbf{multiple representation subspaces} by computing \textbf{multiple sets of attention scores}. Each set of attention scores depends on the strength of different chemical relationships, which are used to scale value messages to incorporate those interactions in the overall message.

Instead of using multiple sets of query, key, and value kernels like the traditional sequence Transformer, the SE(3)-Transformer performs multi-head attention by \textbf{splitting the channels of the query, key, and value embeddings} generated with the same set of query, key, and value kernels as in single-head attention into multiple heads.

For an edge, each channel of the key embedding is generated with a set of key matrices defined by a unique set of learnable radial weights trained specifically for that channel, so the key embeddings across multiple channels already capture multiple representation subspaces. This is also the case across channels of the query and value embeddings. All we need to do is compute multiple attention weights for each edge that provide \textbf{multiple paths} for the model to learn more complex relationships.

\begin{figure}[h!]
    \centering
    \includegraphics[width=\linewidth]{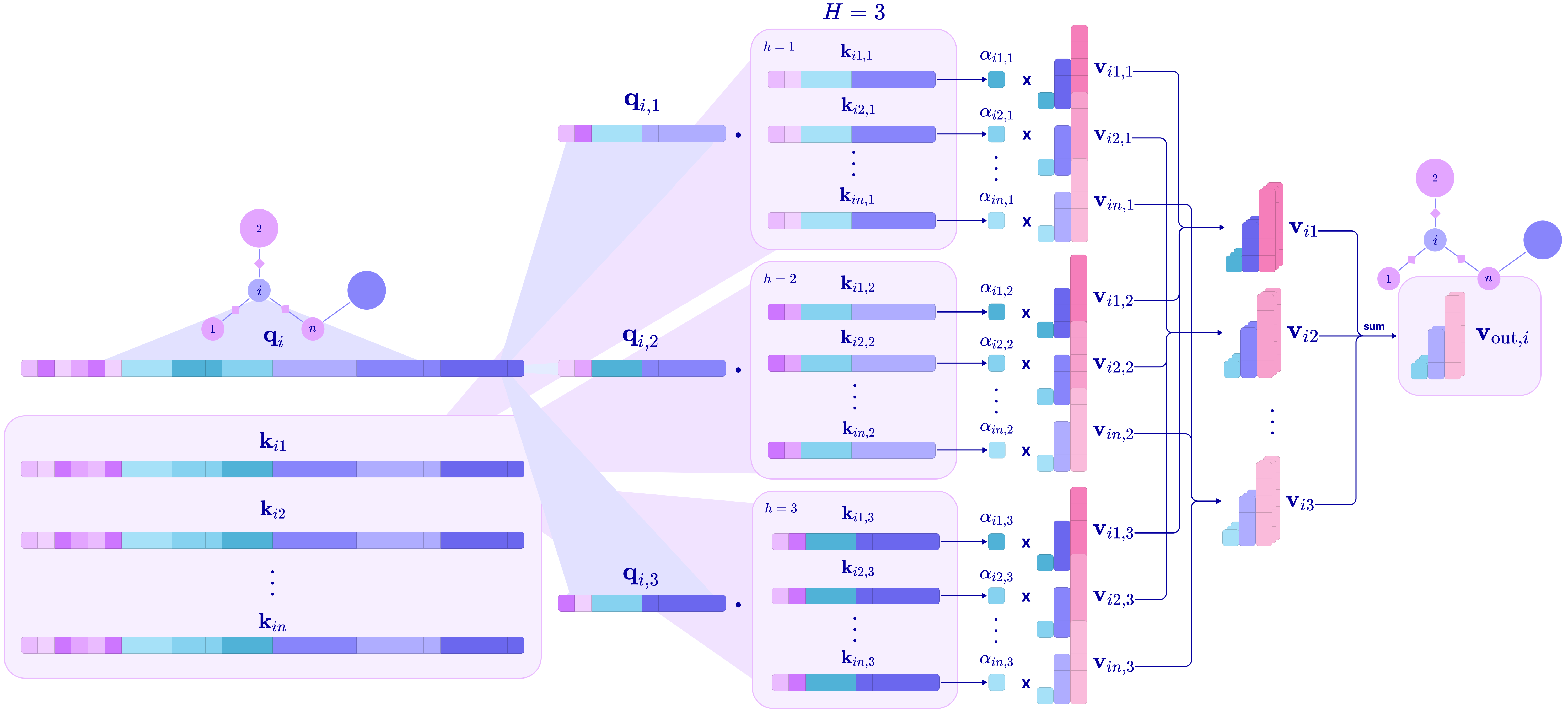}
    \caption{Diagram showing the mechanism of multi-head attention with 3 heads. The full query and key embeddings have 6 type-0 channels, 3 type-1 channels, and 3 type-2 channels, which are split into 3 query embeddings and 3 sets of key embeddings with 2 type-0 channels, 1 type-1 channel, and 1 type-2 channel per embedding. At each head, the dot product between a query embedding and the set of key embeddings of incoming edges is calculated to generate 3 distinct sets of attention scores for 3 attention heads. The value embeddings are also split into 3 sets, where each set contains a subset of channels from each edge. Finally, each set of attention scores is used to scale one set of value messages, and the value embeddings corresponding to the same edge are concatenated to form a full value embedding for each of the $n$ edges. Finally, we take the sum of all $n$ value messages to get the updated feature tensor.}
    \label{fig:multi}
\end{figure}

\textbf{SE(3)-Transformers implement multi-head self-attention by simply splitting the channels of features of the same type into several subsets} (indexed by $h$ with a total of $H$ subsets) \textbf{for which attention scores are calculated separately}. For instance, if there are 16 channels for every type of feature in \texttt{f\_mid\_out}, we can perform multi-head attention with 8 attention heads ($H = 8$) where each head computes a unique attention weight from the key and query embeddings of 2 channels of each degree, which is used to scale 2 value embedding channels for each degree.

Instead of generating a single attention weight for each edge, there are \textbf{$H$ attention weights for each edge that are used to scale the corresponding subset of value embeddings}.

\begin{enumerate}
    \item First, we divide the query and key embedding channels of each type into the number of heads, such that there are $H$ query vectors and $H$ key vectors that both have dimensions equal to the full key dimension $d$ divided by $H$. We reshape and store the query and key embeddings for each head in two arrays with shape $(H, d/H)$ by calling the \texttt{fiber2head} function defined earlier:
\begin{lstlisting}[language=Python]
# reshapes key and query into shape (edges, heads, d/H)
G.edata['k'] = fiber2head(k, self.n_heads, self.f_key, squeeze=True) 
G.ndata['q'] = fiber2head(q, self.n_heads, self.f_key, squeeze=True)
\end{lstlisting}
    
    \item At each node $i$ and head $h$, we take the \textbf{dot product} between the component of the split query embedding of node $i$ assigned to the head $h$ and the key embeddings at $h$ from each incoming edge, repeating $H$ times for every head. This generates \textbf{$H$ sets of $n$ raw attention scores}, where each adjacent edge has $H$ attention scores, which are used to scale subsets of the value messages generated at that edge.
    \begin{align}
        \{\mathbf{q}_{i,h}\cdot\mathbf{k}_{ij,h}|j\in [1\dots n]\}_h
    \end{align}
    
    \item Then, we apply the \textbf{softmax function separately to each head} such that the scores for a single head sum to 1. The attention score for a single head $h$ for the edge $ij$ is computed with the equation below:
    \begin{align}
        \alpha_{ij,h}=\frac{\exp\left(\frac{\mathbf{q}_{i,h}\cdot\mathbf{k}_{ij,h}}{\sqrt{d}}\right)}{\sum_{j'\neq i}^n\exp\left(\frac{\mathbf{q}_{i,h}\cdot\mathbf{k}_{ij',h}}{\sqrt{d}}\right)}
    \end{align}
    The attention scores for every head and every edge in the graph are computed in parallel with the \texttt{e\_dot\_v} DGL function, like in single-head attention, except the array of attention scores has shape (edges, $H$):
\begin{lstlisting}[language=Python]
# computes H dot products for each edge and stores as array with shape (edges, H) in edge data with label 'e'
G.apply_edges(fn.e_dot_v('k', 'q', 'e'))

# extract dot products
e = G.edata.pop('e')
# divide each attention score by the full key dimension
e = e / np.sqrt(self.f_mid_in.dimension)
# apply softmax seperately for each head for each set of edges pointing to the same node and store array of attn scores in edge data labeled 'a'
G.edata['a'] = edge_softmax(G, e)
\end{lstlisting}

    \item Similarly to the query and key, we split the value embeddings into $H$ heads, such that for each feature type $l$, there is a corresponding value array with shape (edges, $H$, channels per head, $2l + 1$).
\begin{lstlisting}[language=Python]
for m, d in self.f_value.structure:
    # extract the value embedding for degree d and reshape into (edges, H, channels per head, 2*d+1)
    G.edata[f'v{d}'] = v[f'{d}'].view(-1, self.n_heads, m//self.n_heads, 2*d+1)
\end{lstlisting}
    
    \item Now, we can \textbf{scale each of the $H$ subsets of value embeddings of a single type $l$ by $H$ distinct attention scores} $\alpha_{ij,h}\mathbf{v}^l_{ij,h}$ with the following user-defined function in DGL:
\begin{lstlisting}[language=Python]
def udf_u_mul_e(self, d):
    def fnc(edges):
        # extract attention scores from edge data
        # shape (edges, H) 
        attn = edges.data['a']
        # extracts value array with shape (edges, H, type do channels per head, 2*d+1)
        value = edges.data[f'v{d}']
            
        # reshapes attention scores to (edges, H, 1, 1) which are broadcasted to match value array
        attn = attn.unsqueeze(-1).unsqueeze(-1)
        # broadcasts attn to have the same shape as value and performs element-wise multiplication
        # every component of each channel that shares the same head index is multiplied by the same attention score
        msg = attn * value
        # return dict with msg array with shape (edges, H, type do channels per head, 2*d+1) with label 'm'
        return {'m': msg}
  return fnc
\end{lstlisting}
    
    \item Finally, for every output degree $l$, we \textbf{(1)} take the \textbf{sum over the weighted value messages at each attention head} and then \textbf{(2)} take the \textbf{sum over every incoming edge} to generate the complete type-$l$ neighbor-to-center message:
    \begin{align}
        \mathbf{f}^l_{\text{out},i}=\underbrace{\sum_{j\neq i}^n\underbrace{\sum_{h=1}^H\alpha_{ij,h}\mathbf{v}^l_{ij,h}}_{\text{(1) sum over heads}}}_{\text{(2) sum over all adjacent nodes}}
    \end{align}
    The code below loops through all the output degrees, scales the output value messages from the same \texttt{udf\_u\_mul\_e} function as the key embedding, takes the sum of the messages across all attention heads and incoming edges, and finally updates the node features with the aggregated messages.
\begin{lstlisting}[language=Python]
for do in self.f_out.degrees:
    # generates msg and sums over attention heads and edges
    G.update_all(self.udf_u_mul_e(do), fn.sum('m', f'out{do}'))
\end{lstlisting}
    
    We also return a dictionary where the keys correspond to the output degrees and the values are arrays with shape (nodes, type-$l$ channels, $2l+1$) to use for attentive self-interaction:
\begin{lstlisting}[language=Python]
# initialize dict
output = {}
for mo, do in self.f_out.structure:
    # extract weighted messages from node data and add dimension to separate channels of the same type
    output[f'{do}'] = G.ndata[f'out{do}'].view(-1, mo, 2*do+1)
return output
\end{lstlisting}
    
\end{enumerate}

Now, we will discuss how to incorporate a skip connection and project the value messages to match the output fiber using attentive self-interaction.

\subsection{Attentive Self-Interaction}
\label{subsec:att-self-interact}
\purple[]{
The SE(3)-Transformer paper introduced the \textbf{attentive self-interaction} layer, which generates self-interaction weights by taking the dot product between pairs of type-$l$ input features and value messages and feeding these raw attention scores into a feed-forward network. Every type-$l$ output channel is a weighted sum of every type-$l$ input channel and every type-$l$ value message scaled by a set of attention scores, capturing patterns across input features from the center node itself and neighbor-to-center messages to generate the output feature tensor.
}

\begin{figure}[h!]
\centering
\includegraphics[width=\linewidth]{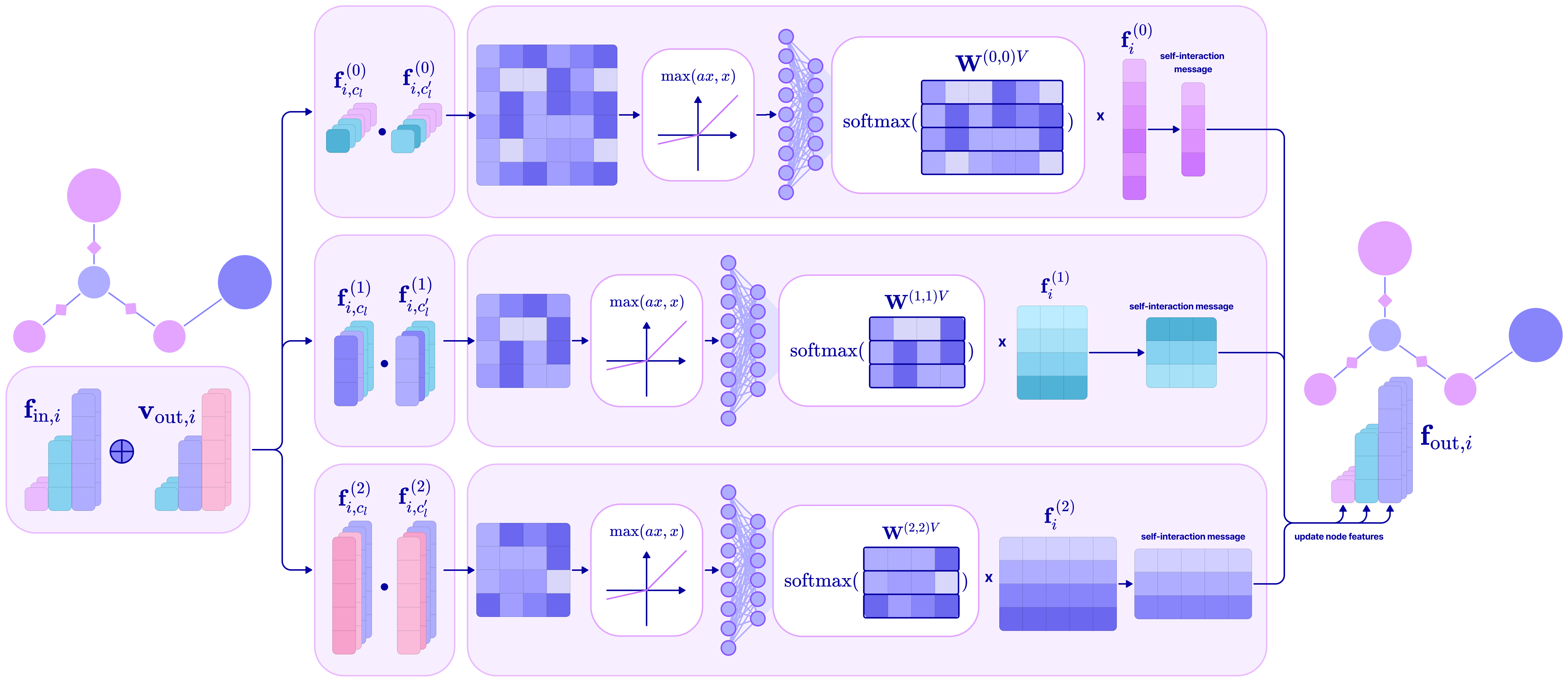}
\caption{Diagram showing the process of computing the attentive-self-interaction message for type-0, type-1, and type-2 output messages to node $i$. First, we concatenate the input features from node $i$ to the value messages from the self-attention layer. Next, we take the dot product between every pair of features with the same degree in the concatenated tensor, producing a square matrix of raw attention scores for each degree. Then, the scores are fed into a leaky ReLU function and an FFN to output an updated matrix of weights with columns corresponding to catenated channels and rows corresponding to output channels. The softmax function is applied to each row to get the final matrix of attention scores, and we take its product with the matrix of concatenated features. The output tensor list is used to update node $i$.}
\label{fig:att-self-interact}
\end{figure}

Attentive self-interaction is a mix of linear self-interaction and channel mixing, as it takes a weighted sum of \textbf{all input channels and all value message channels of the same type} to generate the final output tensor, repeating for every output channel with a unique set of attention scores. But instead of initializing the weights randomly and sharing weights across nodes, attentive self-interaction leverages the attention mechanism to compute unique weights for each node and each input-to-output channel pair.

Before generating the attention weights, we take the saved input features of node $i$ before any updates are made and concatenate them to the output of the multi-head self-attention layer. To do this, we define a class \texttt{GCat} that \textbf{concatenates all the type-$l$ channels in the input fiber to the type-$l$ channels of the value message for every feature type}. We can define an instance variable in the constructor of \texttt{GCat} called \texttt{f\_cat} that is a fiber object with the same structure as the fiber of value messages \texttt{f\_mid\_out} added to the multiplicities of the fiber of input features \texttt{f\_in}:

\begin{lstlisting}[language=Python]
# initialize dictionary of degree:multiplicity key:value pairs
f_cat = {}
# loop through all degrees in value message fiber
for d in f_mid_out.degrees:
    # add the multiplicity of type d in value msg fiber to f_cat
    f_cat[d] = f_mid_out.dict[d]
    # loop through all degrees in input fiber
    if d in f_in.degrees:
		# add the multiplicity of type d in input fiber to f_cat
        f_cat[d] += f_in.dict[d]
# set the instance variable to a fiber with structure f_cat
self.f_cat = Fiber(dictionary=f_cat)
\end{lstlisting}

The forward function of \texttt{GCat} concatenates the input features to the value messages for all degrees in \texttt{f\_in} and \textbf{returns the concatenated feature tensor} in a dictionary called \texttt{out}, where each degree is a label linked to a feature array with shape (nodes, mcat, $2l+1$), where mcat denotes the total number of type-$l$ channels in the concatenated tensor.

\begin{lstlisting}[language=Python]
# forward function of GCat module
def forward(self, msg, inpt):
  	# initialize dictionary of degree:(mcat, 2*do+1) key:value pairs
  	out = {}
  	# loop through all degrees in f_cat
  	for do in self.f_out.degrees:
  	    do = str(do)
  	    # if degree is in input fiber, concatenate all type-do input and msg features along the channel dimension
  	    if do in inpt:
  	        out[do] = torch.cat([inpt[do], msg[do]], 1)
  	    # if degree is not in input fiber, just set the features to the msg features
  	    else:
  	        out[do] = msg[do]
  	return out
\end{lstlisting}

Now, we can use the concatenated dictionary of features to compute attentive self-interaction and \textbf{project the concatenated feature tensor to match the structure of the output fiber of the attention layer}, \texttt{f\_out}. The attentive self-interaction layer generates the type-$l$ output feature at channel $c_l$ as the \textbf{weighted sum of all the type-$l$ channels in the concatenated fiber}, with the equation below:

\begin{align}
    \mathbf{f}^l_{\text{out},i,c_l}=\sum_{c_l'}w^{l l}_{(i,c_l,c_l')}\mathbf{f}^l_{\text{cat},i,c_l'}
\end{align}
where $c_l’$ are the indices of the type-$l$ features in the concatenated fiber and $c_l$ denotes the output type-$l$ channel to which the weighted sum is passed.

\textbf{Let’s break down the steps to generating the attentive self-interaction term for all the type-$l$ output channels for node $i$.}

First, we generate the set of raw attention scores by taking the \textbf{dot product} of \textbf{every pair of type-$l$ channels $c_l$ and $c_l’$ in the concatenated tensor}. Since mcat denotes the total number of type-$l$ input channels, a total of mcat*mcat raw attention scores are generated for the type-$l$ features that can be assembled into a (mcat*mcat)-dimensional vector.

\begin{align}
    \bigoplus_{c_l}\bigoplus_{c_l'}\mathbf{f}^{l}_{\text{cat,}i,c_l}\cdot \mathbf{f}^l_{\text{cat,}i,c_l'}
\end{align}

To implement this in code, we loop through every degree and corresponding feature array with shape $(\text{mcat}, 2l+1)$ stored in the dictionary. At each iteration, we use PyTorch’s \texttt{einsum} function to calculate the sum of the element-wise product along the last dimension (dot product) for every pair of type-$l$ features along the second last dimension of the type-$l$ feature array, generating an $\text{mcat} \times \text{mcat}$ matrix of raw attention scores.

\begin{lstlisting}[language=Python]
# f: array containing all type-l channels with shape (nodes, mi, 2l+1)
# calculate the dot product of all channel pairs in f and stores them in scalars
# a, b: indexes along the channel axis
# c: index along the tensor-component axis
scalars = torch.einsum('...ac,...bc->...ab', [f, f])
\end{lstlisting}

Then, we flatten the scores into a (mcat$^2$)-dimensional vector and ensure that the absolute value of each score is greater than a small constant ($10^{-12}$) to avoid vanishing gradients.

\begin{lstlisting}[language=Python]
# flattens the square matrix into a vector for each node
scalars = scalars.view(*first_dims, mi*mi)
# stores the signs of each attention score
sign = scalars.sign()
# clamps the absolute value of each score to a minimum of 1e-12 to prevent vanishing gradients
scalars = scalars.abs_().clamp_min(self.eps)
# reapplies the signs to each score
scalars *= sign
\end{lstlisting}

Since the number of output channels (mo) is not always equal to the number of concatenated channels, we need to generate a total of mcat*mo weights for every degree $l$ such that each type-$l$ output channel has a set of mcat weights that are used to calculate a weighted sum of all the type-$l$ input features and value messages.

To convert the (mcat*mcat)-dimensional vector into a (mcat*mo)-dimensional vector of weights, we construct a FFN for each feature type with a layer normalization, a non-linear activation function applied element-wise to each attention score, and a linear layer that transforms the (mcat*mcat)-dimensional vector into a (mcat*mo)-dimensional vector. To construct the FNN for each feature degree, we call the function below:

\begin{lstlisting}[language=Python]
def _build_net(self, mcat: int, mo: int):
    # dimension of hidden layer
    n_hidden = mcat * mo
    # input dimension
    cur_inpt = mcat * mcat
    # initializes empty list to store network layers
    net = [] 
    # loop to construct num_layers-1 linear layers
    # num_layers defaulted to 2 to construct a single linear layer
    for i in range(1, self.num_layers):
		# layer normalization of input vector
        net.append(nn.LayerNorm(int(cur_inpt))) 
        # leaky ReLU nonlinearity
        net.append(self.nonlin) 
        # fully connected layer that transforms input dimension to n_hidden and adds bias term only to the last layer
        net.append(
            nn.Linear(cur_inpt, n_hidden, bias=(i == self.num_layers - 1)))
        # initializes weights matrix using Kaiming initialization
        nn.init.kaiming_uniform_(net[-1].weight)
        # sets the input to the next layer to be the output dimension of the current layer
        cur_inpt = n_hidden
    return nn.Sequential(*net)
\end{lstlisting}

We then loop through all the degrees in the concatenated fiber and use the function above to initialize the FFN with the hidden layer dimension set to the mcat*mo:

\begin{lstlisting}[language=Python]
self.transform = nn.ModuleDict()
# loop through every degree:multiplicity pair in f_cat
for do, mcat in self.f_cat.structure_dict.items():
	# extract number of output channels for degree do
    mo = self.f_out.structure_dict[do]
    # build FFN with hidden dimension mcat*mo
    self.transform[str(do)] = self._build_net(mcat, mo)
\end{lstlisting}

Then, the flattened vector of raw attention scores for degree $l$ is fed into the unique type-$l$ FFN to generate an updated set of mcat*mo attention scores with added dependence on learnable parameters and nonlinearities. This vector is reshaped into a mo $\times$  mcat matrix, where each row corresponds to the set of attention scores used to compute self-interaction for a single type-$l$ output channel.

\begin{align}
    \mathbf{W}^{l l V}=\text{FFN}_l\left(\bigoplus_{c_l}\bigoplus_{c_l'}\mathbf{f}^{l}_{\text{cat,}i,c_l}\cdot \mathbf{f}^l_{\text{cat,}i,c_l'}\right)
\end{align}

\begin{lstlisting}[language=Python]
# extract the type do FFN feed the raw attn scores in as input 
att_weights = self.transform[str(do)](scalars)
# reshape the output vector into a mo x mcat matrix
att_weights = att_weights.view(*first_dims, mo, mcat)
\end{lstlisting}

We can now apply the softmax function separately to each row of the attention score matrix (along the last dimension) such that the scores of each row add up to 1. \textbf{This normalizes the set of weights used to scale all the concatenated channels for a single type-$l$ output channel into probabilities that sum to 1.}

\begin{align}
    w^{l l}_{i,c_l,c_l'}\mapsto \frac{\exp(w^{l l}_{i,c_l,c_l'})}{\sum_{c_l'}\exp(w^{l l}_{i,c_l,c_l'})}
\end{align}

\begin{lstlisting}[language=Python]
import torch.nn.functional as F

# apply softmax along the last dimension of the attn weights array of shape (nodes, mo, mcat)
att_weights = F.softmax(input=att_weights, dim=-1)
\end{lstlisting}

Finally, we can generate a \textbf{self-interaction message for each output channel $c_l$ with the weighted sum of every type-$l$ input feature and value message scaled by their attention score}. The attention scores measure the similarity of the channel feature to all the other channels of the same type, which produces a self-interaction message that effectively captures patterns from both the input features of the node itself and the value messages from neighboring nodes. 

This step is performed in parallel for every output channel of a single degree via matrix-matrix multiplication between the mo $\times$ mcat matrix of attention weights and the mcat $\times (2l+1)$ matrix of type-$l$ features in the concatenated feature array.

\begin{align}
    \mathbf{W}^{l l V}\mathbf{f}^l_{\text{cat},i}=\begin{bmatrix}w^{l l}_{(1,1)}&w^{l l}_{(1,2)}&\dots&w^{l l}_{(1,\text{mcat})}\\\\w^{l l}_{(2,1)}&\ddots&&\vdots\\\\\vdots&&\ddots&\vdots\\\\w^{l l}_{(\text{mo},1)}&\dots&\dots&w^{l l}_{(\text{mo,mcat})}\end{bmatrix}_{(\text{mo,mcat})}\begin{bmatrix}\mathbf{f}^l_{\text{cat},i,1}\\\\\mathbf{f}^l_{\text{cat},i,2}\\\\\vdots\\\\\vdots\\\\\mathbf{f}^l_{\text{cat},i,\text{mcat}}\end{bmatrix}_{(\text{mcat,}2l+1)}
\end{align}

The matrix multiplication can be implemented using the \texttt{einsum} function that takes the \textbf{sum of the element-wise products} of the elements along the columns of the attention matrix with the elements along the rows of the feature matrix to produce a (mo, $2l+1$) array of self-interaction messages for every type-$l$ output channel.

\begin{lstlisting}[language=Python]
# attn_weights has shape (nodes, mo, mcat)
# f has shape (nodes, mcat, 2*do+1)
# each element of output has shape (nodes, mo, 2*do+1)
output[do] = torch.einsum('...nm,...md->...nd', [att_weights, f])
\end{lstlisting}

Finally, we can \textbf{update the feature tensor at node $i$} with the attentive self-interaction messages that match the output fiber structure, \texttt{f\_out}:

\begin{align}
    \mathbf{f}^l_{\text{out},i}=\mathbf{W}^{l l V}\mathbf{f}^l_{\text{cat},i}
\end{align}

The purpose of the attentive self-interaction is \textbf{fourfold}:

\begin{enumerate}
    \item In the sequence-based Transformer model, every element in the sequence attends to all other elements, including itself. Since the neighbor-to-center message does not include features from the center node, attentive self-interaction introduces a way for the node to attend to its own features, allowing the model to learn the contributions of the features of a single atom or molecular unit to the prediction task.

    \item Unlike linear self-interaction and channel-mixing, which use a single weight for each feature channel shared across all nodes, \textbf{attentive self-interaction generates unique weights for each node} to produce self-interaction messages that reflect the contextual patterns across input features and value messages of each node.
    
    \item The attention scores place additional weight on features that are similar to the other features of the same type to ensure that a \textbf{cohesive set of output features is used to update the center node}.
    
    \item Since the value messages primarily carry information from neighborhood nodes and edges, \textbf{concatenating the input features to the value messages introduces a skip connection} that ensures that the original node information is not lost. The attention weights process the concatenated features with learnable parameters to determine the \textbf{relative importance of each piece of information}. The skip connection also allows the gradient to skip past the potentially gradient-diminishing message-calculation transformations during backpropagation to avoid vanishing gradients.
\end{enumerate}

Furthermore, the attentive self-interaction weights and the weighted sum operation satisfy the rules for equivariant layers defined in Section \ref{subsec:rules-equiv}.

Now that we have defined all the components of the attention block, we can initialize all the relevant module objects in the constructor of the \texttt{GSE3Res} module:

\begin{lstlisting}[language=Python]
# in constructor
# initialize dictionary of attention modules
self.GMAB = nn.ModuleDict()

# modules for computing query, key, value embeddings 
self.GMAB['v'] = GConvSE3Partial(f_in, self.f_mid_out, edge_dim=edge_dim, x_ij=x_ij)
self.GMAB['k'] = GConvSE3Partial(f_in, self.f_mid_in, edge_dim=edge_dim, x_ij=x_ij)
self.GMAB['q'] = G1x1SE3(f_in, self.f_mid_in)

# module for attention calculations with value fiber set to f_mid_out and key and query fiber set to f_mid_in
self.GMAB['attn'] = GMABSE3(self.f_mid_out, self.f_mid_in, n_heads=n_heads)
# define cat function to concatenate value fiber and input fiber
self.cat = GCat(self.f_mid_out, f_in)
# define project function to project f_cat to f_out with attentive self-interaction
self.project = GAttentiveSelfInt(self.cat.f_cat, f_out)
\end{lstlisting}

In the forward pass, these functions are called in the following order to generate the updated feature tensors for every node in the graph:

\begin{lstlisting}[language=Python]
def forward(self, features, G, **kwargs):
    # generate queries, keys and values for all nodes and edges in G
    v = self.GMAB['v'](features, G=G, **kwargs)
    k = self.GMAB['k'](features, G=G, **kwargs)
    q = self.GMAB['q'](features, G=G)

    # calculate attention value messages
    z = self.GMAB['attn'](v, k=k, q=q, G=G)
    # concatenate the value messages z to the input features
    z = self.cat(z, features)
    # project concatenated messages to f_out
    z = self.project(z)
    
    # return a dictionary with output degrees as labels linked to an array of the output feature tensor for every node in the graph with shape (nodes, mo, 2*do+1)
    return z
\end{lstlisting}

\subsection{Norm Nonlinearity}
\label{subsect:norm-nonlin}
Following each attention block, the SE(3)-Transformer incorporates an \textbf{SE(3)-equivariant ReLU normalization and nonlinearity module} with a feed-forward network. This module \textbf{stabilizes the distribution of features across layers} for smoother gradient flow and adds an extra degree of nonlinearity to the attention block without breaking equivariance.

The norm nonlinearity layer is applied to \textbf{each feature type separately}. The following steps are performed on all type-$l$ output features in a single node in the graph:

\begin{enumerate}
    \item The Euclidean norm (L2 norm) is applied across all type-$l$ features, which calculates the magnitude (or length) of each type-$l$ feature vector in $(2l+1)$-dimensional Euclidean space. All the norms of the type-$l$ feature channels are stored in a single mo-dimensional vector, where mo denotes the number of type-$l$ channels in the output fiber of the attention block. The norm for a single type-$l$ channel is given by the equation below, where the double-bar and subscript denote the L2 norm:
    \begin{align}
        \|\mathbf{f}^l_{c_l}\|_2=\sqrt{\sum_{m=-l}^l(f^l_m)^2}
    \end{align}
    The following code generates an array with shape (mo, $2l+1$) where all the elements along the $c_l$th row are equal to the norm corresponding to the channel $c_l$, in preparation for element-wise division.
\begin{lstlisting}[language=Python]
# f: array of type-l features with shape (mo, 2l+1)
# self.eps: small constant 1e-12 to avoid division by zero
# clamps the norm values to a min value 1e-12 and then expands the norm value across the last dimension of f
norm = f.norm(2, -1, keepdim=True).clamp_min(self.eps).expand_as(f)
\end{lstlisting}
    
    \item Then, each type-$l$ feature is \textbf{divided element-wise by the corresponding norm} to generate a \textbf{unit-length vector in the direction of the feature} (similar to generating the angular component of the displacement vector).

    \begin{align}
        \frac{\mathbf{f}_{c_l}^l}{\|\mathbf{f}_{c_l}^l\|_2}=\begin{bmatrix}\frac{f^l_{-l}}{\|\mathbf{f}_{c_l}^l\|_2}\\\vdots\\\frac{f^l_l}{\|\mathbf{f}_{c_l}^l\|_2}\end{bmatrix}
    \end{align}
\begin{lstlisting}[language=Python]
# extract unit-length features 
phase = f / norm
\end{lstlisting}
    
    \item To incorporate learnability and nonlinearity to the normalization of type-$l$ output, we construct an \textbf{FFN that takes the norms across all type-$l$ output channels and applies the following operations}: \textbf{(1)} a layer normalization step that shifts the mean to 0 and variance to 1 across all type-$l$ norms, \textbf{(2)} a ReLU nonlinearity that converts all negative norms from step 1 to 0 while keeping all positive norms the same and \textbf{(3)} an (optional) linear layer that transforms the norms without changing the dimension through matrix multiplication with learnable weights and addition of a bias vector.
    \begin{align}
        \text{FFN}(\|\mathbf{f}^l\|_2)=\underbrace{\mathbf{W}_{(\text{mo}\times\text{mo)}}\underbrace{\max\left(0,\text{LayerNorm}(\underbrace{\bigoplus_{c_l}\|\mathbf{f}^l_{c_l}\|_2}_{\text{all type-}l\text{ norms}})\right)}_{\text{ReLU nonlinearity}}+\mathbf{b}}_{\text{linear layer}}
    \end{align}
    This FFN can be built by calling the function below:
\begin{lstlisting}[language=Python]
def _build_net(self, mo: int):
    net = []
    # if num_layers > 1 construct FFN with linear layers
    for i in range(self.num_layers):
        # normalize the norms across all channels of the same type 
        net.append(BN(int(mo)))
        # ReLU nonlinearity
        net.append(self.nonlin)
        # linear layer with input and output dim set to mo
        net.append(nn.Linear(mo, mo, bias=(i==self.num_layers-1)))
        nn.init.kaiming_uniform_(net[-1].weight)
        
    # if num_layers = 0 construct FFN with only normalization and ReLU step
    if self.num_layers == 0:
        net.append(BN(int(mo)))
        net.append(self.nonlin)
    return nn.Sequential(*net)
\end{lstlisting}
    
    We call the function above in a loop to construct a unique FFN for every feature type:
    \begin{lstlisting}[language=Python]
    self.transform = nn.ModuleDict()
    for mo, do in self.fiber.structure:
        self.transform[str(do)] = self._build_net(int(mo))
    \end{lstlisting}
    
    \item Next, we call the transform function on the mo-dimensional vector of norms, which generates a weight for each type-$l$ channel. Then, we add a singleton dimension that is broadcast for element-wise multiplication in the next step.
\begin{lstlisting}[language=Python]
# norm[...,0] converts the expanded norms back to a single vector
# call transform function on the vector of norms and adds a singleton dimension
transformed = self.transform[str(k)](norm[...,0]).unsqueeze(-1)
\end{lstlisting}
    \item Finally, we scale each type-$l$ channel divided by its L2 norm with the corresponding nonlinear weight generated from the FFN to get the final normalized feature at channel $c_l$.
    \begin{align}
        \text{NormNonlinear}(\mathbf{f}^l_{c_l})=\text{FFN}(\|\mathbf{f}^l\|_2)_{c_l}\begin{bmatrix}\frac{f^l_{-l}}{\|\mathbf{f}_{c_l}^l\|_2}\\\vdots\\\frac{f^l_l}{\|\mathbf{f}_{c_l}^l\|_2}\end{bmatrix}
    \end{align}
\begin{lstlisting}[language=Python]
# broadcast transformed weights to have shape (mo, 2l+1) and multiply elementwise with normalized features
output[do] = (transformed * phase).view(*v.shape)
\end{lstlisting}
\end{enumerate}

\subsection{Incorporating Edge Features}
\label{subsec:edge-features}
\textbf{Edge features} can store both information on the interaction between nodes and the nodes themselves. Previously, we discussed how to incorporate the \textbf{angular unit displacement vector} (type-1 edge feature) and \textbf{radial distance} (type-0 edge feature) via learnable functions that allow the model to learn dependencies on the spatial positioning of nodes.

Graphs can also contain other edge features (e.g., bond type, shared electrons, etc.) that can be incorporated into equivariant message-passing layers in the following ways:

\begin{enumerate}
    \item If the \textbf{edge feature is a scalar} (type-0 tensor), it can be used to \textbf{scale basis kernels during kernel construction}, similar to the radial functions. This can be done by extending the learnable radial FFN to take multiple scalars (in the form of a vector) as input. This would allow the radial network to learn dependencies between the radial distance and additional scalar edge features and produce kernel weights dependent on these additional edge features.
    \begin{align}
    \mathbf{W}^{l k}(\mathbf{x}_{ij})=\sum_{J=|k-l|}^{k+l}\varphi^{l k}_J(\|\mathbf{x}_{ij}\|, \mathbf{f}_{ij}^{(0)})\mathbf{W}^{l k}_J(\mathbf{x}_{ij})\tag{$\mathbf{f}_{ij}^{(0)}\in \mathbb{R}$}
    \end{align}
    The function denoted by phi takes two type-0 spherical tensors, including radial distance, and converts them into a scalar weight given by $\varphi^{l k}_J:\mathbb{R}^2\to\mathbb{R}$. The code below \textbf{concatenates type-0 edge features to the radial distance} and feeds the vector of type-0 features into the radial network to generate the unique set of weights used to scale the basis kernels:
\begin{lstlisting}[language=Python]
# extract type-0 features from edge data 
w = G.edata['w'] 
# concatenate it with the radial distance along the last dimension to form a vector of type-0 edge features
feat = torch.cat([w, r], -1)

# use feat as input to radial functions to generate weight kernels for each input and output degree pair
for (mi, di) in self.f_in.structure:
    for (mo, do) in self.f_out.structure:
        etype = f'({di},{do})'
        # this line generates a kernel using the radial function that takes feat as input
        G.edata[etype] = self.kernel_unary[etype](feat, basis)
\end{lstlisting}
    
    \item An alternative way to incorporate scalar edge features is to \textbf{construct separate learnable FFNs} specifically trained to learn dependencies on the specific edge feature, producing a scalar weight that is used in conjunction with the radial weight to scale the basis kernels.
    \begin{align}
        \mathbf{W}^{l k}(\mathbf{x}_{ij})=\sum_{J=|k-l|}^{k+l}\psi^{l k}_J(\mathbf{f}_{ij}^{(0)})\varphi^{l k}_J(\|\mathbf{x}_{ij}\|)\mathbf{W}^{l k}_J(\mathbf{x}_{ij})\tag{$\mathbf{f}_{ij}^{(0)}\in \mathbb{R}$}
    \end{align}
    Psi ($\Psi$) denotes the FFN that takes a scalar edge feature to a set of ($2\min(l, k)+1$)*(input channels)*(output channels) weights used to scale each basis kernel for every pair of input and output channels:
    \begin{align}
        \psi^{l k}(\mathbf{f}_{ij}^{(0)}):\mathbb{R}\to\mathbb{R}^{(2min(l,k)+1)(\text{mi})(\text{mo})} 
    \end{align}
    
    \item \textbf{Higher-degree ($k \geq 1$) edge features can be transformed from type-$k$ to type-$l$ tensors} via equivariant kernels and then used in the self-attention mechanism just like the node features. This is done by \textbf{concatenating the edge features to the features of the source node of the same type} just before they are transformed into type-$l$ key embeddings for computing attention weights and value embeddings for computing the neighbor-to-center messages.
\end{enumerate}
To incorporate edge features, the SE(3)-Transformer module \textbf{concatenates the full relative displacement vector} (as opposed to the unit displacement vector used as input to spherical harmonics) as a type-1 edge feature \textbf{with the type-1 node features} before calculating the key and value embeddings.

\begin{figure}[h!]
\centering
\includegraphics[width=0.6\linewidth]{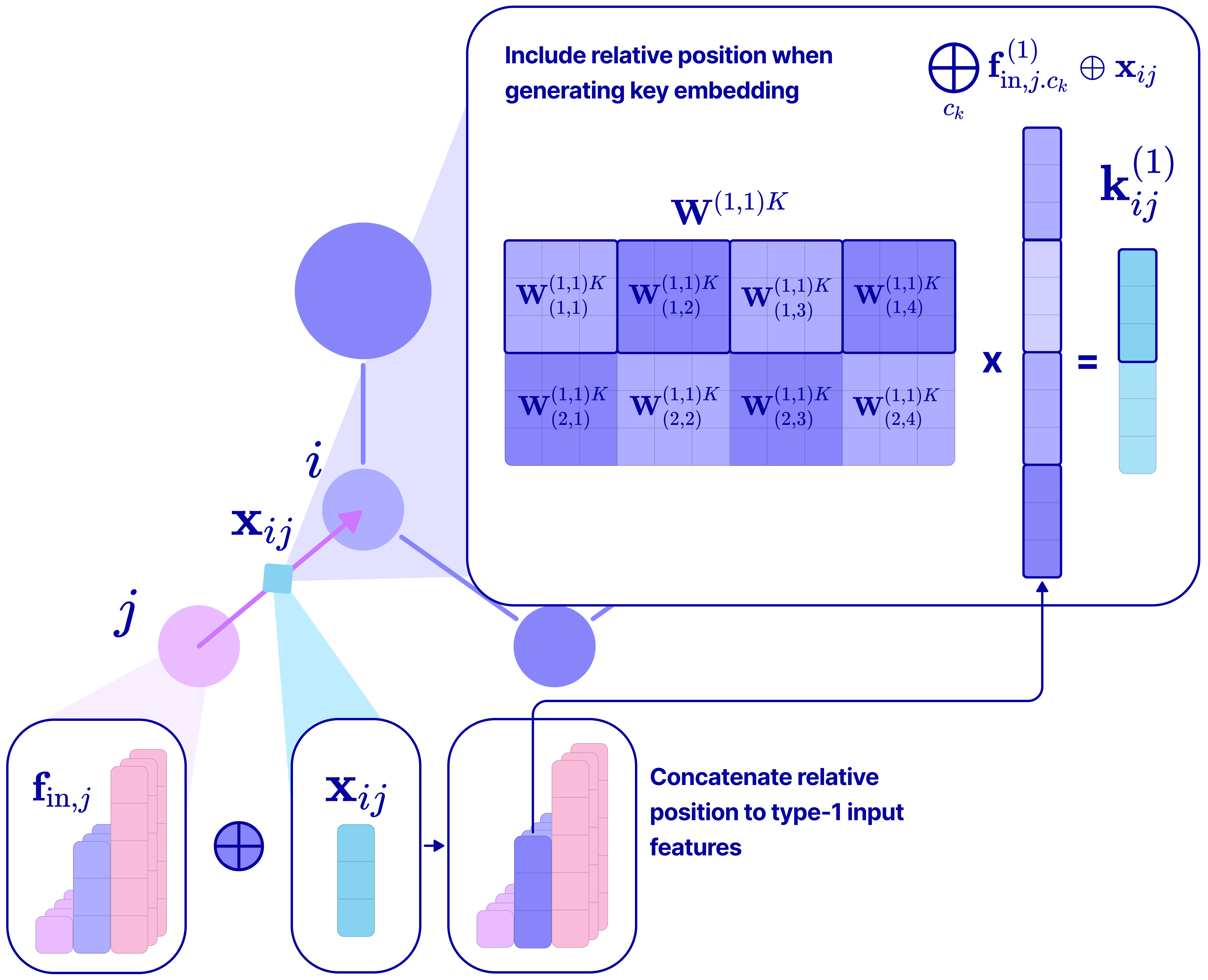}
\caption{Diagram showing the concatenation of the type-1 displacement vector to the input features from node $j$. The concatenated type-1 tensors are multiplied by the key kernel that transforms between type-1 input and type-1 output tensors to generate a part of the type-1 key embeddings, which will then be added to the type-1 key embeddings transformed from all other input degrees. Note that the concatenated type-1 features are also used to generate the key embeddings of every degree defined in the key fiber.}
\label{fig:displacement}
\end{figure}

First, the multiplicity for degree 1 in the input fiber is incremented by 1 if the displacement is concatenated (multiplicity is unchanged for addition). 

\begin{lstlisting}[language=Python]
# in constructor
# adding/concatinating relative position input fiber to generate key/value embeddings
# 'cat' concatenates relative position & existing feature vector
# 'add' adds relative position to , but only if multiplicity is at least 1
assert x_ij in [None, 'cat', 'add']
self.x_ij = x_ij
if x_ij == 'cat':
	# increments the multiplicity of degree 1 in the input fiber by 1 to account for type-1 relative position
    self.f_in = Fiber.combine(f_in, Fiber(structure=[(1,1)]))
else:
    self.f_in = f_in
\end{lstlisting}

Now, we can \textbf{incorporate the type-1 displacement vector stored in the edge either through concatenation or addition} to the type-1 input features \textit{before} multiplying with the key and value kernels and generating the key and value embeddings:

\begin{lstlisting}[language=Python]
# in udf that generates the key and value arrays for a single output degree do
def fnc(edges):
	# neighbor -> center messages
	msg = 0
	# loop through all input degrees in the input fiber
	for mi, di in self.f_in.structure:
	    # executes when the loop reaches degree 1, and the flag is set to 'cat'
	    if self.x_ij == 'cat' and d_in == 1:
	        # calculate type-1 relative position vector between source   and destination node
	        rel = (edges.dst['x'] - edges.src['x']).view(-1, 3, 1)
	        # since the multiplicity of degree 1 is incremented in the constructor, m_ori is the original num of type-1 channels
	        m_ori = mi - 1
	        if m_ori == 0:
	            # if there are no type-1 channels, just use relative position
	            src = rel
	        else: # if other type-1 channels exist in the source node, concatenate rel to channels
	            # stack all channels into a single dimension
	            src = edges.src[f'{d_in}'].view(-1, m_ori*(2*di+1), 1)
	            # concatenate the relative position vector to type-1 channels
	            src = torch.cat([src, rel], dim=1)
	    # executes if flag is set to 'add' and at least 1 type-1 feature
	    elif self.x_ij == 'add' and d_in == 1 and m_in > 1:
	        src = edges.src[f'{d_in}'].view(-1, m_in*(2*d_in+1), 1)
	        # calculate relative position and reshape to (3, 1) 
	        rel = (edges.dst['x'] - edges.src['x']).view(-1, 3, 1)
	        # add to the first type-1 feature at source node
	        # :3, :1 adds to the first three elements of the second last dim and first of the last dim
	        src[..., :3, :1] = src[..., :3, :1] + rel
	    else:
	        src = edges.src[f'{d_in}'].view(-1, mi*(2*di+1), 1)
	    # extract key or value kernel that transforms types di to do 
	    edge = edges.data[f'({di},{do})']
	    # multiply kernel with shape (do*mo, 2*do+1) with all input features of degree do of shape (mi*(2*di+1), 1) 
	    msg = msg + torch.matmul(edge, src)
	msg = msg.view(msg.shape[0], -1, 2*do+1)
\end{lstlisting}

\newpage
\section{Chemical Property Prediction}
\label{sec:6}
\purple[]{
To illustrate how the spherical equivariant graph transformers can be used in practice, we describe an experiment in \citet{fuchs2020se} that applies it for \textbf{prediction of system-level chemical properties}, which are invariant to translations and rotations in 3-dimensions, on the QM9 dataset \citep{ramakrishnan2014quantum} of 134K small molecules. 
}

\begin{figure}[h!]
\centering
\includegraphics[width=\linewidth]{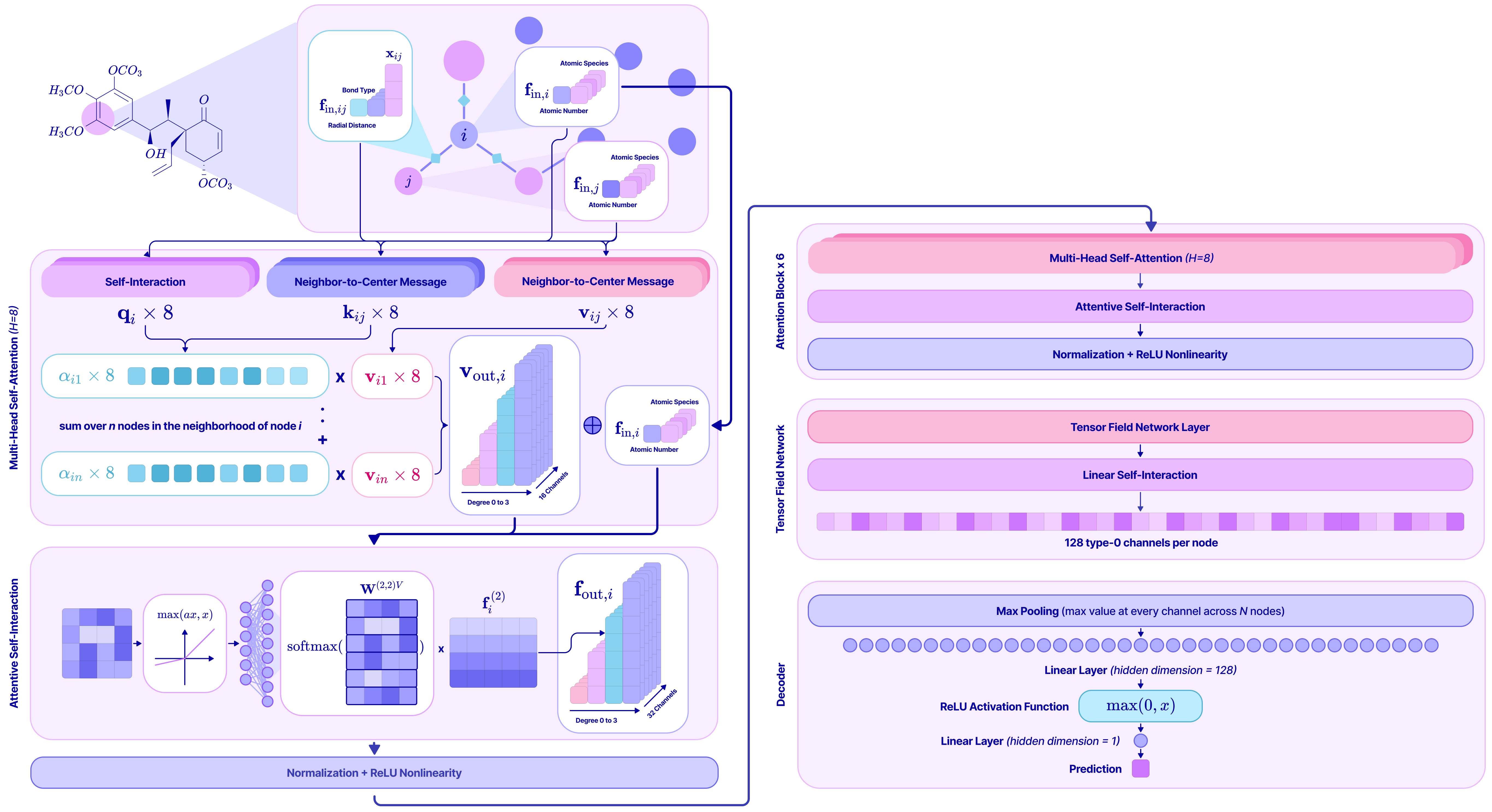}
\caption{The full SE(3)-Transformer architecture used for chemical property prediction on the QM9 dataset.}
\label{fig:chemical}
\end{figure}

Molecular property prediction provides a natural setting for equivariant models, as most quantum-mechanical properties of molecules are invariant to global rotation and translation, and the interactions between atoms depend on their relative geometry. In this section, we walk through the full modeling pipeline, from constructing a geometric molecular graph to encoding features with an SE(3)-Transformer to decoding invariant predictions.

\subsection{Initializing the Molecular Graph}
We represent each molecule as a geometric graph, where nodes correspond to atoms and edges encode pairwise bonding relationships. 

\begin{figure}[h!]
\centering
\includegraphics[width=0.8\linewidth]{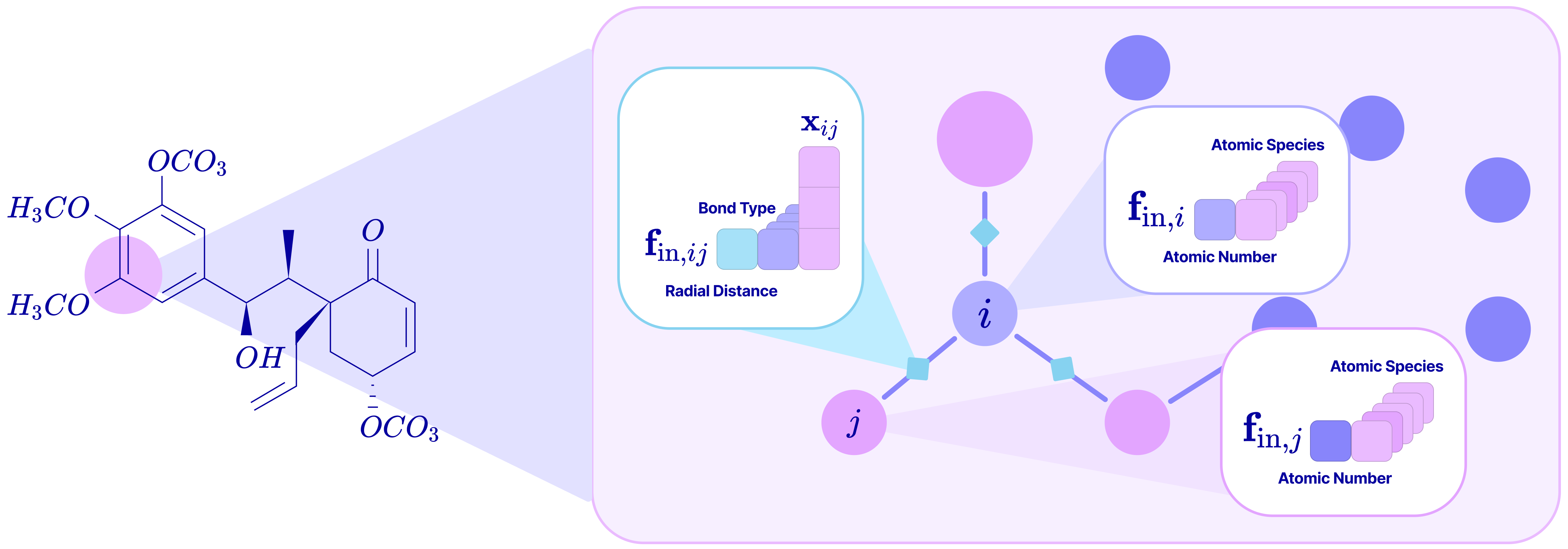}
\caption{The small molecule SMILES strings are converted into a graph where the nodes represent atoms, and the edges represent covalent bonds between atoms. The initial node feature tensor consists of 6 type-0 channels storing the number of protons and the type of atom. The initial edge feature tensor consists of 5 type-0 channels storing the radial distance and bond type, and one type-1 channel with the relative displacement vector.}
\label{fig:chem1}
\end{figure}

Each \textbf{node of the graph represents an atom in the molecule}, which is represented by a 6-dimensional embedding that encodes the following \textbf{node features}:

\begin{enumerate}
    \item The \textbf{atomic species} (Hydrogen, Carbon, Oxygen, Nitrogen, or Fluorine) is represented as a 5-dimensional one-hot-embedding vector with a 1 corresponding to the species of the node and zeros everywhere else.
    
    \item The \textbf{atomic number} or number of protons in the atom is represented as a scalar integer value.
\end{enumerate}

All node features are type-0 tensors, so there are a total of 6 type-0 channels in the initial node feature tensor.

The graph is a sparse molecular graph where \textbf{only the bonded atoms are connected by a bidirectional edge}, meaning the nodes at either end of the edge can send and receive messages to each other. Each \textbf{edge} is represented with a 5-dimensional vector that encodes the following \textbf{edge features}:

\begin{enumerate}
    \item The \textbf{type of chemical bond} (single, double, triple, or aromatic bond) is represented as a 4-dimensional one-hot encoding vector corresponding to the bond type between the nodes that it connects.
    
    \item The \textbf{Euclidean distance between atoms} is represented as a scalar.
\end{enumerate}

All edge features are also type-0 tensors, so there are a total of 5 type-0 channels in the initial feature tensor corresponding to each edge. The code below initializes the DGL graph using two arrays storing the source and destination node indices that are aligned such that a position $i$ in the source array and the destination array corresponds to a single edge in the graph. Then, the node and edge features are stored as arrays that can be accessed with the corresponding label.

\begin{lstlisting}[language=Python]
# src: array of source node ids with length corresponding to the num of edges
# dst: array of source node ids with length corresponding to the num of edges
# initialize dgl graph with edges 
G = dgl.DGLGraph((src, dst))

# add 3-dimensional positions of atoms for all atoms to node data with label 'x'
G.ndata['x'] = torch.tensor(x) # shape (num_atoms, 3)
# add atomic species and atomic number as 6 separate scalars to node data with label 'f'
G.ndata['f'] = torch.tensor(np.concatenate([one_hot, atomic_numbers], -1)[...,None]) # (num_atoms, 6, 1)

# add edge features to graph
# add 3-dimensional displacement vector between src and dst nodes to edge data with label 'd'
G.edata['d'] = torch.tensor(x[dst] - x[src]) # (num_atoms, 3)
# add 4-dimensional one-hot encoding of bond type to edge data labeled 'w'
G.edata['w'] = torch.tensor(w) # (num_atoms, 4)
\end{lstlisting}

\subsection{Encoder}
\purple[]{
The encoder consists of equivariant multi-head attention blocks, each of which is followed by an equivariant norm nonlinearity layer. The output is a learned internal representation for each feature channel of each atom that encodes dependencies between bonded atoms.
}

\begin{figure}[h!]
\centering
\includegraphics[width=0.8\linewidth]{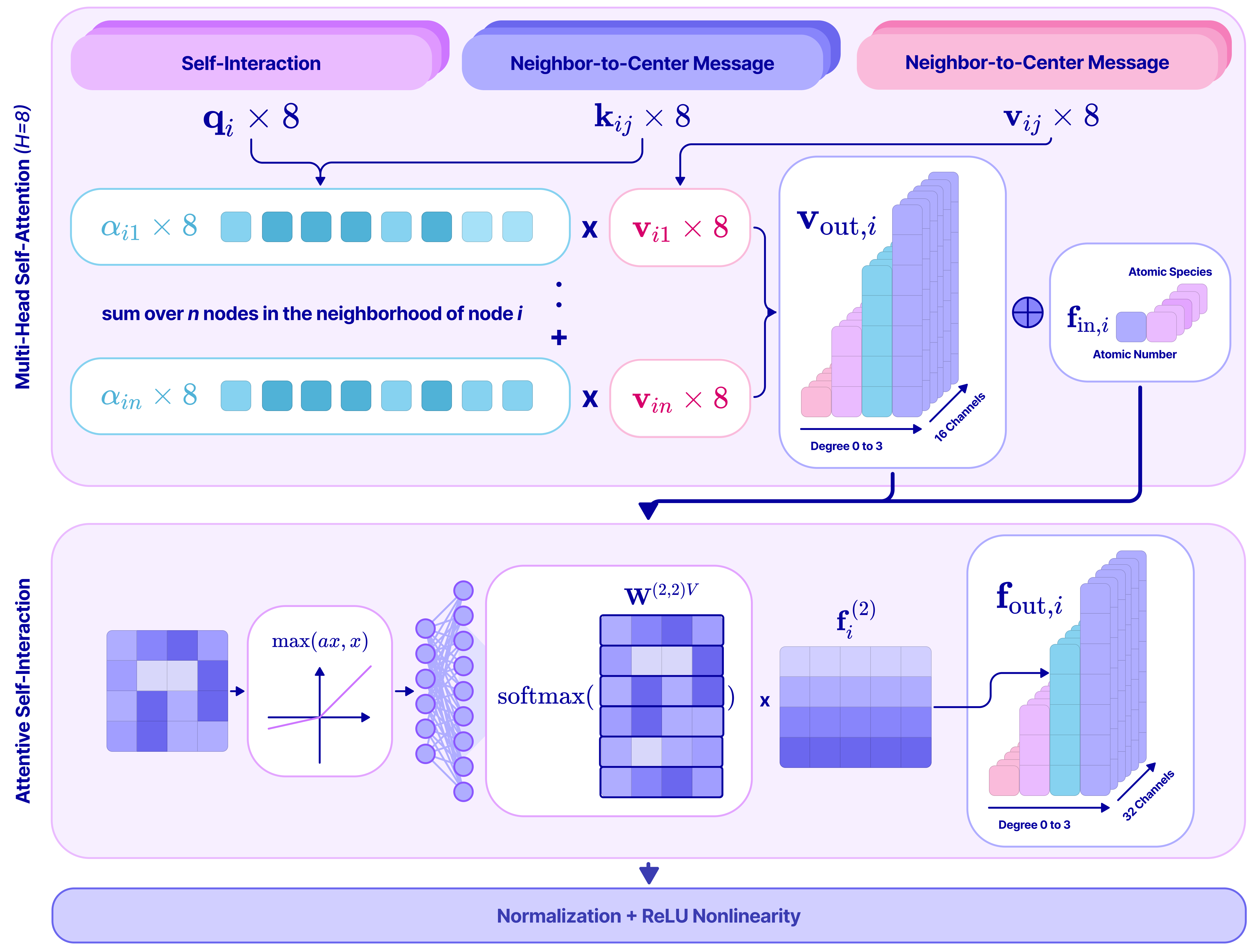}
\caption{The encoder consists of three layers: a multi-head attention layer with 8 heads, an attentive self-interaction layer that concatenates the input features to the neighbor-to-center message and projects the concatenated tensor to match the output fiber, and a norm nonlinearity layer.}
\label{fig:chem2}
\end{figure}

To learn high-dimensional equivariant features and relationships that determine the chemical properties of the molecular graph, the \textbf{encoder leverages equivariant message-passing to iteratively refine node embeddings for downstream prediction with the decoder.}

The encoder is composed of seven multi-head attention blocks, each with 8 attention heads. The first attention block takes six channels of type-0 features as input and outputs a feature tensor with type-0, type-1, type-2, and type-3 spherical tensors, each with 32 channels for every node in the graph. The input, key/query (\texttt{f\_mid\_in}), value (\texttt{f\_mid\_out}), and output fibers are defined below:

\begin{enumerate}
    \item $\texttt{f\_in} = [(\text{input channels}, \text{input degree})] = [(4, 0)]$
    \item $\texttt{f\_mid\_in} = [(16, 0)]$
    \item $\texttt{f\_mid\_out} = [(16, 0), (16, 1), (16, 2), (16, 3)]$
    \item $\texttt{f\_cat} = [(20, 0), (16, 1), (16, 2), (16, 3)]$
    \item $\texttt{f\_out} = [(\text{output channels}, \text{output degrees})]= [(32, 0), (32, 1), (32, 2), (32, 3)]$
\end{enumerate}

This block generates 16 channels of value messages for each output degree before concatenating with the input type-0 features and projecting it to match the output fiber \texttt{f\_out} using attentive self-interaction.

Since the graph contains multiple type-0 scalar edge features in addition to the radial distance, the model concatenates them into a single 5-dimensional vector, where the first dimension corresponds to the radial distance and the remaining four dimensions correspond to the one-hot encoding of the bond type. Instead of just the radial distance, this 5-dimensional vector is fed into the radial function to generate the weights used to scale the basis kernels.

The first attention block executes the multi-head self-attention layer as follows:

\begin{enumerate}
    \item Since the query embeddings are generated via self-interaction, we can only compute type-0 query embeddings from the type-0 input features. For each edge, we generate 16 query embeddings, each of which is a linear combination of the input type-0 channels. Then, we split them into 8 attention heads and concatenate the two embeddings at each head into a single 2-dimensional vector to get a query array with shape (edges, 8, 2).
    
    \item The key embeddings must have the same dimension as the query embedding so we only generate key embeddings for degree 0 with shape (edges, 8, 2). For each edge, we generate 16 key embeddings by transforming the four channels of type-0 input tensors to 16 channels of type-0 key embeddings using 4*16 = 64 unique $1 \times  1$ (scalar) key kernels corresponding to each input and output channel pair. Then, we split them into 8 attention heads and concatenate the two embeddings at each head into a single 2-dimensional vector.
    
    \item For each edge, 8 unique attention weights are generated from the dot product of the query and key vectors for the 8 attention heads, each of which scales a total of 8 value embeddings (2 for each output degree).
    
    \item The value array corresponding to each output degree has shape (edges, 8 (heads), 2 (channels per head), $2l  + 1$) for $l  = 0, 1, 2, 3$. For each edge and each output degree $l$, the 4 type-0 input channels are transformed into 16 channels of type-$l$ value embeddings using 4*16 = 64 unique $(2l+1) \times 1$ value kernels corresponding to each input and output channel pair. These value embeddings are then divided into 8 attention heads with 2 channels per degree per head.
\end{enumerate}

Every channel of every node in the graph is updated with the weighted sum of the value embeddings from incoming edges.

The first attention block is followed by six identical attention blocks with the same input and output fiber. The intermediate value fiber divides the channels in half before projecting them to match the output fiber using attentive self-interaction. The fiber structures for all six attention blocks are defined below:

\begin{enumerate}
    \item $\texttt{f\_in} = [(32, 0), (32, 1), (32, 2), (32, 3)]$
    \item $\texttt{f\_mid\_in} = [(16, 0), (16, 1), (16, 2), (16, 3)]$
    \item $\texttt{f\_mid\_out} = [(16, 0), (16, 1), (16, 2), (16, 3)]$
    \item $\texttt{f\_cat} = [(48, 0), (48, 1), (48, 2), (48, 3)]$
    \item $\texttt{f\_out} = [(32, 0), (32, 1), (32, 2), (32, 3)]$
\end{enumerate}

There are some key differences from the first attention block, given the higher dimension of the input features:

\begin{enumerate}
    \item The value array for each degree has the same shape as the first attention block. Since there are 4*16 = 64 input channels and 64 output channels across all degrees, there are a total of $64^2$ unique value kernels for each input-output channel/degree pair.
    
    \item Instead of only type-0 query embeddings, 16 query embeddings of each type from 0 to 4 are generated from the self-interaction of the input features. These are split into heads and concatenated into a single vector per head, such that the query array has shape (edges, 8, dimension of query/8).
    
    \item The key embedding for a given channel of a given degree in the center node is generated by aggregating transformed messages from every channel of every degree in the source node. Just like the query embedding, they are split into heads and concatenated into an array with shape (edges, 8, dimension of query/8).
\end{enumerate}

Each attention block is followed by a norm nonlinearity layer that normalizes across all the features of the same degree for every node in the graph and incorporates the ReLU nonlinearity. The output of the final attention block is a fiber with structure $[(32, 0), (32, 1), (32, 2), (32, 3)]$ that is fed into the decoder.

\subsection{Decoder}

\purple[]{The decoder consists of a TFN module that transforms the high-dimensional features back down to scalar (type-0) feature channels. These are invariant to rotation, so they can be passed into a FFN to generate the final prediction.}

\begin{figure}[h!]
\centering
\includegraphics[width=0.8\linewidth]{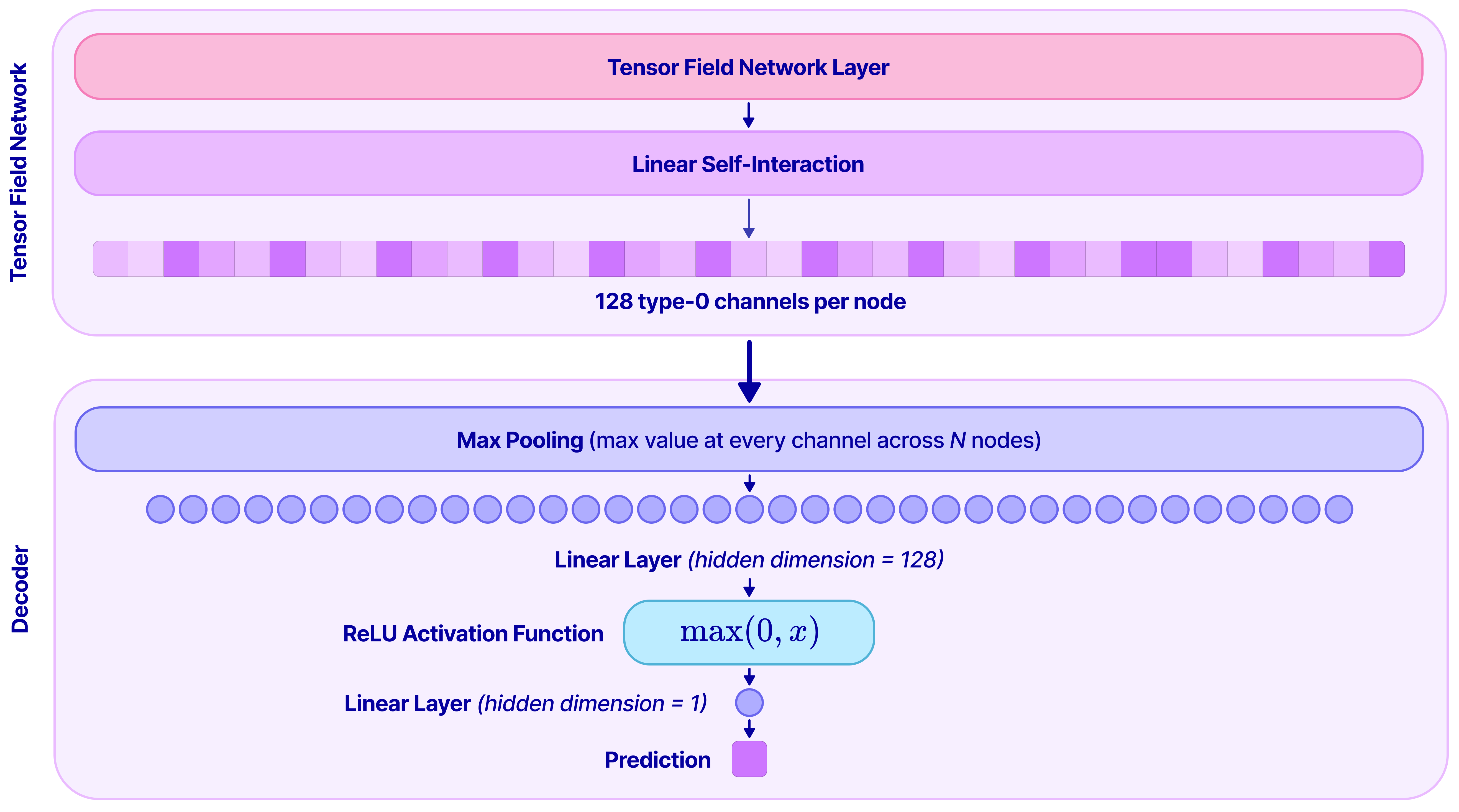}
\caption{The decoder is composed of a TFN layer that reduces all features into 128 type-0 channels, a max pooling layer that reduces the graph into a single 128-dimensional vector, a linear layer with hidden dimension 128, a ReLU activation function, and a final linear layer that generates a scalar prediction.}
\label{fig:chem3}
\end{figure}

\textbf{To generate the final prediction, the decoder reduces the equivariant node representations to rotation-invariant scalar (type-0) features through equivariant processing and invariant aggregation}.

The first block of the decoder is a TFN layer that converts the fiber with structure [(32, 0), (32, 1), (32, 2), (32, 3)] to a fiber with structure [(128, 0)]. Each type-0 output channel of the TFN layer is generated by adding a neighbor-to-center message with a linear self-interaction message.

\begin{enumerate}
    \item For each edge, there are 128 total input channels across all degrees and 128 type-0 output channels, so the TFN layer uses a total of $128^2$ unique $1 \times (2k+1)$ equivariant kernels that transform from type-$k$ input channels for $k = 0, \dots,  3$ to type-0 output channels. The neighbor-to-center message for a single node is the average of the type-0 messages across all incoming edges.
    \item A linear self-interaction message is computed for all 128 type-0 output channels as a weighted sum of the 32 type-0 input channels. This type-0 message is added to the neighbor-to-center message and acts as a skip connection in gradient descent.
\end{enumerate}

The output of the TFN module is a graph with a 128-dimensional vector at every node, where each element of the vector is a type-0 scalar invariant to rotation. This means that we can apply any function to these vectors without breaking the equivariance of the model.

The resulting graph is fed into a max pooling layer that reduces the feature data across all nodes into a single set of 128 type-0 channels by extracting the maximum type-0 scalar at each channel $c_k$ across every node in the graph.

\begin{align}
    \mathbf{f}^{(0)}_{c_k}=\max_{i\in N}(\mathbf{f}^{(0)}_{i,c_k})
\end{align}

The implementation of the max pooling mechanism for graph inputs is given below:

\begin{lstlisting}[language=Python]
from dgl.nn.pytorch.glob import MaxPooling

class GMaxPooling(nn.Module):
    def __init__(self):
        super().__init__()
        self.pool = MaxPooling()

    @profile
    def forward(self, features, G, **kwargs):
        # extracts the type-0 features in the graph reshapes into an array with shape (nodes, type-0 channels)
        h = features['0'][...,-1]
        # returns the maximum scalar across nodes for each type-0 channel 
        return self.pool(G, h)
\end{lstlisting}

An alternative way to reduce the type-0 feature data across the graph is by using an average pooling layer, where instead of taking the maximum across all nodes, we take the average.

The pooling layer reduces the graph into a single $128$-dimensional vector that is fed into an FFN. The first linear layer of the FFN transforms the vector without reducing its dimension with a $128 \times 128$ learnable weight matrix and $128$-dimensional bias vector. Then, a ReLU activation function is applied element-wise to the vector. The final linear layer transforms the 128-dimensional vector into a single scalar prediction with a $128 \times  1$ weight matrix and a single bias.

\newpage
\section{Conclusion}
This guide aims to serve as a self-contained and intuitive introduction to spherical equivariant graph transformers, to demystify both the mathematical foundations and the architectural design choices that underlie modern SE(3)-equivariant models. By starting from basic concepts in group theory and representation theory and progressively building toward full neural architectures, we have shown how equivariance is not an abstract constraint imposed after the fact, but rather a guiding principle that shapes every component of the model, from feature representations and kernels to message passing and attention.

Equivariant architectures have become a central tool across machine learning, particularly in scientific applications where geometry and symmetry play a fundamental role \citep{han2025survey, zhang2025artificial}. Within this broader landscape, spherical equivariance has emerged as a critical component of modern generative and predictive models. It underpins state-of-the-art approaches to protein modeling and design \citep{krishna2024generalized}, target-aware molecular generation \citep{hu2025target}, and machine-learned interatomic potentials for molecular dynamics and materials simulation \citep{park2024scalable, wang2024machine}. Spherical equivariant architectures have also driven recent advances in crystal structure prediction \citep{lin2025equivariant} and molecular representation learning \citep{wang2024enhancing}.

One of the key messages of this guide is that equivariance is not an add-on feature, but a design principle inherent in the architecture. Spherical equivariant graph transformers differ from familiar graph neural networks and attention-based models fundamentally through spherical tensor representations and equivariant kernels. This perspective not only demystifies existing equivariant architectures but also provides a foundation for developing new models that respect additional symmetries, incorporate dynamics, or operate within generative frameworks.

Looking ahead, spherical equivariant models are likely to play an increasingly important role in scientific machine learning \citep{zhang2025artificial}. Ongoing work on equivariant diffusion models \citep{schneuing2024structure}, flow-based generative models \citep{hassan2024flow}, and symmetry-aware molecular simulators \citep{wang2024machine} suggests a future in which geometric inductive biases are tightly integrated with probabilistic modeling and large-scale learning. We hope that this guide serves both as a practical reference and as an entry point for researchers and practitioners seeking to understand, apply, and extend spherical equivariant graph transformers in emerging applications.

\paragraph{Acknowledgements} I am thankful for the positive feedback from the community, which inspired the publication of this guide with a permanent DOI. I hope it continues to support researchers and learners interested in geometric deep learning.

\newpage
\printbibliography

\end{document}